%% file: acl_latex.tex
\pdfoutput=1

\documentclass[11pt]{article}

\usepackage[final]{acl}

\usepackage{times}
\usepackage{latexsym}

\usepackage[T1]{fontenc}

\usepackage[utf8]{inputenc}

\usepackage{microtype}

\usepackage{inconsolata}

\usepackage{graphicx}

\usepackage{enumitem}
\usepackage{hyperref}
\usepackage{amssymb}
\usepackage{url}
\usepackage{multirow}    
\usepackage{booktabs}    
\usepackage{geometry}    
\usepackage{subcaption}
\usepackage{ulem}
\usepackage{float}
\usepackage{placeins}
\usepackage{wrapfig}
\usepackage{caption}    
\usepackage{array}      
\usepackage{geometry}   
\usepackage{subcaption}
\usepackage{amsmath}
\usepackage{cleveref}
\usepackage{arydshln}
\usepackage{makecell}
\title{Beyond One-Size-Fits-All: Tailored Benchmarks for Efficient Evaluation}

\author{Peiwen Yuan$^1$\footnotemark[1], Yueqi Zhang$^1$\footnotemark[1], Shaoxiong Feng$^2$, Yiwei Li$^1$, Xinglin Wang$^1$, Jiayi Shi$^1$\\ {\bf Chuyi Tan$^1$, Boyuan Pan$^2$, Yao Hu$^2$, Kan Li$^{1}$\footnotemark[2]}\\
  $^1$School of Computer Science and Technology, Beijing Institute of Technology \\
  $^2$Xiaohongshu Inc \\
  \texttt{\{peiwenyuan,zhangyq,liyiwei,wangxinglin,shijiayi,tanchuyi,likan\}@bit.edu.cn} \ \ \\ 
  \texttt{\{shaoxiongfeng2023,whd.thu\}@gmail.com} \ \  \texttt{\{panboyuan,xiahou\}@xiaohongshu.com}}

\begin{document}
\maketitle
\renewcommand{\thefootnote}{\fnsymbol{footnote}} 
\footnotetext[1]{Equal contribution.} 
\footnotetext[2]{Corresponding author.} 
\input{abstract}
\input{intro}

\input{relatedwork}
\input{method}
\input{experiment}

\input{conclusion}

\bibliography{custom}
\clearpage
\appendix
\input{appendix}

\end{document}

%% file: abstract.tex
\begingroup
  \renewcommand{\thefootnote}{\arabic{footnote}}
  \setcounter{footnote}{0} 
\begin{abstract}
Evaluating models on large benchmarks is very resource-intensive, especially during the  period of rapid model evolution.
Existing efficient evaluation methods estimate the performance of target models by testing them only on a small and static coreset of the benchmark, which is derived from the publicly available evaluation results of source models.
These methods rely on the assumption that target models have high prediction consistency with source models. However, we demonstrate that it doesn’t generalize well in practice.
To alleviate the inconsistency issue, we present \textsc{TailoredBench}, a method that conducts customized evaluation tailored to each target model. 
Specifically, a Global-coreset is first constructed as a probe to identify the most consistent source models for each target model with an adaptive source model selection strategy. 
Afterwards, a scalable K-Medoids clustering algorithm is proposed to extend the Global-coreset to a tailored Native-coreset for each target model. 
According to the predictions on Native-coresets, we obtain the performance of target models on the whole benchmark with a calibrated estimation strategy. 
Comprehensive experiments on 5 benchmarks across over 300 models demonstrate that compared to best performing baselines, \textsc{TailoredBench} achieves an average reduction of 31.4\% in MAE of accuracy estimates under the same inference budgets, showcasing strong effectiveness and generalizability\footnote{Our code is available at \url{https://github.com/marvelcell/TailoredBench}}.
\end{abstract}
\endgroup

%% file: intro.tex
\section{Introduction}

Scaling up models in multiple dimensions has led to remarkable advancements in their capabilities \citep{llama2,insgpt}, which also presents significant challenges for efficiently assessing them. For instance, \citet{liang2022holistic} reports that evaluating a model with approximately 10 billion parameters on the HELM leaderboard costs over \$1,700 via APIs or more than 1,200 GPU hours. Moreover, these costs scale by a factor of $X$ when exploring and comparing $X$ different training or inference configurations during the development or deployment phase.



To achieve efficient evaluation, some studies \citep{AP,tiny} have explored the following paradigm: \textit{step 1.} constructing example embeddings according to the predictions from a set of \textbf{\textit{source models}} (which are freely available for popular leaderboards\footnote{\url{https://huggingface.co/open-llm-leaderboard}}\textsuperscript{,}\footnote{\url{https://rank.opencompass.org.cn}}\textsuperscript{,}\footnote{\url{https://crfm.stanford.edu/helm}}); \textit{step 2.} clustering the benchmark and selecting the cluster centroids to form a coreset (typically less than 100 examples); \textit{step 3.} approximating the performance of \textbf{\textit{target models}} under evaluation based on their predictions on the coreset. Underlying this approach is the assumption that performance patterns generalize: if source models respond similarly to two examples $a$ and $b$, then a target model’s performance on $a$ can be used to estimate its performance on $b$.
\begin{figure*}[t]
    \centering
    \begin{subfigure}[b]{0.95\textwidth}
        \includegraphics[width=\textwidth]{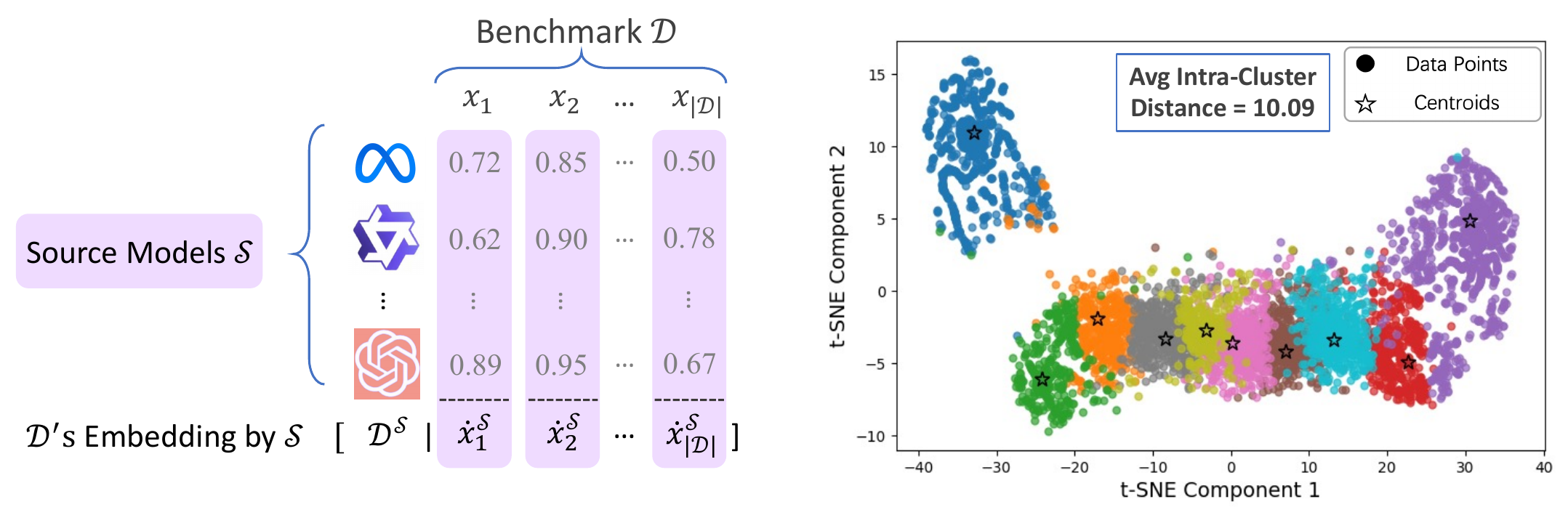}
        \vspace{-0.6cm}
        \caption{Hellaswag benchmark represented by source-model embeddings $\mathcal{D}^\mathcal{S}$.}
        \label{FigDistShift_a}
    \end{subfigure}\\
    \begin{subfigure}[b]{0.95\textwidth}
        \includegraphics[width=\textwidth]{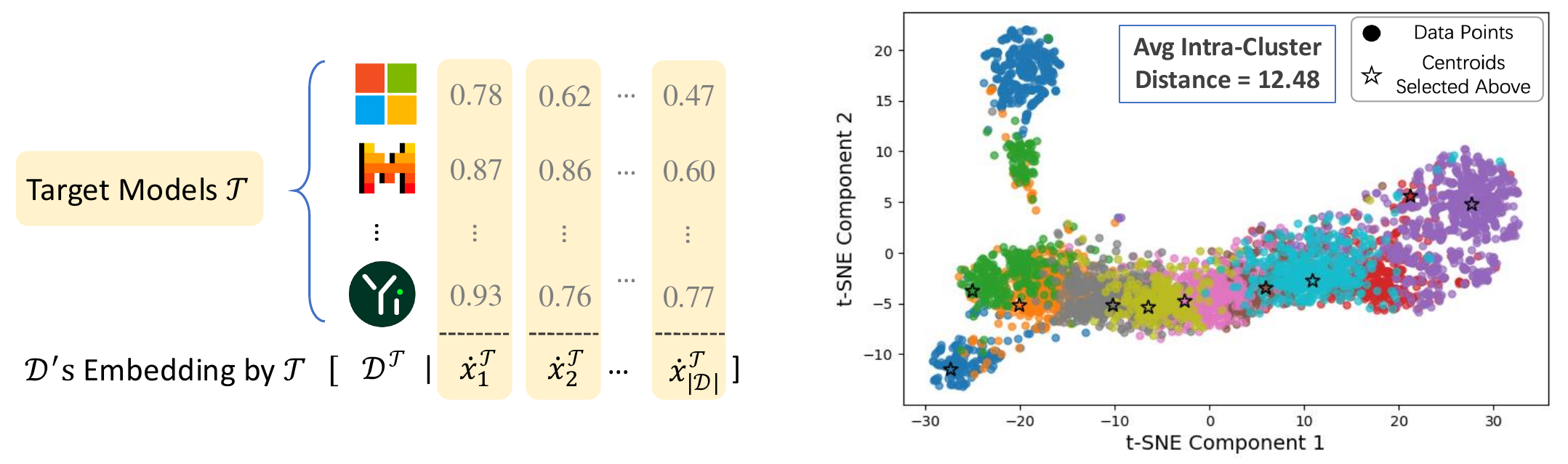}
        \vspace{-0.6cm}
        \caption{Hellaswag benchmark represented by target-model embeddings $\mathcal{D}^\mathcal{T}$.}
        \label{FigDistShift_b}
    \end{subfigure}
    \vspace{-0.2cm}
    \caption{The t-SNE visualization of the Hellaswag benchmark using embeddings derived from source (above) and target (below) models' predictions. The increased average distance between examples and their cluster centroids in the target-based embedding indicates that the coreset (centroids) obtained from source-based embeddings no longer effectively represents the entire benchmark for target models.}
    \label{FigDistShift}
    \vspace{-0.6cm}
\end{figure*}

Nevertheless, we find that such generalizability between source and target models does not necessarily hold.
Following \textsc{AnchorPoint} \citep{AP}, we construct an embedding based on the correctness (e.g., the probability of the correct option) of all source models for each example and visualize them using t-SNE algorithm \citep{tsne}.
In these embeddings (Figure~\ref{FigDistShift_a}), nearby examples elicit similar predictions from the source models, allowing cluster centroids (marked by stars) to serve as representative points. Yet, when we adopt embeddings derived from the correctness of target models instead (Figure~\ref{FigDistShift_b}), the average distance between the example and its centroid increases from 10.09 to 12.48, indicating that the previously chosen centroids fail to represent their respective clusters effectively.
This reveals a discrepancy in prediction behaviors between source and target models, which we term \textit{prediction consistency}—the extent to which their predictions align on the same examples. When prediction consistency is low, source-model-derived coresets fail to generalize, resulting in inaccurate performance estimates for target models.

To address the aforementioned issue, we propose the \textsc{TailoredBench} method, which adaptively constructs model-specific evaluation coreset in a global to native manner for accurate and efficient evaluation.
Specifically, we first construct a static G-set (Global-coreset) based on the prediction results of all the source models.
By applying an adaptive source model selection strategy, the predictions of target models on the G-set are used as a probe to select a native source model set for each target model that has stronger prediction consistency with them. 
Based on this posterior, we design a scalable K-Medoids clustering technique to expand the G-set into an N-set (Native-coreset) for each target model, according to the benchmark embeddings under the metric of corresponding native source models. Finally, we approximate the overall performance of target models by employing a calibrated estimation strategy based on their predictions on the N-set. 

We conduct extensive experiments on five benchmarks across more than 300 models, involving tasks in the fields of natural language and multi-modality. Compared to non-customized efficient evaluation baselines, \textsc{TailoredBench} can more accurately estimate the performance of models (attaining an average of 31.4\% MAE degradation improvement on accuracy) under the same small-size inference budgets (generally 20\textasciitilde 40 examples). Our contributions are summarized as follows:
\vspace{-0.2cm}
\begin{itemize}[leftmargin=20pt]
\setlength{\itemsep}{0pt}
\setlength{\parsep}{0pt}
\setlength{\parskip}{0pt}
\item We analyze that the existing efficient evaluation methods overestimate the prediction consistency across models, thus the source-model-based static coreset may fail to assess the target models accurately.
\item We propose the \textsc{TailoredBench} method to conduct tailored evaluation on adaptively constructed N-set for each target model to attain more accurate evaluation results.
\item We conduct comprehensive experiments and analyses on multiple settings to validate the excellent effectiveness and strong generalizability of \textsc{TailoredBench}.
\end{itemize}

%% file: relatedwork.tex
\section{Related Works}
As LLMs proliferate and version updates accelerate, the cost of thoroughly evaluating each model across all benchmarks has become prohibitive, leading to methods that subsample the most representative subsets from each benchmark for more efficient evaluation. \cite{AP} clusters examples directly using the confidence scores provided by source models, leveraging these scores to select an optimal subset. Similarly, \cite{tiny} employs an Item Response Theory (IRT) model, trained on the success matrix of each source model across various examples, to derive the latent representations of examples for clustering. \cite{100instances} introduces a generic assessor framework that predicts the performance of a new LLM on unseen instances using its results on a small reference set, achieving comparable accuracy to full-scale evaluations. \cite{perlitz2023efficient} proposes Flash-HELM, which dynamically adjusts the sizes of randomly selected subsets based on model ranking, where higher-ranked models are evaluated with greater precision. \cite{prabhu2024lifelong} proposes the Sort \& Search (S\&S) strategy, which leverages the difficulties of examples and dynamic programming to select the coreset. (See more related works in Appendix~\ref{apd:related_works}.) 

%% file: method.tex
\begin{figure*}[ht]
    \centering
    \includegraphics[scale=0.36]{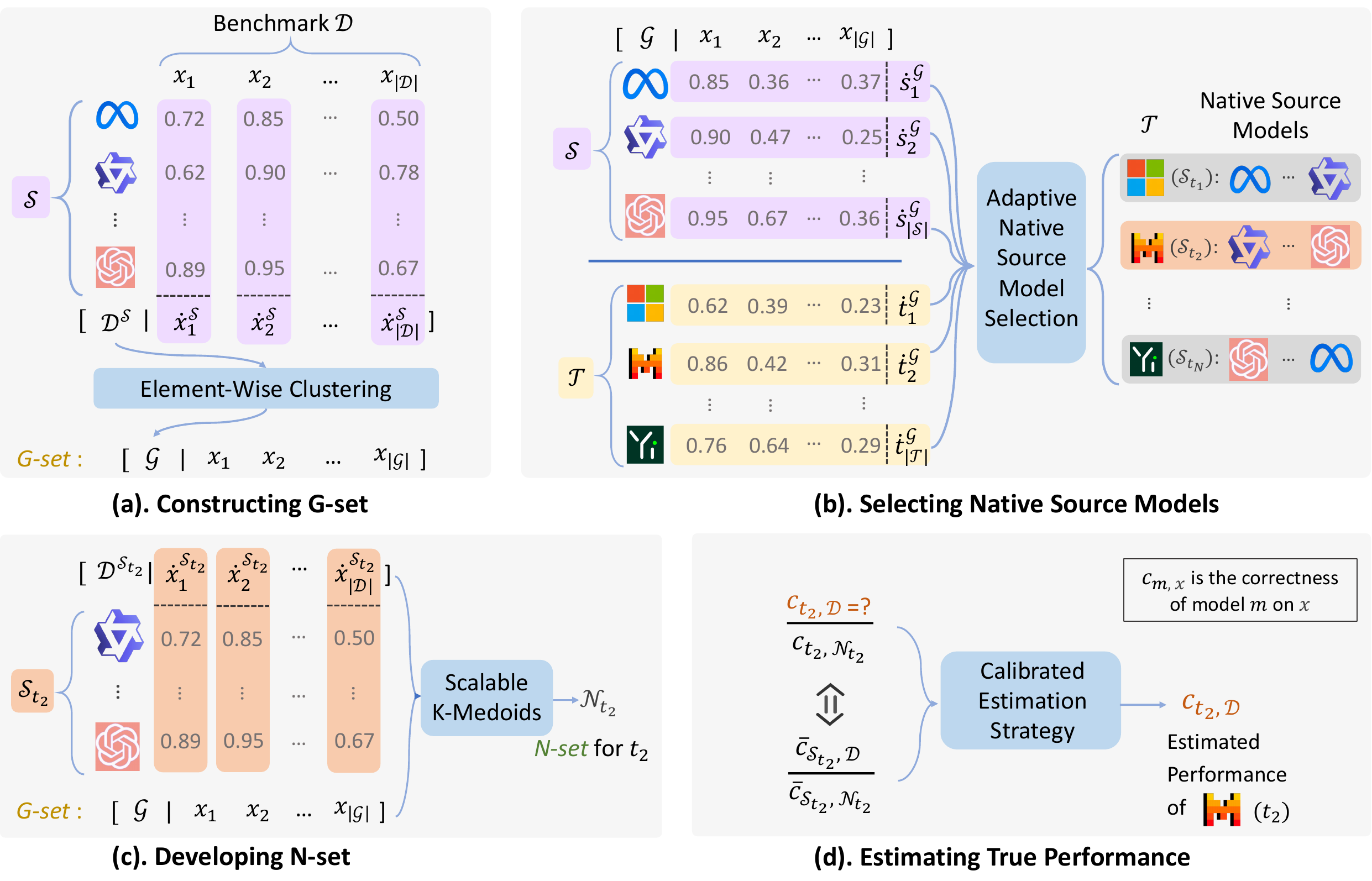}
    \vspace{-0.1cm}
    \caption{Overview of \textsc{TailoredBench}.}
    \label{FigMethod}
    \vspace{-0.4cm}
\end{figure*}
\section{TailoredBench Approach}

The \textsc{TailoredBench} approach centers on dynamically selecting prediction-consistent source models and crafting an N-set that faithfully represents the entire benchmark for each target model.
Its formulation proceeds through four tightly integrated steps: constructing a globally representative G-set (\S\ref{sec:G-set}), identifying native source models (\S\ref{sec:AdpSM}) and developing N-set for each target model (\S\ref{sec:N-set}), and finally estimating the target models' overall performance (\S\ref{sec:EsmPerform}).

\subsection{Task Set-Up}
Let $\mathcal{D} = \left\{(x_{k}, y_{k})\right\}_{k=1}^{|\mathcal{D}|}$ denotes a benchmark, where $x_{k}$ is the input and $y_{k}$ is the corresponding ground truth output. We define the set of target models under evaluation as $\mathcal{T} = \{ t_{m} \}_{m=1}^{|\mathcal{T}|}$. Additionally, we denote the source model set as $\mathcal{S} = \{ s_{n} \}_{n=1}^{|\mathcal{S}|}$, for which we have access to their predictions across all examples in $\mathcal{D}$. Following previous works, we ensure that $\mathcal{T} \cap \mathcal{S} = \varnothing$. Our objective is to accurately estimate the performance $P_{t_m}$ of each target model $t_m \in \mathcal{T}$ and determine the ranking relationships within $\mathcal{T}$, while minimizing the model inference cost.

\subsection{Constructing G-set}
\label{sec:G-set}
We first construct the G-set $\mathcal{G}$, which is designed as a probe for each target model to identify a set of source models with the highest prediction consistency. 
Consequently, it is intended to be a small yet relatively representative subset of the benchmark, ensuring its generalizability across target models. 

Following prior works \citep{AP}, we employ clustering based on the correctness of source models to construct the G-set. Here, correctness can be either the predictive probability of the correct option (continuous value [0, 1]) or whether the model answers the example correctly (discrete binary value \{0, 1\}).

For each example $x_k$ in the benchmark $\mathcal{D}$, we compute an embedding using the correctness scores $c_{s_n,x_k}$ from each source model $s_n$:
\begin{equation}
    \dot{x}_{k}^{\mathcal{S}} = \begin{pmatrix} c_{s_1, x_k} \\ c_{s_2, x_k} \\ \vdots \\ c_{s_{|\mathcal{S}|}, x_k} \end{pmatrix}
\end{equation}
The superscript $\mathcal{S}$ indicates that the embedding is derived from source models' correctness, and each embedding is $|\mathcal{S}|$-dimensional. The collection of these embeddings constitutes the benchmark's representation $\mathcal{D}^{\mathcal{S}} = \{\dot{x}_k^\mathcal{S}\}_{k=1}^{|\mathcal{D}|}$.


Based on $\mathcal{D}^{\mathcal{S}}$, we apply K-Medoids clustering \citep{pam} to select the G-set with the objective function below:
\begin{equation}
    \label{eq1}
    \min_{\{\mathcal{G}, \mathcal{C}_{g}\}} \sum_{x_{g} \in \mathcal{G}} \sum_{x_{k} \in \mathcal{C}_{g} \setminus \{x_{g}\}} \texttt{Dis} (\dot{x}_{g}^{\mathcal{S}}, \dot{x}_{k}^{\mathcal{S}})
\end{equation}
where $x_{g}$ is an example in the G-set $\mathcal{G} = \{x_{g}\}_{g=1}^{|\mathcal{G}|}$, and $\mathcal{C}_{g}$ is the cluster for which $x_{g}$ is the centroid. 
$\texttt{Dis}$ denotes the distance metric in clustering.

To maximize the generalization capability of our method, the choice of distance metric is critical. Previous approaches \citep{AP, miller2021accuracy, baek2022agreement, mehra2024predicting} using correlation distance \citep{ref_pearsonr} to measure example consistency often assume linear relationships in scoring patterns among models or examples. However, this assumption may not hold for discrete numerical embeddings, leading to significant performance degradation. In contrast, element-wise distance (e.g., Manhattan distance) can effectively capture individual discrepancies in correctness vectors, thereby accommodating various correctness formats. 
By default, we adopt manhattan distance as $\texttt{Dis}$ for our \textsc{TailoredBench} method. 

\subsection{Adaptive Native Source Model Selection}
\label{sec:AdpSM}
After constructing G-set $\mathcal{G}$, we attain the prediction results of target models $\mathcal{T}$ on it, which we use as a probe to construct a Native Source Model Set $\mathcal{S}_{t_m}$ that exhibits the highest prediction consistency for each $t_m \in \mathcal{T}$.

Specifically, we first embed all the source models $s_n \in \mathcal{S}$ and target models $t_m \in \mathcal{T}$ based on their prediction correctness on $\mathcal{G}$ as follows:
\begin{equation}
    \label{eq2}
    \begin{gathered}
    \dot{s}_n^{\mathcal{G}} = \begin{pmatrix} c_{s_n, x_1} \\ c_{s_n, x_2} \\ \vdots \\ c_{s_n, x_{|\mathcal{G}|}} \end{pmatrix}, \quad
    \dot{t}_m^{\mathcal{G}} = \begin{pmatrix} c_{t_m, x_1} \\ c_{t_m, x_2} \\ \vdots \\ c_{t_m, x_{|\mathcal{G}|}} \end{pmatrix}
    \end{gathered}
\end{equation}
Here, the superscript $\mathcal{G}$ denotes that each embedding dimension is derived from the model's prediction correctness on the G-set. Leveraging these embeddings, we compute the average prediction consistency $\bar{d}$ among all the models (both source and target) on the G-set as follows: 
\begin{equation}
    \begin{gathered}
    \bar{d} = \frac{2}{(|\mathcal{S}| + |\mathcal{T}|)(|\mathcal{S}| + |\mathcal{T}| - 1)} \sum_{i<j} d_{ij}, \\
    \text{where } d_{ij} = \texttt{Dis}\left(\dot{\phi}_i^{\mathcal{G}}, \dot{\phi}_j^{\mathcal{G}}\right)
    \end{gathered}
\end{equation}
In this context, $i, j \in [1, |\mathcal{S}| + |\mathcal{T}|]$ and $\phi$ represents any model from $\mathcal{S} \cup \mathcal{T}$. By computing $\bar{d}$ across all models, we establish a robust threshold that reflects the model set's similarity landscape, enabling a consistent and effective selection of native source models for each target model.

On this basis, we determine $\bar{n}$, the size of the native source model set for target models, by calculating the average number of source models whose prediction consistency with each target model exceeds the threshold $\bar{d}$ as follows:
\begin{equation}
    \begin{gathered}
    \bar{n} = \left\lfloor \frac{1}{|\mathcal{T}|} \sum_{m=1}^{|\mathcal{T}|} |\mathcal{S}_{t_m}| \right\rfloor \label{eq:bar_n}, \\
    \text{where } \mathcal{S}_{t_m} = \left\{ s_n \in \mathcal{S} \ \bigg| \ \texttt{Dis}\left(\dot{s}_n^{\mathcal{G}}, \dot{t}_m^{\mathcal{G}}\right) < \bar{d} \right\}   
    \end{gathered}
\end{equation}
For a target model $t_m$, the top $\bar{n}$ source models exhibiting the highest prediction consistency are selected to form its dynamic source model set $\mathcal{S}_{t_m}$. By standardizing the number of native source models across all target models, we ensure that each target model's feature representation maintains consistent dimensionality and informational richness during subsequent clustering.

\subsection{Developing N-set}
\label{sec:N-set}

Leveraging the selected native source models $\mathcal{S}_{t_m}$, we construct the most representative N-set $\mathcal{N}_{t_m}$ for each target model $t_m$. To maximize the utilization of the observed prediction results of target models on $\mathcal{G}$, we propose a \textsc{Scalable K-Medoids Clustering} algorithm to extend $\mathcal{G}$ into the N-set. 

Initially, each example $x_k \in \mathcal{D}$ is represented by a feature vector $\dot{x}_k^{\mathcal{S}_{t_m}}$, which is based on the correctness of its native source models $\mathcal{S}_{t_m}$. Then, our \textsc{Scalable K-Medoids Clustering} algorithm operates as follows:

\paragraph{Anchored Medoid Initialization:} We fix the examples in G-set $|\mathcal{G}|$ as initial medoids. To reach an N-set size of $|\mathcal{N}_{t_m}|$, we randomly add $|\mathcal{N}_{t_m}| - |\mathcal{G}|$ additional examples from $\mathcal{D} \setminus \mathcal{G}$ to form the initial medoid set.

\paragraph{Cluster Assignment:} Assign each example $x_k \in \mathcal{D}$ to the nearest medoid $x_{\mu}$ to form the cluster $\mathcal{C}_{\mu}$:
\begin{equation}
    \begin{gathered}
    x_k \in \mathcal{C}_{\mu},\\ \text{ where } \mu=\arg\min_{\mu} \texttt{Dis} \left( x_{\mu}^{S_{t_m}}, x_k^{S_{t_m}}\right)
    \end{gathered}
\end{equation}
\paragraph{Dynamic Medoid Refinement:} For each cluster $\mathcal{C}_{\mu}$ with a non-G-set medoid, update the medoid $x_{\mu}$ by selecting the example within $\mathcal{C}_{\mu}$ that minimizes the total distance to all other examples in $\mathcal{C}_{\mu}$:
\begin{equation}x_{\mu}=\arg\min_{x_i\in\mathcal{C}_{\mu}}\sum_{x_j\in{\mathcal{C}_{\mu}}\setminus{\{x_i\}}}\texttt{Dis}\left(x_i^{\mathcal{S}_{t_m}}, x_j^{\mathcal{S}_{t_m}}\right)
\end{equation}
Medoids corresponding to G-set remain fixed during this process.

\paragraph{Convergence Verification:} Repeat the Cluster Assignment and Dynamic Medoid Refinement steps until convergence is achieved, i.e., when medoids no longer change or a maximum number of iterations is reached.

By incorporating the G-set examples as fixed medoids, the clustering process ensures that these pivotal examples guide the formation of clusters and the selection of additional N-set examples.

\subsection{Calibrated Performance Estimation}
\label{sec:EsmPerform}
After establishing the N-set $\mathcal{N}_{t_m}$ for a target model $t_{m}$, for previous methods \cite{AP,tiny}, they may estimate the model’s overall performance by first evaluating it on these centroid examples and then weighting the results according to each centroid’s coverage of the benchmark. However, simply relying on medoids overlooks subtle variations in how individual examples within each cluster are predicted, potentially leading to less accurate global estimates.

To address this, we leverage the prediction consistency between the target model $t_m$ and its native source models $\mathcal{S}_{t_m}$ to obtain the calibrated correctness estimates for the target model. For a given cluster with medoid $x$, consider any non-medoid example $x^{\prime}$ in the same cluster. We compute a scaling factor based on the native source models' average correctness, which reflects how the prediction patterns at $x^{\prime}$ relate to those at $x$:
\begin{equation}
    \mathrm{Scale}(x^{\prime})=\frac{\bar{c}_{\mathcal{S}_{t_m},x^{\prime}}+0.5}{\bar{c}_{\mathcal{S}_{t_m},x}+0.5}
\end{equation}
Here, $\bar{c}_{S_{t_m},x}$ and $\bar{c}_{S_{t_m},x^{\prime}}$ denote the average correctness of $\mathcal{S}_{t_m}$ on the medoid $x$ and the non-medoid $x^{\prime}$, respectively. The addition of 0.5 ensures numerical stability by preventing the denominator from becoming zero. Given that $t_m$ and $\mathcal{S}_{t_m}$ exhibit similar prediction consistencies, we assume this scaling factor can be applied to estimate the target model's correctness on $x^{\prime}{:}$
\begin{equation}
    c_{t_m,x^{\prime}}=(c_{t_m,x}+0.5)\cdot\mathrm{Scale}(x^{\prime})-0.5
\end{equation}
By integrating these inferred correctness values across all examples in the benchmark $\mathcal{D}$, we obtain a more faithful global performance estimation without re-evaluating the entire dataset:
\begin{equation}
    P_{t_m}=\frac1{|\mathcal{D}|}\sum_{x^{\prime}\in\mathcal{D}}c_{t_m,x^{\prime}}
\end{equation}

%% file: experiment.tex
\section{Experiments}
\subsection{Experimental Setup}
\label{sec:exp-3.1}
\begin{table*}[ht]
\renewcommand\arraystretch{1.2}
\centering
\small
\setlength{\tabcolsep}{0.49em} 
\begin{tabular}{cl*{5}{cc}}
\toprule
\multirow{2}{*}{\textbf{Benchmarks}} & \multirow{2}{*}{\textbf{Inference counts}} & \multicolumn{2}{c}{20} & \multicolumn{2}{c}{25} & \multicolumn{2}{c}{30} & \multicolumn{2}{c}{35} & \multicolumn{2}{c}{40} \\
&  & \textbf{$\tau \uparrow$} & {\scriptsize \textbf{MAE} $\downarrow$}  & \textbf{$\tau \uparrow$} & {\scriptsize \textbf{MAE} $\downarrow$}  & \textbf{$\tau \uparrow$} & {\scriptsize \textbf{MAE} $\downarrow$}  & \textbf{$\tau \uparrow$} & {\scriptsize \textbf{MAE} $\downarrow$}  & \textbf{$\tau \uparrow$} & {\scriptsize \textbf{MAE} $\downarrow$}  \\
\midrule
\multirow{2}{*}{ARC Challenge}&\textsc{Best Baseline} & 0.662 & 0.046 & 0.663 & 0.046 & 0.676 & 0.036 & 0.713 & 0.036 & 0.714 & 0.029 \\
                           & \textsc{TailoredBench} & \textbf{0.711} & \textbf{0.031} & \textbf{0.737} & \textbf{0.029} & \textbf{0.756} & \textbf{0.028} & \textbf{0.766} & \textbf{0.027} & \textbf{0.773} & \textbf{0.027} \\
\hdashline
\multirow{2}{*}{Hellaswag}&\textsc{Best Baseline} & 0.860 & 0.060 & 0.880 & 0.053 & 0.877 & 0.043 & 0.897 & 0.038 & 0.898 & 0.032 \\
                           & \textsc{TailoredBench} & \textbf{0.900} & \textbf{0.020} & \textbf{0.909} & \textbf{0.018} & \textbf{0.913} & \textbf{0.018} & \textbf{0.914} & \textbf{0.017} & \textbf{0.918} & \textbf{0.017} \\
\hdashline
\multirow{2}{*}{GSM8K}&\textsc{Best Baseline} & 0.811 & 0.055 & 0.828 & 0.047 & 0.839 & 0.041 & 0.847 & 0.038 & 0.858 & 0.034 \\
                           &\textsc{TailoredBench} & \textbf{0.852} & \textbf{0.035} & \textbf{0.858} & \textbf{0.034} & \textbf{0.863} & \textbf{0.033} & \textbf{0.869} & \textbf{0.031} & \textbf{0.878} & \textbf{0.029} \\
\hdashline
\multirow{2}{*}{Winogrande}&\textsc{Best Baseline} & 0.472 & 0.041 & 0.487 & 0.038 & 0.514 & 0.038 & 0.521 & 0.036 & 0.518 & 0.034 \\
                           & \textsc{TailoredBench} & \textbf{0.565} & \textbf{0.028} & \textbf{0.568} & \textbf{0.026} & \textbf{0.604} & \textbf{0.024} & \textbf{0.608} & \textbf{0.023} & \textbf{0.618} & \textbf{0.022} \\
\hdashline
\multirow{2}{*}{POPE}&\textsc{Best Baseline} & 0.488 & 0.038 & 0.510 & 0.037 & 0.518 & 0.034 & 0.547 & 0.033 & 0.556 & \textbf{0.031} \\
                           &\textsc{TailoredBench} & \textbf{0.521} & \textbf{0.036} & \textbf{0.547} & \textbf{0.035} & \textbf{0.562} & \textbf{0.031} & \textbf{0.563} & \textbf{0.031} & \textbf{0.574} & 0.032 \\
\bottomrule
\end{tabular}
\caption{Results on all benchmarks. For each setting, we take the best result from multiple baselines to compare with \textsc{TailoredBench}. The detailed performance of each baseline under each setting is presented in Table~\ref{tab:apdallresult}. Values in bold represent the best results.}
\label{tab:allresult}
\vspace{-0.4cm}
\end{table*}

\paragraph{Benchmarks and Models} 
We validate \textsc{TailoredBench} on five diverse benchmarks spanning natural language and multimodal tasks. ARC Challenge \citep{clark2018arc} consists of 1,172 scientific reasoning questions, with predictions from 153 models. Hellaswag \citep{zellers2019hellaswag} provides 6,000 commonsense inference examples (a subset of its validation set) and outputs from 139 models. GSM8K \citep{cobbe2021GSM8K} includes 1,319 math reasoning problems tested on 150 models. Winogrande \citep{sakaguchi2021winogrande} has 1,267 pronoun resolution examples with 150 models evaluated. POPE \citep{li2023pope} features 5,127 instances for assessing multimodal hallucination, accompanied by results from 99 models. A complete list of models used for each benchmark is provided in Appendix \ref{apd:modellist}. We randomly split models into source and target sets for each benchmark, ensuring that their intersection is empty. 

For ARC Challenge and Hellaswag, model correctness is represented by continuous probabilities, while GSM8K, Winogrande, and POPE use binary correctness \{0, 1\}. Predictions for ARC Challenge, Hellaswag, GSM8K, and Winogrande come from the Open LLM Leaderboard \citep{open-llm-leaderboard}, and those for POPE are from the OpenCompass Leaderboard \citep{2023opencompass}.

\paragraph{Baseline and Evaluation Metrics} We compare \textsc{TailoredBench} against three baselines: a \textit{Random Sampling} strategy that randomly selects a subset of examples from the benchmark to estimate model performance, serving as a basic reference point; the \textit{Anchor Points} method \citep{AP}, which uses K-Medoids clustering on source-model predictions to identify a fixed representative coreset; and \textit{gp-IRT} \citep{tiny}, which employs an Item Response Theory model trained on the predictions of the source models to estimate target models' performance on the full benchmark. In all cases, we use the same source models and target models to ensure a fair comparison.


We employ two metrics to assess these methods. \textit{Kendall's $\tau$ Correlation Coefficient} evaluates the ordinal agreement between estimated and true model rankings, indicating how well the relative performance order is preserved. \textit{Mean Absolute Error (MAE)} measures the average absolute deviation between estimated and true performance scores, thereby capturing the precision of performance estimation for individual target models.

\begin{figure*}[ht]
    \centering
    \begin{subfigure}[b]{0.49\textwidth} 
        \centering 
        \includegraphics[width=\linewidth]{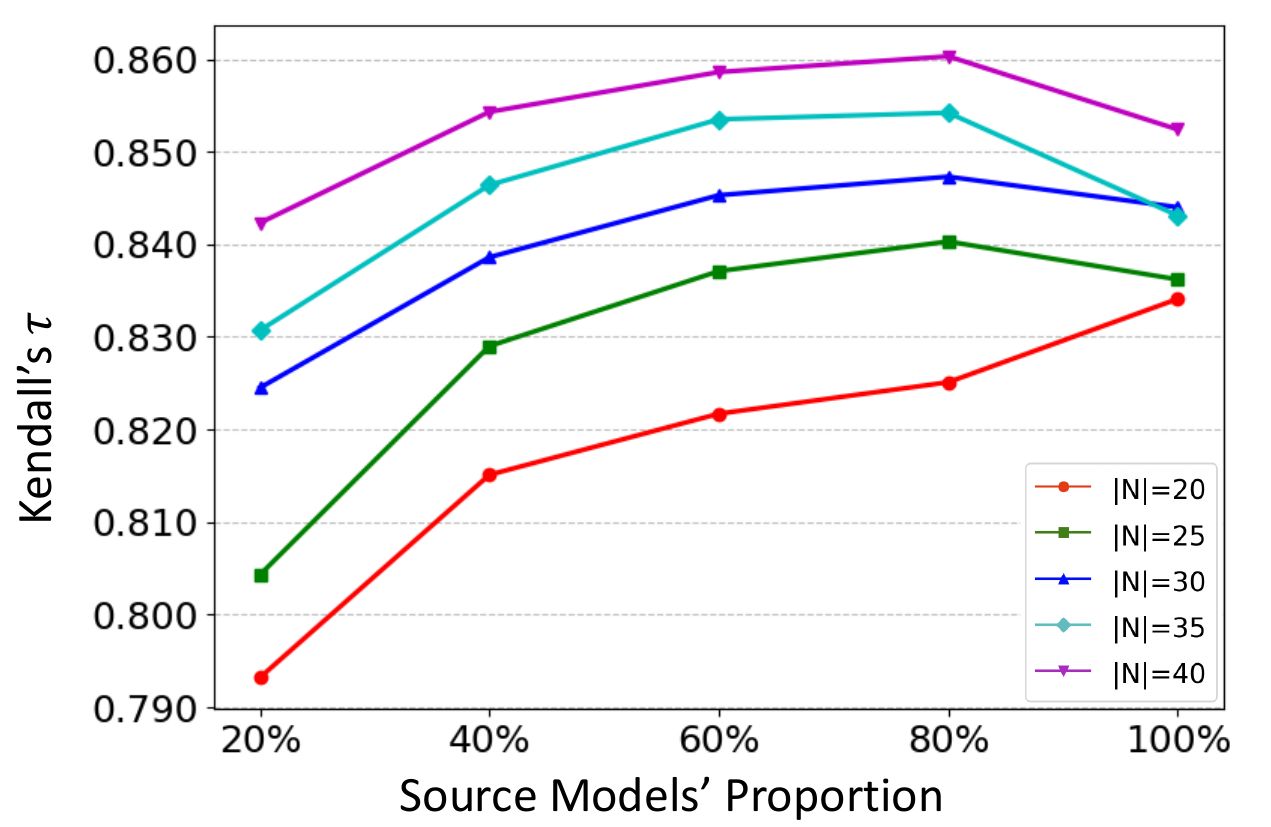}
        \caption{The impact of the quantity of Native Source Models (with prediction consistency kept the same).}
        \label{fig:GSM8K_modelratio_kendall}
    \end{subfigure}
    \hfill 
    \begin{subfigure}[b]{0.49\textwidth} 
        \centering 
        \includegraphics[width=\linewidth]{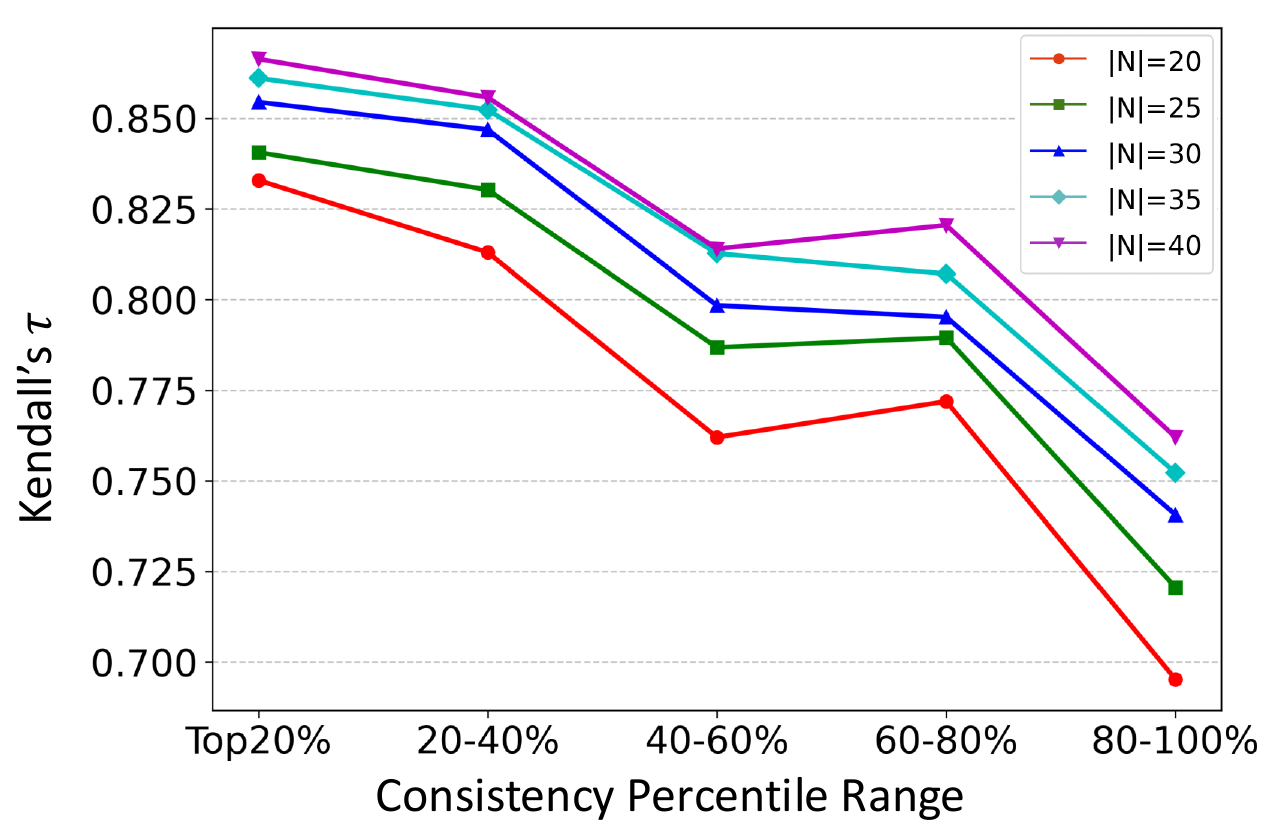}
        \caption{The impact of prediction consistency between the Native Source Model and Target Model (with quantity kept the same).}
        \label{fig:GSM8K_similarity_kendall}
    \end{subfigure}
    \caption{Investigating the impact of native source model quantity and prediction consistency with target model on GSM8K using the controlled variable method.}
    \label{fig:GSM8K_combined}
    \vspace{-0.4cm}
\end{figure*}






\subsection{Main Results}
\paragraph{TailoredBench: Effective Ranking and Estimation of Model Performances} Table~\ref{tab:allresult} present a comprehensive comparison between our \textsc{TailoredBench} method and the best baseline approaches for each metric across all benchmarks. Full results are available in Appendix \ref{apd:all_result}. In our experiments, we allocated 10 examples to the G-set and averaged the outcomes over 100 randomized trials to ensure statistical reliability. The inference count—defined as the number of examples in the N-set for our method—varied from 20 to 40.



As demonstrated in the table, our method consistently outperforms baseline approaches in both Kendall's $\tau$ and MAE metrics across all inference counts and benchmarks featuring different correctness types. When the inference count increases, the performance of our method continues to improve, evidenced by a steady increase in Kendall's $\tau$ and a continuous decrease in MAE. Notably, compared to best performing baselines, our approach achieves nearly a 31.4\% reduction in MAE. These results indicate that our method effectively estimates the relative performance among target models and provides more accurate estimations of their performance on the entire benchmark. Furthermore, compared to the static AnchorPoints method, our approach significantly improves both Kendall's $\tau$ and MAE metrics, highlighting its effectiveness in adaptively selecting a more representative N-set for each target model and thereby improving estimation accuracy. We also calculate the accuracy of our method in ranking the performance between every pair of target models. The results show that the accuracy reached 96.0\% on the Hellaswag benchmark and 93.6\% on the GSM8K benchmark. In terms of robustness, Appendix \ref{apd:variance} demonstrates that our method exhibits significantly lower variance compared to the baselines. 

Moreover, across all benchmarks and inference counts, we conduct a one-sided Z-test over 100 repeated experiments. Whenever our method outperformed the baselines, the p-values remained below 0.05, confirming a statistical advantage.

    

\subsection{Ablation Studies}
\paragraph{Element-Wise Distance Effectively Facilitates Handling Various Data Forms} 
Our method uses element-wise Distance (specifically Manhattan distance) to effectively handle both continuous and discrete values. As shown in Table \ref{tab:distance}, with 30 inference counts, element-wise Distances outperform the correlation distance used by AnchorPoints. This confirms its effectiveness in improving our method’s performance. Detailed per-dataset results are provided in Appendix \ref{apd:distance}.
\begin{table}[htbp]
\renewcommand\arraystretch{1}
\centering
\setlength{\tabcolsep}{1.25em} 
\begin{tabular}{l cc}
\toprule
\textbf{Distance}& \textbf{$\tau \uparrow$} & {\scriptsize \textbf{MAE} $\downarrow$}  \\ 
\midrule
\textsc{Correlation}    & 0.720 & 0.032 \\
\textsc{Cosine}         & \uline{0.736} & \uline{0.028}  \\
\textsc{Manhattan}      & \textbf{0.740} & \textbf{0.027} \\
\bottomrule
\end{tabular}
\caption{Average performance with different types of distance across benchmarks.}
\label{tab:distance}
\vspace{-0.5cm}
\end{table}

\paragraph{Calibrated Estimation Strategy Improves Performance Estimation} 
We compare \textsc{TailoredBench} with and without calibration. As shown in Table \ref{tab:calibrate}, with 30 inference counts, the calibrated variant achieves higher Kendall’s $\tau$ and lower MAE, confirming that calibration enhances the accuracy of performance estimation. Detailed per-dataset results are provided in Appendix \ref{apd:ablation_calibration}.

\paragraph{More Ablation Studies} 
We conduct additional ablation studies on our method in Appendices \ref{apd:dynamic-native-source-models} and \ref{apd:not-fix-gset}. The results show that (1) \textbf{using a fixed number of native source models for each target model stabilizes performance}; (2) \textbf{fixing the G-set as part of the N-set strikes a balance between effectiveness and the inference budget}.

\subsection{Analyses}
\paragraph{Impact of Native Source Model Selection on Method Performance} Here, we isolate the effects of both the number of native source models and their prediction consistency with the target model by independently varying these factors.

\textit{When native source models share a fixed level of prediction consistency with the target model, increasing their number enhances performance.} To investigate this, we randomly select models designated as native source models, from 20\% to 100\%. As shown in Figure \ref{fig:GSM8K_modelratio_kendall}, performance improves as more native source models are included, since a larger set of models offers a greater chance of obtaining a more robust embedding. See Appendix \ref{apd:quantity} for results on more benchmarks.

\textit{When the number of native source models is fixed, higher prediction consistency with the target model enhances performance.} To examine this, we select a fixed number of native source models at various consistency levels relative to the target model (top 20\%, 20\textasciitilde 40\%, up to 80\textasciitilde 100\%). 
\begin{table}[htbp]
\renewcommand\arraystretch{1}
\centering
\setlength{\tabcolsep}{1em} 
\begin{tabular}{l cc}
\toprule
\textbf{Method Variants}& \textbf{$\tau \uparrow$} & {\scriptsize \textbf{MAE} $\downarrow$}  \\ 
\midrule
\textsc{Non-Calibrated}    & 0.724 & 0.030 \\
\textsc{Calibrated}         & \textbf{0.740} & \textbf{0.027}  \\
\bottomrule
\end{tabular}
\caption{Average performance with and without calibration across benchmarks.}
\label{tab:calibrate}
\vspace{-0.5cm}
\end{table}
As shown in Figure \ref{fig:GSM8K_similarity_kendall} (with the horizontal axis representing the Consistency Percentile Range for these intervals), Kendall’s $\tau$ decreases sharply as the consistency percentile range expands. See Appendix \ref{apd:similarity} for additional benchmark results.

\begin{figure*}[t]
  \includegraphics[width=0.49\linewidth]{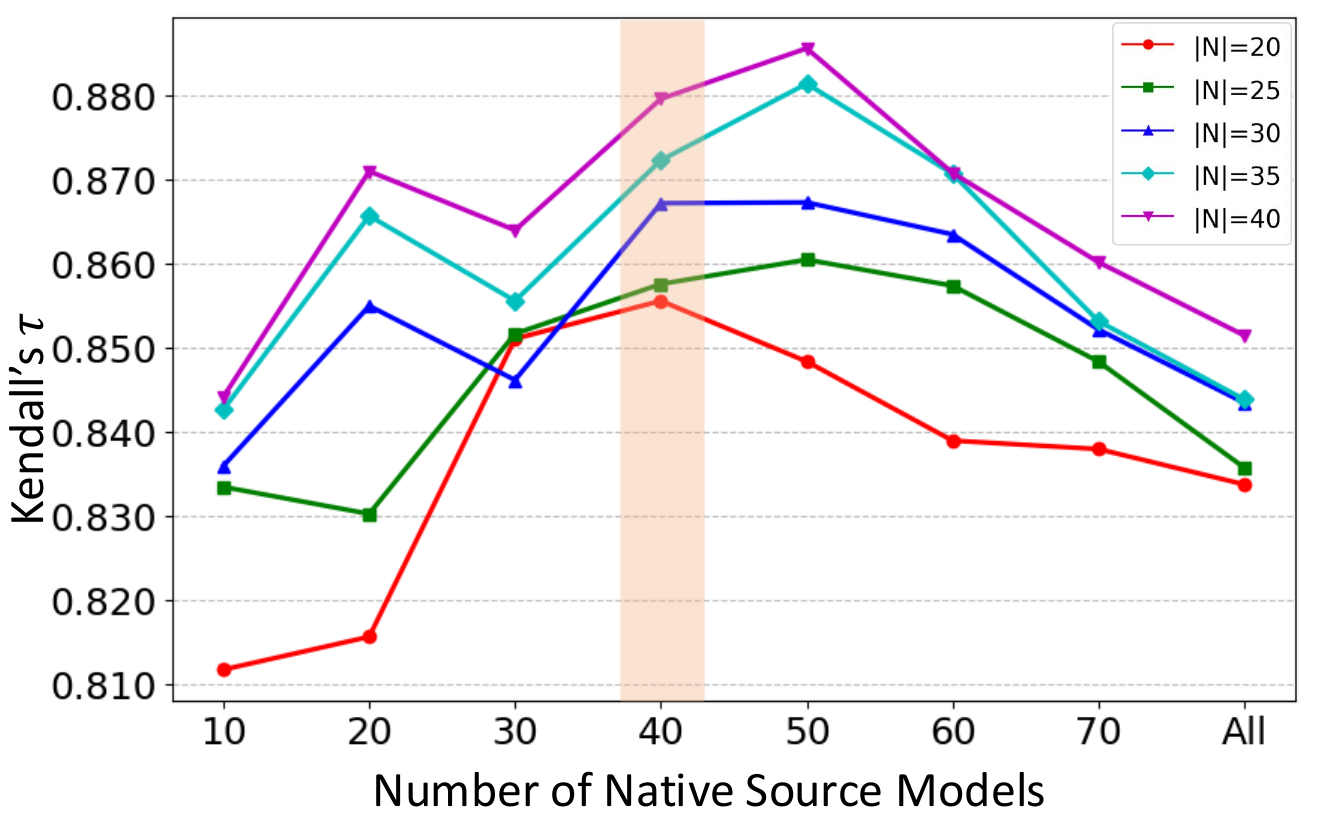} \hfill
  \includegraphics[width=0.49\linewidth]{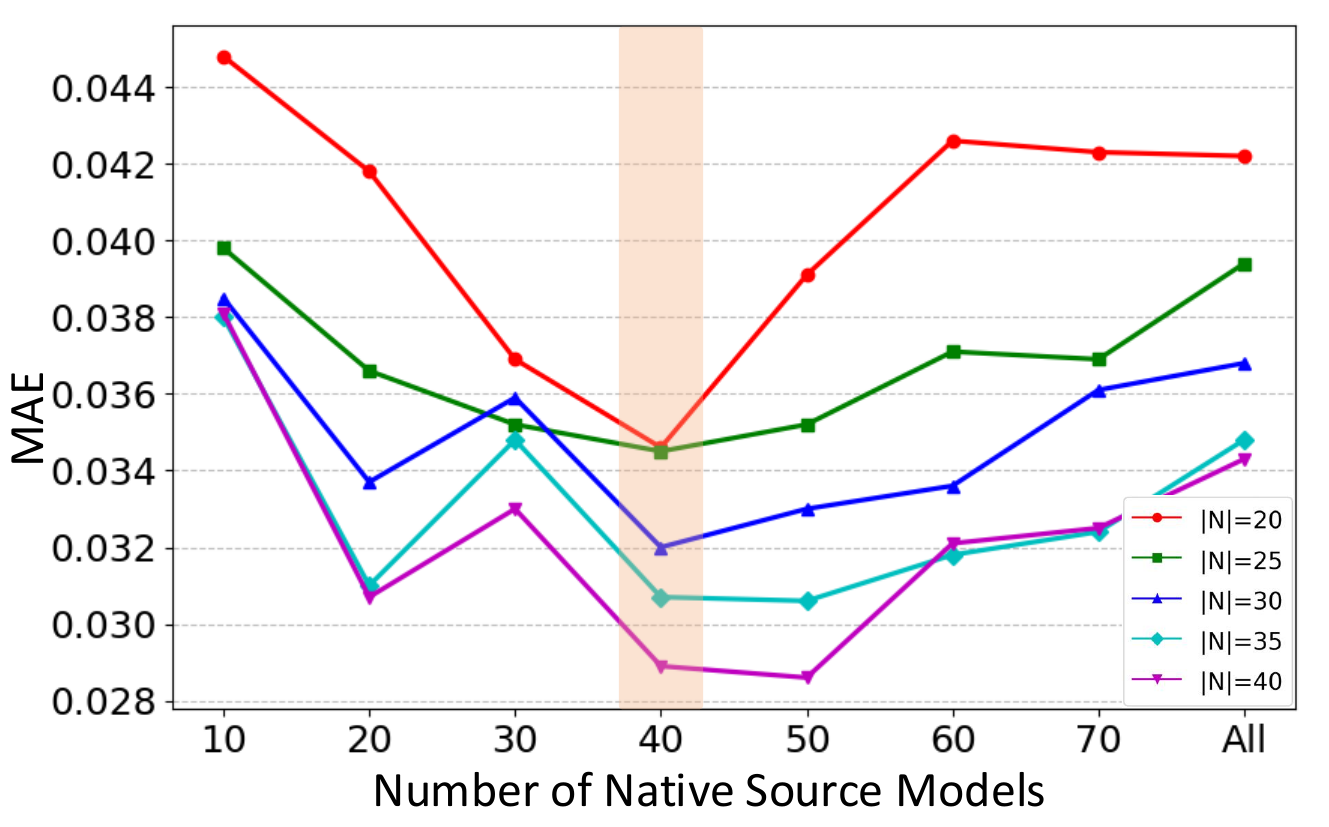}
  \caption {Performance of \textsc{TailoredBench} with varying numbers of Native Source Models on GSM8K benchmark. The shaded area indicates the adaptively selected number of native source models and the corresponding performance of our method.}
\vspace{-0.4cm}
\label{FigMainAnalysis}
\end{figure*}

\paragraph{TailoredBench Method Adaptively Selects Optimal Native Source Model Sets} Here, we analyze the ability of our method to select the optimal native source model sets. Figure \ref{FigMainAnalysis} shows the performance of our method on the GSM8K benchmark, where source models with the top-k prediction consistency to the target model are selected as Native source models. The results reveal that Kendall's $\tau$ coefficient initially increases and then decreases as the number of native source models grows, while the MAE first decreases and then increases. This trend aligns with our observations in Figure \ref{fig:GSM8K_combined}. Specifically, when only a few native source models are selected, their high consistency with the target model is offset by the noise introduced due to the small sample size, which reduces clustering performance. Increasing the number of native source models helps mitigate this issue and improves performance until an optimal point is reached. However, selecting too many native source models incorporates models with lower prediction consistency to the target model, which diminishes effectiveness. Our method addresses this by adaptively selecting the near-optimal number of native source models across all benchmarks. For example, as shown in Figure \ref{FigMainAnalysis}, our approach selects 40 native source models for each target model on the GSM8K benchmark, achieving near-optimal performance. Further experiments pertaining to this section are detailed in Appendix \ref{apd:main_analysis}.

Additionally, we observe that target models preferentially select native source models from their own family, which can better capture the nuances and prediction patterns distinctive to their respective model lineages and contribute to more accurate performance estimations. This intra-family selection bias and the performance of our method when target models significantly differ from source models is further explored in detail in Appendix \ref{apd:intra-family-models}.

\paragraph{10 Examples are Sufficient for the Probe} 
We investigate how G-set size affects our method's performance by fixing the N-set at 30 examples and varying the G-set from 5 to 25 examples across all benchmarks. As shown in Table \ref{tab:gset}, Kendall's $\tau$ peaks and MAE reaches a minimum at a G-set size of 10. Smaller G-set fail to capture the prediction consistency between source and target models, limiting effective N-set selection. Conversely, a larger G-set reduces N-set representativeness by being dominated by G-set points, leading to diminished performance. Detailed per-dataset results are provided in Appendix \ref{apd:gset}.

\begin{table}[htbp]
\renewcommand\arraystretch{1}
\centering
\setlength{\tabcolsep}{1.25em} 
\begin{tabular}{ccc}
\toprule
\textbf{$\lvert$G-set$\rvert$} & \textbf{$\tau \uparrow$} & {\scriptsize \textbf{MAE} $\downarrow$}  \\ 
\midrule
\textsc{5}    & 0.734 & 0.030 \\
\textsc{10}   & \textbf{0.740} & \textbf{0.027} \\
\textsc{15}   & \uline{0.736} & \uline{0.028} \\
\textsc{20}   & 0.735 & \uline{0.028} \\
\textsc{25}   & 0.731 & 0.029 \\
\bottomrule
\end{tabular}
\caption{Average performance with different G-set size across benchmarks.}
\label{tab:gset}
\vspace{-0.5cm}
\end{table}

\paragraph{Performance with Larger Inference Count On the Hellaswag Benchmark.} 
We further evaluate our method with larger inference counts on the Hellaswag benchmark. 
Table \ref{tab:largeNset} shows that as inference counts increase from 50 to 150, \textsc{TailoredBench} consistently improves model performance prediction and ranking, maintaining a clear advantage over baseline methods, demonstrating its effectiveness with larger inference budgets.

\begin{table}[htbp]
\renewcommand\arraystretch{1.41}
\centering
\small
\setlength{\tabcolsep}{0.39em} 
\begin{tabular}{c *{3}{cc}}
\toprule
\multirow{2}{*}{\makecell{\textbf{Inference} \\ \textbf{counts}}} & \multicolumn{2}{c}{50} & \multicolumn{2}{c}{100} & \multicolumn{2}{c}{150} \\  
\noalign{\vskip -0.17em}
 & \textbf{$\tau \uparrow$} & {\scriptsize \textbf{MAE} $\downarrow$}  & \textbf{$\tau \uparrow$} & {\scriptsize \textbf{MAE} $\downarrow$}  & \textbf{$\tau \uparrow$} & {\scriptsize \textbf{MAE} $\downarrow$} \\ 
\midrule 
\textsc{Random}         & 0.887 & 0.053 & 0.920 & 0.038 & 0.935 & \uline{0.030} \\
\hdashline
\makecell{\textsc{Anchor} \\ \textsc{Points}}   & \uline{0.915} & 0.046 & \uline{0.931} & 0.040 & \uline{0.940} & 0.040 \\
\hdashline
\textsc{gp-IRT}         & 0.869 & \uline{0.026} & 0.915 & \uline{0.015} & 0.936 & \textbf{0.012} \\
\hdashline
\makecell{\textsc{Tailored} \\ \textsc{Bench}}  & \textbf{0.923} & \textbf{0.016} & \textbf{0.934} & \textbf{0.014} & \textbf{0.943} & \textbf{0.012} \\
\bottomrule
\end{tabular}
\caption{Performance of compared methods on the Hellaswag benchmark with larger inference counts.}
\label{tab:largeNset}
\vspace{-0.6cm}
\end{table}

%% file: conclusion.tex
\section{Conclusions}
In this paper, we propose the \textsc{TailoredBench} method, which mainly includes an adaptive source model set selection strategy, a scalable K-Medoids clustering algorithm and a calibrated performance estimation strategy. 
Abandoning the one-size-fits-all approach, we have customized the evaluation on the constructed native coreset for each target model.
This approach enables a more accurate reconstruction and ranking of the model's performance on the entire benchmark with a small-size inference budget. 
Comprehensive experiments show that \textsc{TailoredBench} can achieve more accurate model evaluation (an average of 31.4\% estimation MAE loss degradation) with few inference costs.

\section*{Limitations} A primary limitation of mainstream approaches in benchmark compression, including \citep{AP, tiny}, and our method, is their dependence on comprehensive evaluation results from existing models across all examples within a benchmark. As described above, these results are typically readily accessible through public leaderboards. 
However, obtaining initial model performance results is necessary for new or certain private benchmarks, which introduces additional inference overhead. Nonetheless, we maintain that this initial cost is justified, as it is offset by the significant resource savings achieved through numerous subsequent rapid evaluations facilitated by our method.

\section*{Ethics Statement}
All of the datasets used in this study were publicly available, and no annotators were employed for our data collection. We confirm that the datasets we used did not contain any harmful content and was consistent with their intended use (research). We have cited the datasets and relevant works used in this study.

\subsection*{Acknowledgments}
This work is supported by Beijing Natural Science Foundation (No.4222037, L181010).

%% file: appendix.tex
\appendix
\section{Related Works}
\label{apd:related_works}
\paragraph{Models Correlation in Predictive Consistency:} Prior works \citep{taori2020measuring, miller2021accuracy, awadalla2022exploring} have demonstrated a certain level of correlation between in-distribution (ID) and out-of-distribution (OOD) performances across diverse models and tasks. Building on this foundation, \cite{baek2022agreement} and \cite{mehra2024predicting} advance this relationship by showing the phenomenon that the agreement between two models on ID data is linearly correlated with their agreement on OOD data, where the accuracy holds the similar linear relationship, enabling accurate estimation of model's OOD accuracy based solely on ID data. Our work extends this phenomenon to address the challenge of benchmark compression, enabling the selection of more representative subsets for benchmarks. 
\cite{xu2024data} synthesizes several methods and dynamically chooses the optimal subset selection method for each benchmark but requires many examples to determine the best approach. Despite these advancements, these methods often struggle with substantial distribution shifts between the source and target models, caused by the discrepancy between their predictive consistency, potentially causing significant distortion in estimating the target model's performance. 
Extending the approach of \cite{AP}, our work alleviates this issue by dynamically selecting a native source model set with the highest prediction consistency to the target model, ensuring the selection of a tailored coreset for each target model that best represents the benchmark.

\textbf{Scaling Approaches for Model Performance Estimations:} 
Scaling law describes the relationship between model properties (e.g., FLOPs used during training, model parameter size) and model capabilities. 
Recent works \citep{hu2023predicting, ruan2024observational, isik2024scaling} have leveraged scaling laws to predict model performance on various downstream tasks, reducing the computational cost of evaluating models on complex downstream tasks. \cite{zhang2024collaborative} simplifies those approaches by utilizing the relationships between model families and their collaborative overall performance across tasks rather than fitting scaling laws. The aforementioned methods typically rely on overall model performance across several benchmarks and specific design factors (e.g., model size or training data properties) to either fit scaling curves or investigate correlations between models on various tasks. In contrast, our approach addresses a more general case by reducing the evaluation cost for multiple models on a single benchmark, offering a more efficient performance estimation framework.

\section{More Experimental Results}

\subsection{Comprehensive Experimental Results Across All Datasets}
\label{apd:all_result}
In Table \ref{tab:apdallresult}, we present a comprehensive comparison of our approach against all baseline methods across the full range of benchmark datasets. The results indicate that our method consistently outperforms every baseline under all considered inference counts, thereby demonstrating the overall effectiveness of our proposed approach.
\begin{table*}[ht]
\renewcommand\arraystretch{1.2}
\centering
\small
\setlength{\tabcolsep}{0.49em} 
\begin{tabular}{cl*{5}{cc}}
\toprule
\multirow{2}{*}{\textbf{Benchmarks}} & \multirow{2}{*}{\textbf{Inference counts}} & \multicolumn{2}{c}{20} & \multicolumn{2}{c}{25} & \multicolumn{2}{c}{30} & \multicolumn{2}{c}{35} & \multicolumn{2}{c}{40} \\
&  & \textbf{$\tau \uparrow$} & {\scriptsize \textbf{MAE} $\downarrow$}  & \textbf{$\tau \uparrow$} & {\scriptsize \textbf{MAE} $\downarrow$}  & \textbf{$\tau \uparrow$} & {\scriptsize \textbf{MAE} $\downarrow$}  & \textbf{$\tau \uparrow$} & {\scriptsize \textbf{MAE} $\downarrow$}  & \textbf{$\tau \uparrow$} & {\scriptsize \textbf{MAE} $\downarrow$}  \\
\midrule
\multirow{4}{*}{ARC Challenge}&\textsc{Random}        & 0.626 & 0.078 & 0.659 & 0.065 & 0.676 & 0.067 & 0.694 & 0.062 & 0.712 & 0.057 \\
&\textsc{AnchorPoints} & \uline{0.662} & 0.064 & \uline{0.663} & 0.058 & \uline{0.676} & 0.053 & \uline{0.713} & 0.048 & \uline{0.714} & 0.043 \\
&\textsc{gp-IRT}        & 0.589 & \uline{0.046} & 0.620 & \uline{0.046} & 0.662 & \uline{0.036} & 0.681 & \uline{0.036} & 0.695 & \uline{0.029} \\ 
                           & \textsc{TailoredBench} & \textbf{0.711} & \textbf{0.031} & \textbf{0.737} & \textbf{0.029} & \textbf{0.756} & \textbf{0.028} & \textbf{0.766} & \textbf{0.027} & \textbf{0.773} & \textbf{0.027} \\
\hdashline
\multirow{4}{*}{Hellaswag}&\textsc{Random}         & 0.811 & 0.083 & 0.836 & 0.077 & 0.850 & 0.066 & 0.863 & 0.060 & 0.871 & 0.058  \\
&\textsc{AnchorPoints}  & \uline{0.860} & \uline{0.060} & \uline{0.880} & 0.061 & \uline{0.877} & 0.067 & \uline{0.897} & 0.059 & \uline{0.898} & 0.057 \\
&\textsc{gp-IRT}         & 0.724 & 0.062 & 0.776 & \uline{0.053} & 0.810 & \uline{0.043} & 0.827 & \uline{0.038} & 0.849 & \uline{0.032} \\ 
                           & \textsc{TailoredBench} & \textbf{0.900} & \textbf{0.020} & \textbf{0.909} & \textbf{0.018} & \textbf{0.913} & \textbf{0.018} & \textbf{0.914} & \textbf{0.017} & \textbf{0.918} & \textbf{0.017} \\
\hdashline
\multirow{4}{*}{GSM8K}&\textsc{Random}        & \uline{0.811} & 0.062 & \uline{0.828} & 0.055 & \uline{0.839} & 0.052 & \uline{0.847} & 0.049 & \uline{0.858} & 0.044 \\
&\textsc{AnchorPoints} & 0.786 & 0.087 & 0.791 & 0.079 & 0.796 & 0.073 & 0.800 & 0.071 & 0.799 & 0.071 \\
&\textsc{gp-IRT}        & 0.787 & \uline{0.055} & 0.807 & \uline{0.047} & 0.829 & \uline{0.041} & 0.842 & \uline{0.038} & 0.858 & \uline{0.034} \\ 
                           &\textsc{TailoredBench} & \textbf{0.852} & \textbf{0.035} & \textbf{0.858} & \textbf{0.034} & \textbf{0.863} & \textbf{0.033} & \textbf{0.869} & \textbf{0.031} & \textbf{0.878} & \textbf{0.029} \\
\hdashline
\multirow{4}{*}{Winogrande}&\textsc{Random}        & 0.373 & 0.078 & 0.408 & 0.067 & 0.446 & 0.062 & 0.470 & 0.055 & 0.492 & 0.052 \\
&\textsc{AnchorPoints} & \uline{0.472} & 0.086 & \uline{0.487} & 0.085 & \uline{0.514} & 0.075 & \uline{0.521} & 0.087 & \uline{0.518} & 0.073 \\
&\textsc{gp-IRT}        & 0.263 & \uline{0.041} & 0.313 & \uline{0.038} & 0.353 & \uline{0.038} & 0.392 & \uline{0.036} & 0.419 & \uline{0.034} \\ 
                           & \textsc{TailoredBench} & \textbf{0.565} & \textbf{0.028} & \textbf{0.568} & \textbf{0.026} & \textbf{0.604} & \textbf{0.024} & \textbf{0.608} & \textbf{0.023} & \textbf{0.618} & \textbf{0.022} \\
\hdashline
\multirow{4}{*}{POPE}&\textsc{Random}        & \uline{0.488} & 0.058 & \uline{0.510} & 0.054 & 0.507 & 0.048 & 0.515 & 0.044 & 0.547 & 0.040 \\
&\textsc{AnchorPoints} & 0.474 & 0.040 & 0.483 & 0.038 & \uline{0.518} & \uline{0.034} & \uline{0.547} & \uline{0.033} & \uline{0.556} & \textbf{0.031} \\
&\textsc{gp-IRT}        & 0.481 & \uline{0.038} & 0.470 & \uline{0.037} & 0.462 & 0.036 & 0.482 & 0.034 & 0.477 & 0.033 \\ 
                           &\textsc{TailoredBench} & \textbf{0.521} & \textbf{0.036} & \textbf{0.547} & \textbf{0.035} & \textbf{0.562} & \textbf{0.031} & \textbf{0.563} & \textbf{0.031} & \textbf{0.574} & \uline{0.032} \\
\bottomrule
\end{tabular}
\caption{Results on all benchmarks. Values in bold represent the best results, while values that are underlined represent the second-best results.}
\label{tab:apdallresult}
\end{table*}
\subsection{Comprehensive Distance Measures Ablation Across Benchmarks}
\label{apd:distance}
Here, we provide comprehensive results of our ablation study evaluating the impact of different distance measures on our method's performance with 30 inference counts across various benchmarks. Table \ref{apdtab:distance} presents detailed Kendall's $\tau$ and MAE metrics for cosine similarity, Manhattan distance, and correlation distance across all datasets. These results offer deeper insights into the effectiveness of Element-Wise Distance measures in enhancing benchmark compression.

\begin{table*}
\renewcommand\arraystretch{1}
\centering
\setlength{\tabcolsep}{0.3em} 
\begin{tabular}{l *{5}{cc}}
\toprule
\multirow{2}{*}{\textbf{Distances}} & \multicolumn{2}{c}{ARC Challenge} & \multicolumn{2}{c}{Hellaswag} & \multicolumn{2}{c}{GSM8K} & \multicolumn{2}{c}{Winogrande} & \multicolumn{2}{c}{POPE} \\ 
\noalign{\vskip -0.17em}
 & \textbf{$\tau \uparrow$} & {\scriptsize \textbf{MAE} $\downarrow$}  & \textbf{$\tau \uparrow$} & {\scriptsize \textbf{MAE} $\downarrow$}  & \textbf{$\tau \uparrow$} & {\scriptsize \textbf{MAE} $\downarrow$}  & \textbf{$\tau \uparrow$} & {\scriptsize \textbf{MAE} $\downarrow$}  & \textbf{$\tau \uparrow$} & {\scriptsize \textbf{MAE} $\downarrow$}  \\ 
\midrule
\textsc{Correlation}    & \textbf{0.766} & 0.033 & 0.903 & \uline{0.019} & \uline{0.828} & 0.041 & 0.557 & 0.029 & 0.547 & 0.038 \\
\textsc{Cosine}         & 0.746 & \uline{0.031} & \textbf{0.914} & \uline{0.019} & 0.827 & \uline{0.040} & \textbf{0.616} & \textbf{0.024} & \textbf{0.577} & \textbf{0.024} \\
\textsc{Manhattan}      & \uline{0.756} & \textbf{0.028} & \uline{0.913} & \textbf{0.018} & \textbf{0.863} & \textbf{0.033} & \uline{0.604} & \uline{0.024} & \uline{0.562} & \uline{0.031} \\
\bottomrule
\end{tabular}
\caption{Detailed ablation results for distance selection across all benchmarks.}
\label{apdtab:distance}
\end{table*}

\subsection{Detailed Calibration Ablation Results}
\label{apd:ablation_calibration}
Table \ref{tab:ablation_calibration} presents the results of our ablation study, comparing our \textsc{TailoredBench} method with and without the calibrated performance estimation process under 30 inference counts. The calibrated version of our method generally achieves higher Kendall’s $\tau$ scores and lower mean absolute errors (MAE) across various benchmarks and inference counts, demonstrating that the calibrated performance estimation process effectively enhances the performance estimation ability of our method.

\begin{table*}
\renewcommand\arraystretch{1}
\centering
\setlength{\tabcolsep}{0.3em} 
\begin{tabular}{l *{5}{cc}}
\toprule
\multirow{2}{*}{\textbf{Method Variants}} & \multicolumn{2}{c}{ARC Challenge} & \multicolumn{2}{c}{Hellaswag} & \multicolumn{2}{c}{GSM8K} & \multicolumn{2}{c}{Winogrande} & \multicolumn{2}{c}{POPE} \\ 
\noalign{\vskip -0.17em}
 & \textbf{$\tau \uparrow$} & {\scriptsize \textbf{MAE} $\downarrow$}  & \textbf{$\tau \uparrow$} & {\scriptsize \textbf{MAE} $\downarrow$}  & \textbf{$\tau \uparrow$} & {\scriptsize \textbf{MAE} $\downarrow$}  & \textbf{$\tau \uparrow$} & {\scriptsize \textbf{MAE} $\downarrow$}  & \textbf{$\tau \uparrow$} & {\scriptsize \textbf{MAE} $\downarrow$}  \\ 
\midrule
\textsc{Non-Calibrated}    & 0.748 & \textbf{0.026} & 0.910 & \textbf{0.017} & 0.862 & 0.036 & 0.588 & 0.028 & 0.531 & 0.043 \\
\textsc{Calibrated}      & \textbf{0.756} & 0.028 & \textbf{0.913} & 0.018 & \textbf{0.863} & \textbf{0.033} & \textbf{0.604} & \textbf{0.024} & \textbf{0.562} & \textbf{0.031} \\
\bottomrule
\end{tabular}
\caption{Detailed ablation results for calibrated performance estimation process across all benchmarks.}
\label{tab:ablation_calibration}
\vspace{-0.3cm}
\end{table*}

\subsection{Impact of Dynamic Native Source Model Quantity on Performance}
\label{apd:dynamic-native-source-models}
In this ablation study, we investigate the effect of dynamically selecting varying numbers of native source models for each target model, as opposed to using a standardized quantity across all target models. Specifically, instead of treating \( \bar{n} \) (as computed in Eq. \ref{eq:bar_n}) as the fixed number of native source models, we now interpret it as a lower bound—thereby including all source models whose prediction consistency exceeds the threshold.

Table \ref{tab:dynamic-native-source-models} summarizes the results for all benchmarks under an inference count of 30. Notably, we observe improvements on the GSM8K and POPE datasets, while a slight decrease in performance is seen on other datasets.

These findings underscore that the core strength of our method lies in maximizing the consistency between the source and target models. When exactly \( \bar{n} \) native source models are selected for each target model, performance appears to be near optimal. In contrast, adding additional native source models for certain target models may introduce performance variability. Consequently, we adopt a standardized number of native source models to ensure stability.
\begin{table*}
\renewcommand\arraystretch{1}
\centering
\setlength{\tabcolsep}{0.3em} 
\begin{tabular}{l *{5}{cc}}
\toprule
\multirow{2}{*}{\textbf{Method Variants}} & \multicolumn{2}{c}{ARC Challenge} & \multicolumn{2}{c}{Hellaswag} & \multicolumn{2}{c}{GSM8K} & \multicolumn{2}{c}{Winogrande} & \multicolumn{2}{c}{POPE} \\ 
\noalign{\vskip -0.17em}
 & \textbf{$\tau \uparrow$} & {\scriptsize \textbf{MAE} $\downarrow$}  & \textbf{$\tau \uparrow$} & {\scriptsize \textbf{MAE} $\downarrow$}  & \textbf{$\tau \uparrow$} & {\scriptsize \textbf{MAE} $\downarrow$}  & \textbf{$\tau \uparrow$} & {\scriptsize \textbf{MAE} $\downarrow$}  & \textbf{$\tau \uparrow$} & {\scriptsize \textbf{MAE} $\downarrow$}  \\ 
\midrule
\makecell{\textsc{Dynamic native} \\ \textsc{source models number}} & 0.741 & 0.029 & 0.909 & \textbf{0.018} & \textbf{0.875} & \textbf{0.031} & 0.592 & 0.025 & \textbf{0.604} & \textbf{0.031} \\
\makecell{\textsc{Standardized native} \\ \textsc{source models number}} & \textbf{0.756} & \textbf{0.028} & \textbf{0.913} & \textbf{0.018} & 0.863 & 0.033 & \textbf{0.604} & \textbf{0.024} & 0.562 & \textbf{0.031} \\ 
\bottomrule
\end{tabular}
\caption{Ablation Study on Dynamic vs. Standardized Native Source Model Selection.}
\label{tab:dynamic-native-source-models}
\end{table*}

\subsection{Impact of Fixed Medoids in N-set Construction on Performance}
\label{apd:not-fix-gset}
In the Developing N-set module, our Scalable K-Medoids Clustering algorithm employs the G-set examples as fixed (anchored) medoids. To assess the effectiveness of this design choice, we conducted an ablation study comparing our standard approach (with fixed G-set medoids) against a variant where the G-set points are allowed to change during medoid refinement. The results are summarized in Table \ref{tab:not-fix-gset}.

When the N-set size is fixed at 30, our method with fixed medoids shows slightly inferior performance compared to the variant without fixed medoids (see rows 1 and 3 in Table \ref{tab:not-fix-gset}). When the inference budget is fixed at 30, our method outperforms the variant without fixed medoids (see rows 1 and 2 in Table \ref{tab:not-fix-gset}). These findings suggest that anchoring the G-set as fixed medoids helps achieve a balanced trade-off between the size of the N-set and the available inference budget.

\begin{table*}[ht]
\renewcommand\arraystretch{1}
\centering
\begin{tabular}{lccccc}
\toprule
\multirow{2}{*}{\textbf{Method Variants}} & \multirow{2}{*}{$|\text{G-set}|$} & \multirow{2}{*}{$|\text{N-set}|$} & \multirow{2}{*}{Inference counts} & \multicolumn{2}{c}{Average} \\
&  &  &  & \textbf{$\tau \uparrow$} & {\scriptsize \textbf{MAE} $\downarrow$} \\
\midrule
\textsc{Fixed G-set}           & 10  & 30  & 30  &  0.740  & 0.027 \\
\textsc{not-Fixed G-set}      & 10  & 20  & 30  &  0.719  & 0.031 \\
\textsc{not-Fixed G-set}       & 10  & 30  & 40  &  \textbf{0.745} & \textbf{0.027} \\
\bottomrule
\end{tabular}
\caption{Performance Comparison of N-set Construction Methods with and without Fixed G-set Medoids.}
\label{tab:not-fix-gset}
\end{table*}

\subsection{Intra-Family Affinity and Its Impact on Performance under Different Source-Target Model Similarity}
\label{apd:intra-family-models}
We conducted an additional analysis on the GSM8K dataset to investigate whether models within the same family (e.g., Llama, Mistral) tend to select their own family members as native source models. As shown in Table \ref{tab:native_model_selection}, with a similar number of models from each family within the source and target model set, the results indicate a significant intra-family preference. On average, each llama-series model selected approximately 5.5 Mistral models and 6.4 Llama models as their native source models. Similarly, each Mistral-series model chose about 7.6 Mistral models and 3.0 Llama models on average. These findings suggest that models exhibit a bias toward source models with similar architectures, potentially due to shared representation spaces or analogous decision boundaries. This intra-family affinity may facilitate more accurate performance estimation, as the selected native source models can better capture the nuances and prediction patterns distinctive to their respective model lineages.

Furthermore, we conduct experiments on the GSM8K dataset to evaluate the performance of our method when the target models differ significantly from the source models. Specifically, by selecting only Llama series models as the target and using an inference count of 30, we compare the performance of all methods across two sets of source models: one that includes Llama series models and one that does not, with each set comprising an equal number of models. As shown in Table \ref{tab:Source_Models_Type}, the performance of all methods is closely correlated with the similarity between the target and source models; when these models differ, performance declines across all methods. Nonetheless, our method consistently outperforms the baselines regardless of the target–source model similarity, underscoring the generalizability of our approach.

\begin{table}[htbp]
    \renewcommand\arraystretch{1}
    \centering
    \setlength{\tabcolsep}{0.65em} 
    \begin{tabular}{ccc}
    \toprule
    \makecell{\textbf{Model} \\ \textbf{Family}} & 
    \makecell{{Avg. Selected} \\ {Mistral Models}} & 
    \makecell{{Avg. Selected} \\ {Llama Models}} \\
    \midrule
    \textsc{Llama}    & 5.5 & 6.4 \\
    \textsc{Mistral}  & 7.6 & 3.0 \\
    \bottomrule
    \end{tabular}
    \caption{Statistics of native source model selection within model families on GSM8K benchmark.}
    \label{tab:native_model_selection}
    \vspace{-0.1cm}
\end{table}

\begin{table}
    \renewcommand\arraystretch{1}
    \centering
    \setlength{\tabcolsep}{0.3em} 
    \begin{tabular}{l *{2}{cc}}
    \toprule
    \multirow{2}{*}{\makecell{\textbf{Source Models} \\ \textbf{Composition}}} & \multicolumn{2}{c}{w/o Llama} & \multicolumn{2}{c}{with Llama}\\ 
    \noalign{\vskip -0.17em}
     & \textbf{$\tau \uparrow$} & {\scriptsize \textbf{MAE} $\downarrow$}  & \textbf{$\tau \uparrow$} & {\scriptsize \textbf{MAE} $\downarrow$} \\ 
    \midrule
    \textsc{AnchorPoints}   & 0.388 & 0.048 & 0.525 & 0.050\\
    \textsc{GP-IRT}         & 0.505 & 0.064 & 0.526 & 0.038\\
    \textsc{TailoredBench}  & \textbf{0.634} & \textbf{0.031} & \textbf{0.704} & \textbf{0.022}\\
    \bottomrule
    \end{tabular}
    \caption{Methods' Performance under Different Source-Target Model Similarity.}
    \label{tab:Source_Models_Type}
\end{table}


\subsection{Comprehensive G-set Size Evaluation Across Benchmarks}
\label{apd:gset}
In this section, we present a comprehensive evaluation of how varying G-set sizes affect our method’s performance across multiple benchmarks. Table \ref{apdtab:gset} reports Kendall's $\tau$ and MAE metrics for G-set sizes ranging from 5 to 25 for each benchmark while fixing the N-set size as 30. These results provide deeper insights into selecting the optimal G-set size and support the conclusions drawn in the main text.

\begin{table*}
\renewcommand\arraystretch{1}
\centering
\setlength{\tabcolsep}{0.5em} 
\begin{tabular}{c *{5}{cc}}
\toprule
\multirow{2}{*}{\textbf{ $\lvert$G-set$\rvert$ }} & \multicolumn{2}{c}{ARC Challenge} & \multicolumn{2}{c}{Hellaswag} & \multicolumn{2}{c}{GSM8K} & \multicolumn{2}{c}{Winogrande} & \multicolumn{2}{c}{POPE} \\ 
\noalign{\vskip -0.17em}
 & \textbf{$\tau \uparrow$} & {\scriptsize \textbf{MAE} $\downarrow$}  & \textbf{$\tau \uparrow$} & {\scriptsize \textbf{MAE} $\downarrow$}  & \textbf{$\tau \uparrow$} & {\scriptsize \textbf{MAE} $\downarrow$}  & \textbf{$\tau \uparrow$} & {\scriptsize \textbf{MAE} $\downarrow$}  & \textbf{$\tau \uparrow$} & {\scriptsize \textbf{MAE} $\downarrow$}  \\ 
\midrule
\textsc{5}  & 0.719 & 0.031 & \uline{0.912} & \uline{0.019} & \textbf{0.865} & 0.035 & \uline{0.624} & 0.026 & 0.549 & 0.037 \\
\textsc{10} & \textbf{0.756} & \textbf{0.028} & \textbf{0.913} & \textbf{0.018} & \uline{0.863} & \textbf{0.033} & 0.604 & \textbf{0.024} & \textbf{0.562} & \textbf{0.031} \\
\textsc{15} & \uline{0.751} & \uline{0.029} & 0.911 & \textbf{0.018} & 0.854 & \uline{0.034} & 0.608 & \uline{0.025} & \uline{0.556} & \uline{0.033} \\
\textsc{20} & 0.740 & \uline{0.029} & 0.910 & \textbf{0.018} & 0.862 & \uline{0.034} & 0.621 & 0.026 & 0.541 & 0.034 \\
\textsc{25} & 0.725 & 0.030 & 0.909 & \uline{0.019} & 0.851 & 0.036 & \textbf{0.638} & 0.026 & 0.533 & 0.036 \\
\bottomrule
\end{tabular}
\caption{Detailed results for G-set size across all benchmarks.}
\label{apdtab:gset}
\end{table*}


\subsection{Demonstration of Method Effectiveness with Variance}
\label{apd:variance}
\begin{figure*}[ht]
  \includegraphics[width=0.49\linewidth]{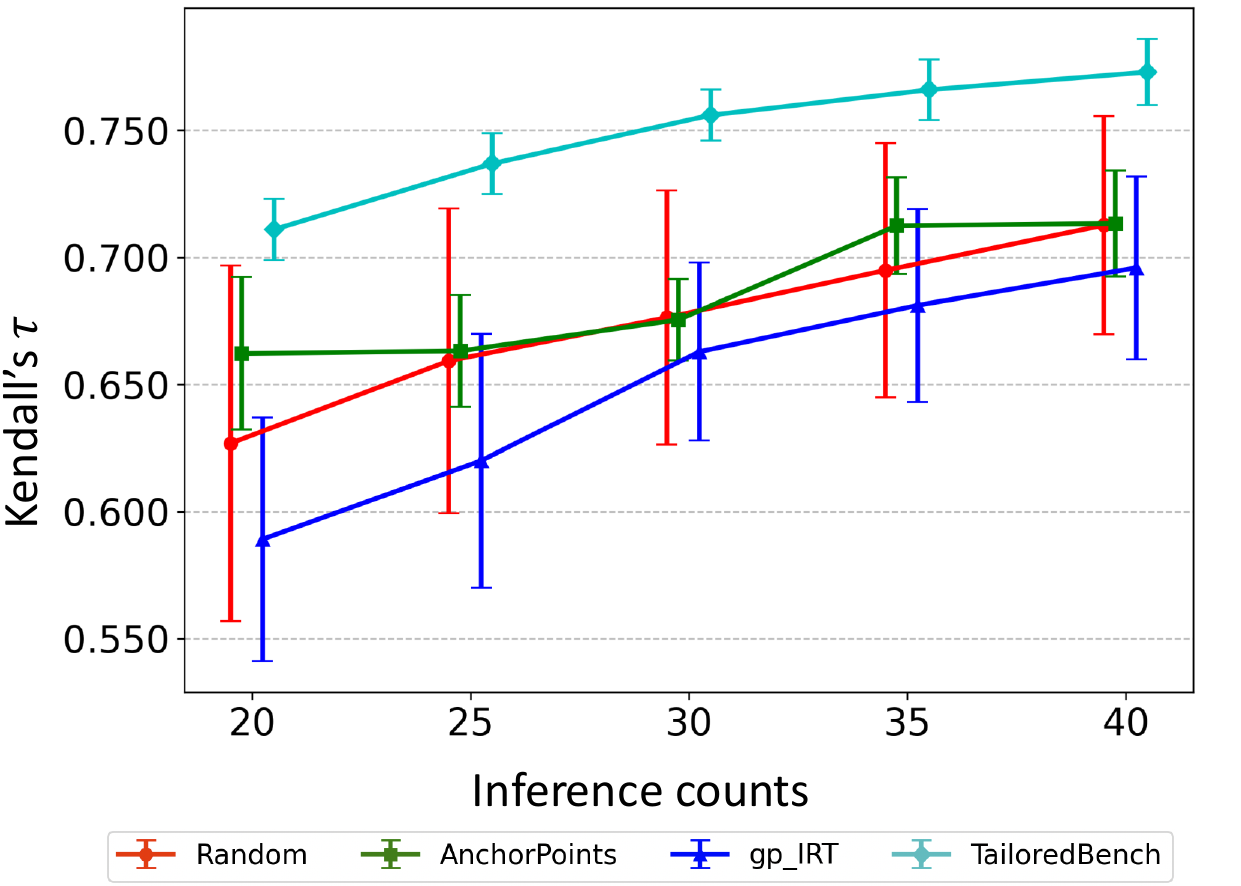} \hfill
  \includegraphics[width=0.49\linewidth]{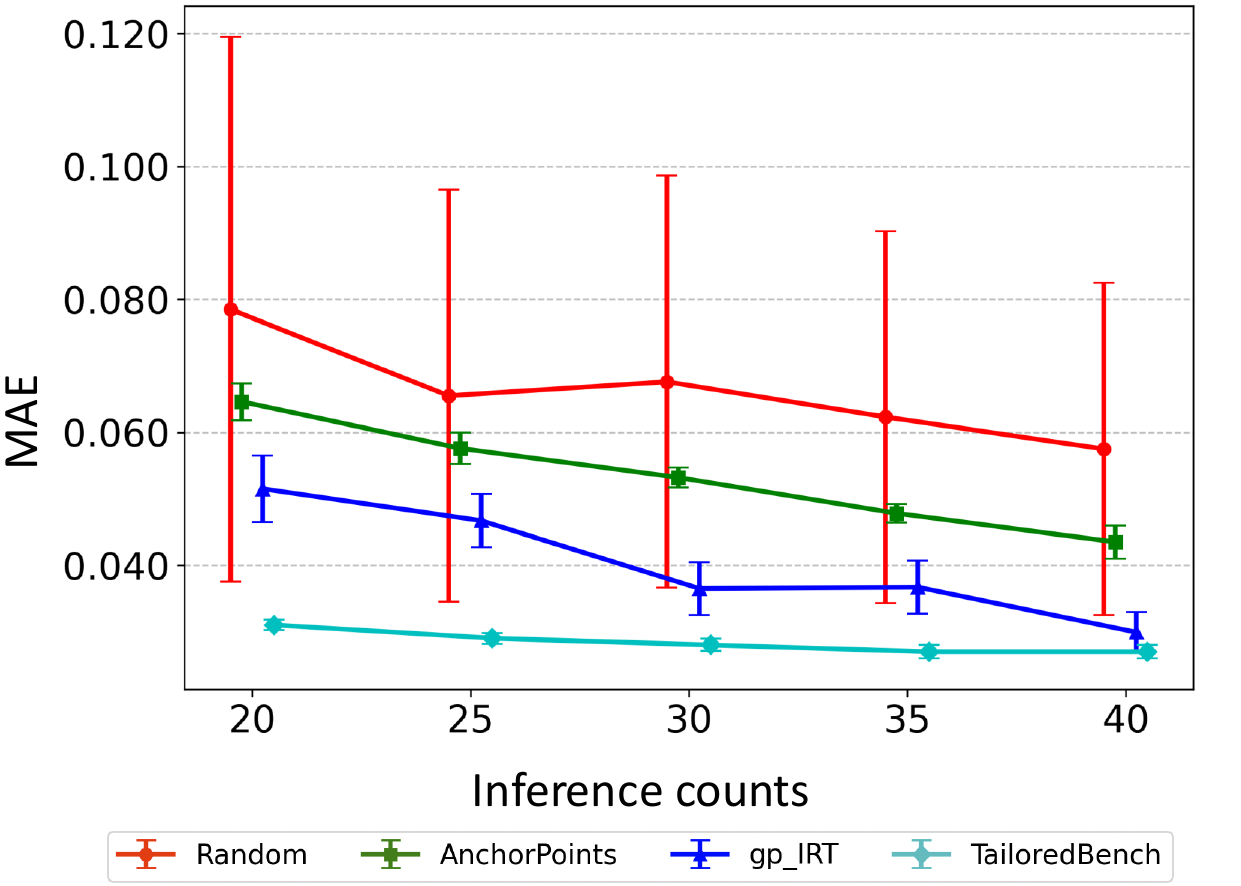}
  \caption {Demonstration of method effectiveness with variance on ARC Challenge benchmark.}
\vspace{-0.3cm}
\label{Variance_arc}
\end{figure*}

\begin{figure*}[ht]
  \includegraphics[width=0.49\linewidth]{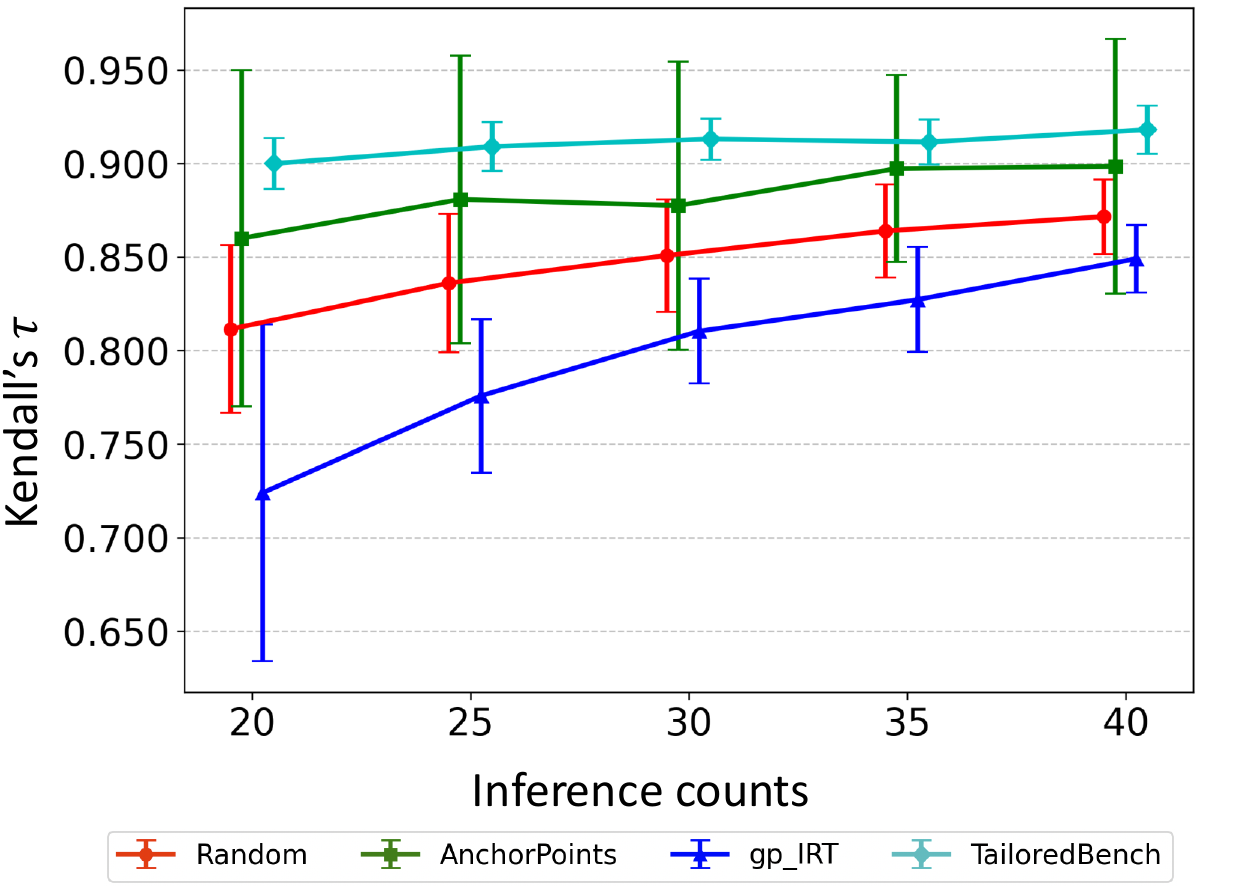} \hfill
  \includegraphics[width=0.49\linewidth]{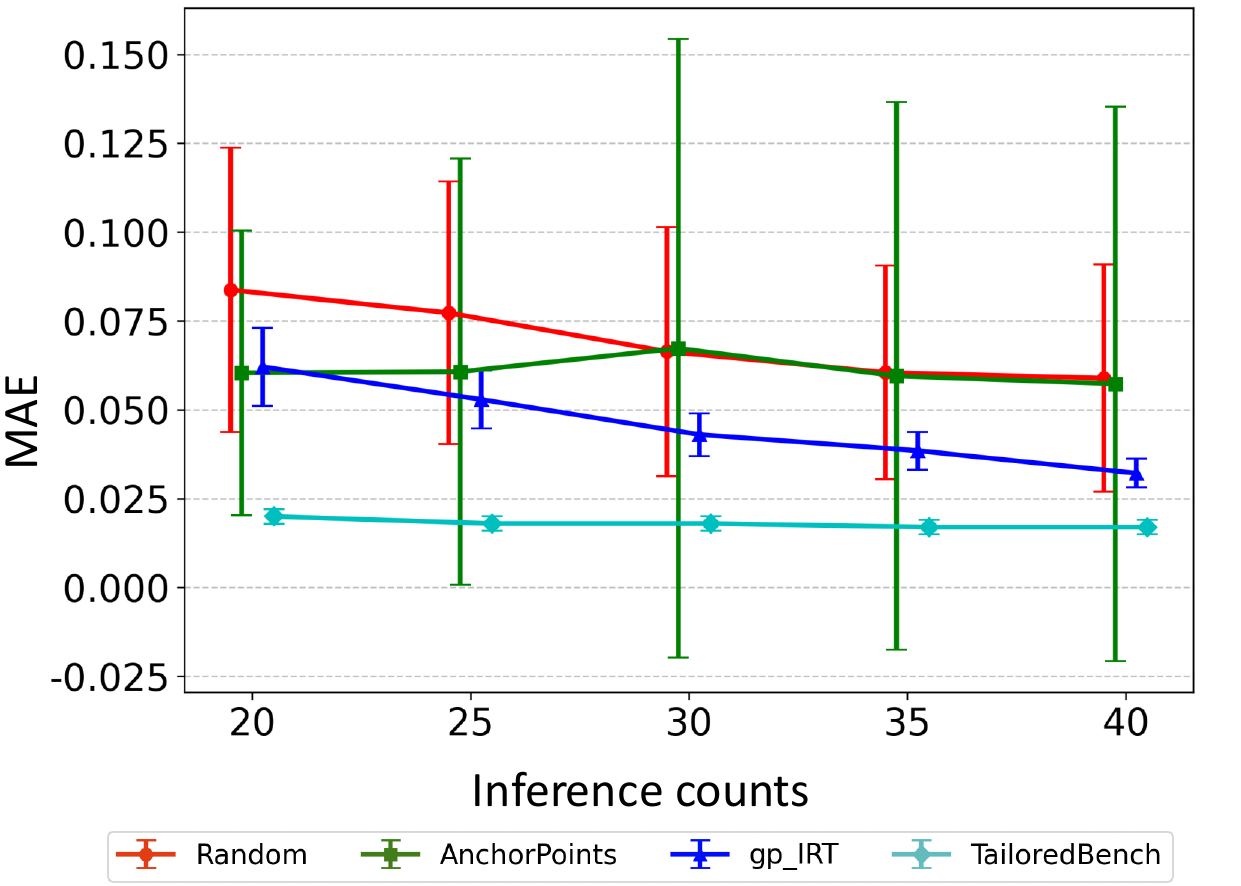}
  \caption {Demonstration of method effectiveness with variance on Hellaswag benchmark.}
\vspace{-0.3cm}
\label{Variance_hella}
\end{figure*}

\begin{figure*}[ht]
  \includegraphics[width=0.49\linewidth]{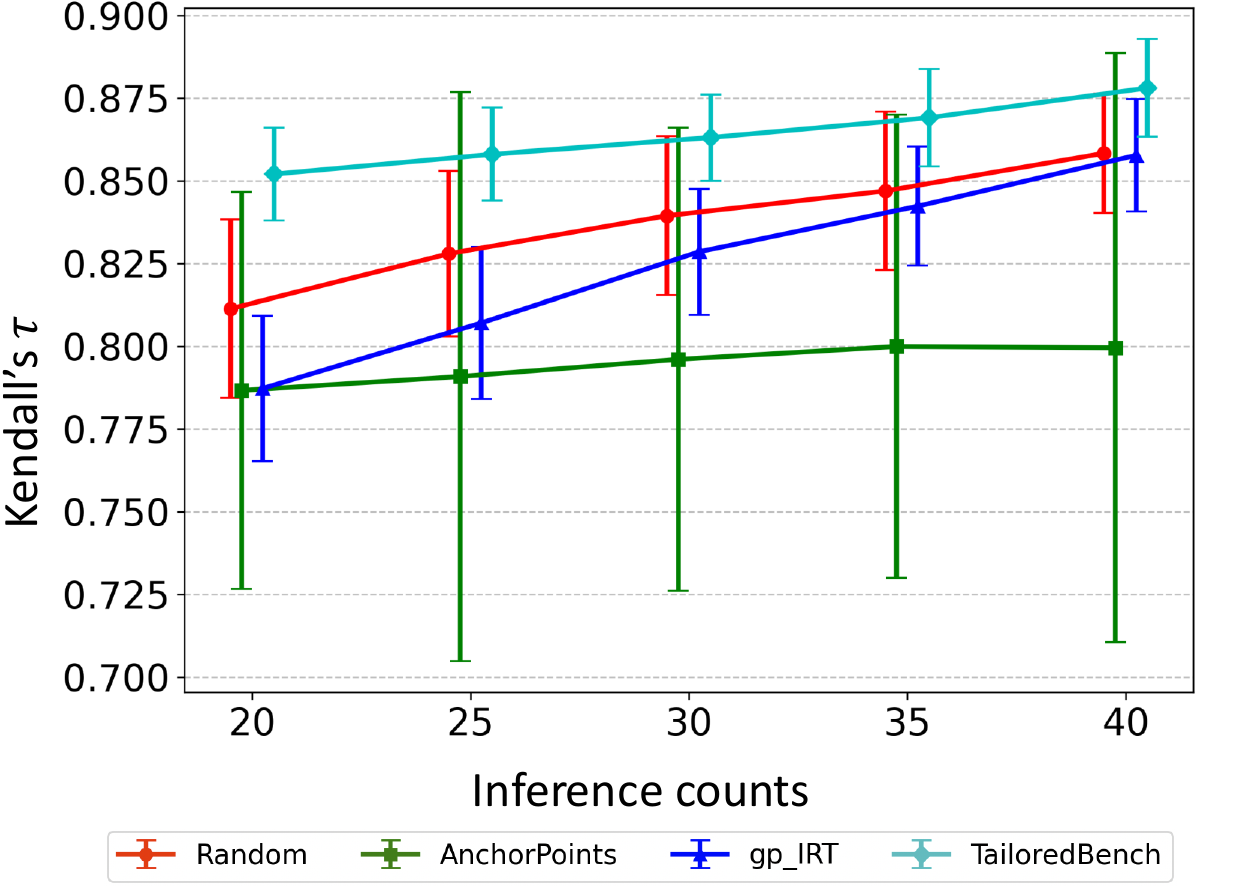} \hfill
  \includegraphics[width=0.49\linewidth]{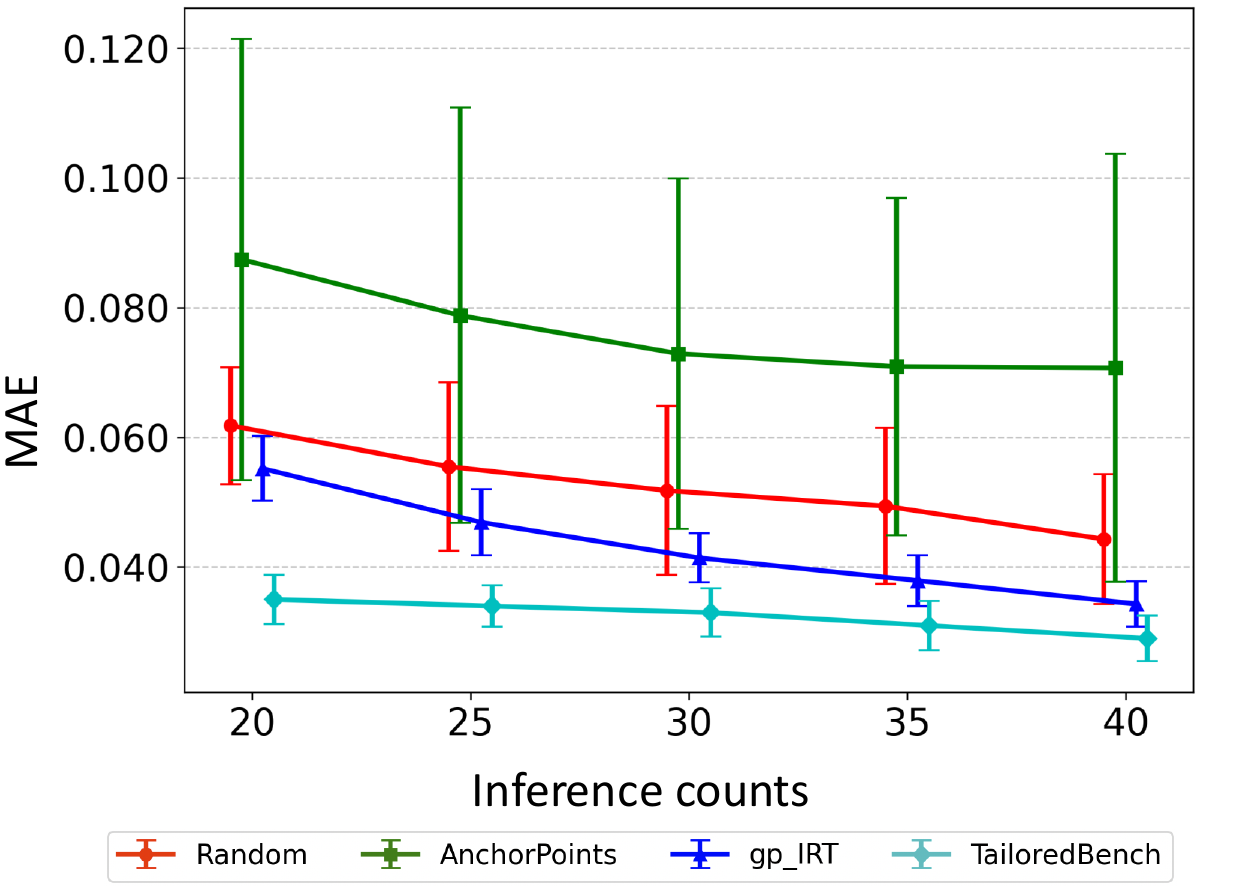}
  \caption {Demonstration of method effectiveness with variance on GSM8K benchmark.}
\vspace{-0.3cm}
\label{Variance_GSM8K}
\end{figure*}

\begin{figure*}[ht]
  \includegraphics[width=0.49\linewidth]{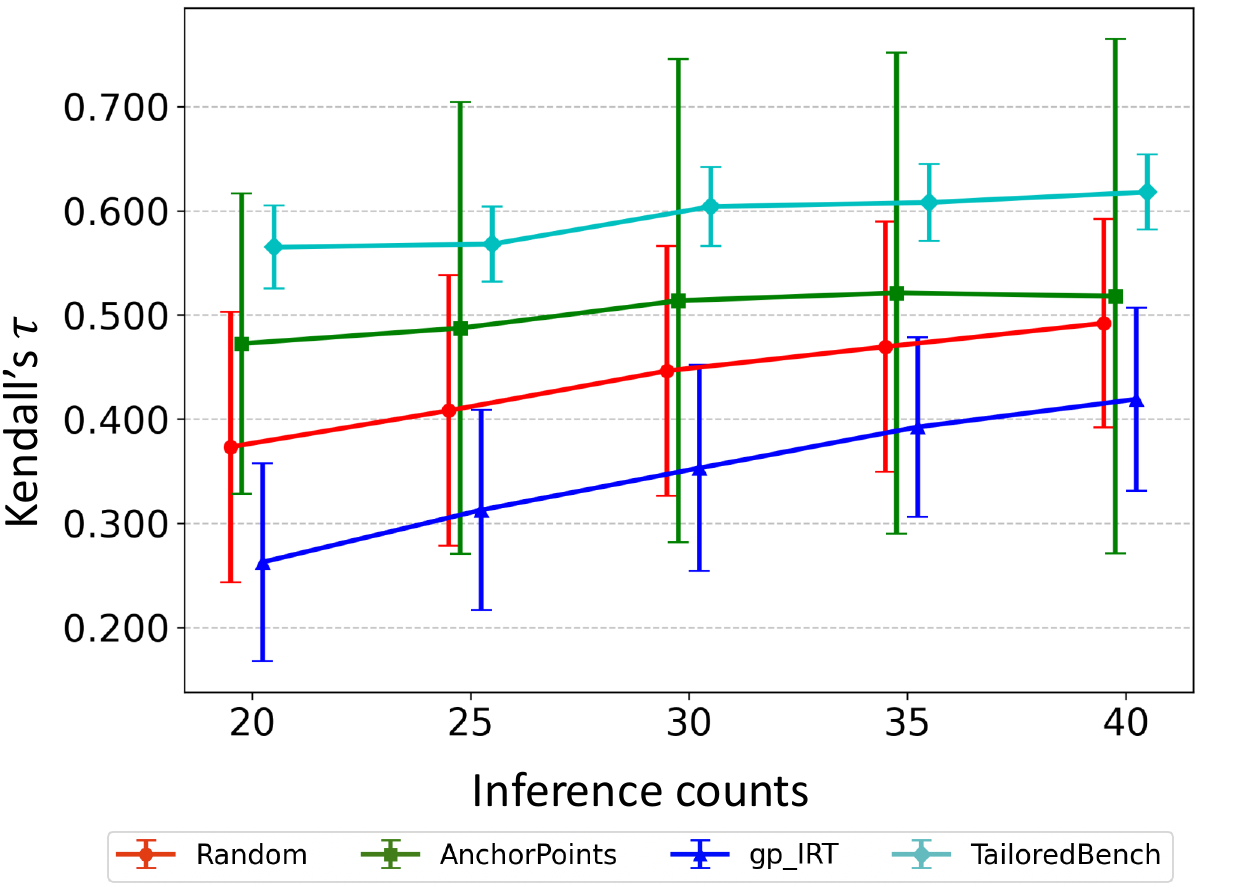} \hfill
  \includegraphics[width=0.49\linewidth]{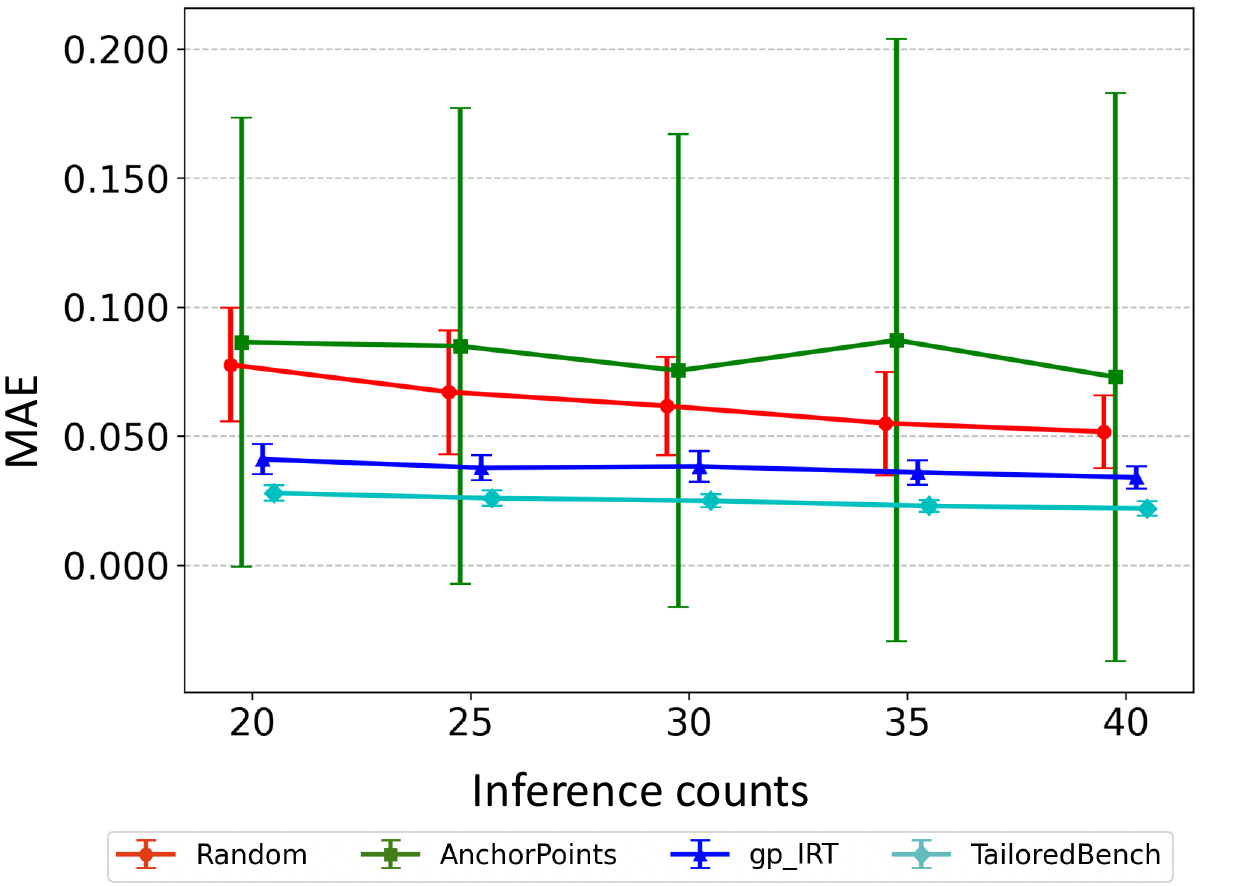}
  \caption {Demonstration of method effectiveness with variance on winogrande benchmark.}
\vspace{-0.3cm}
\label{Variance_winogrande}
\end{figure*}

\begin{figure*}[ht]
  \includegraphics[width=0.49\linewidth]{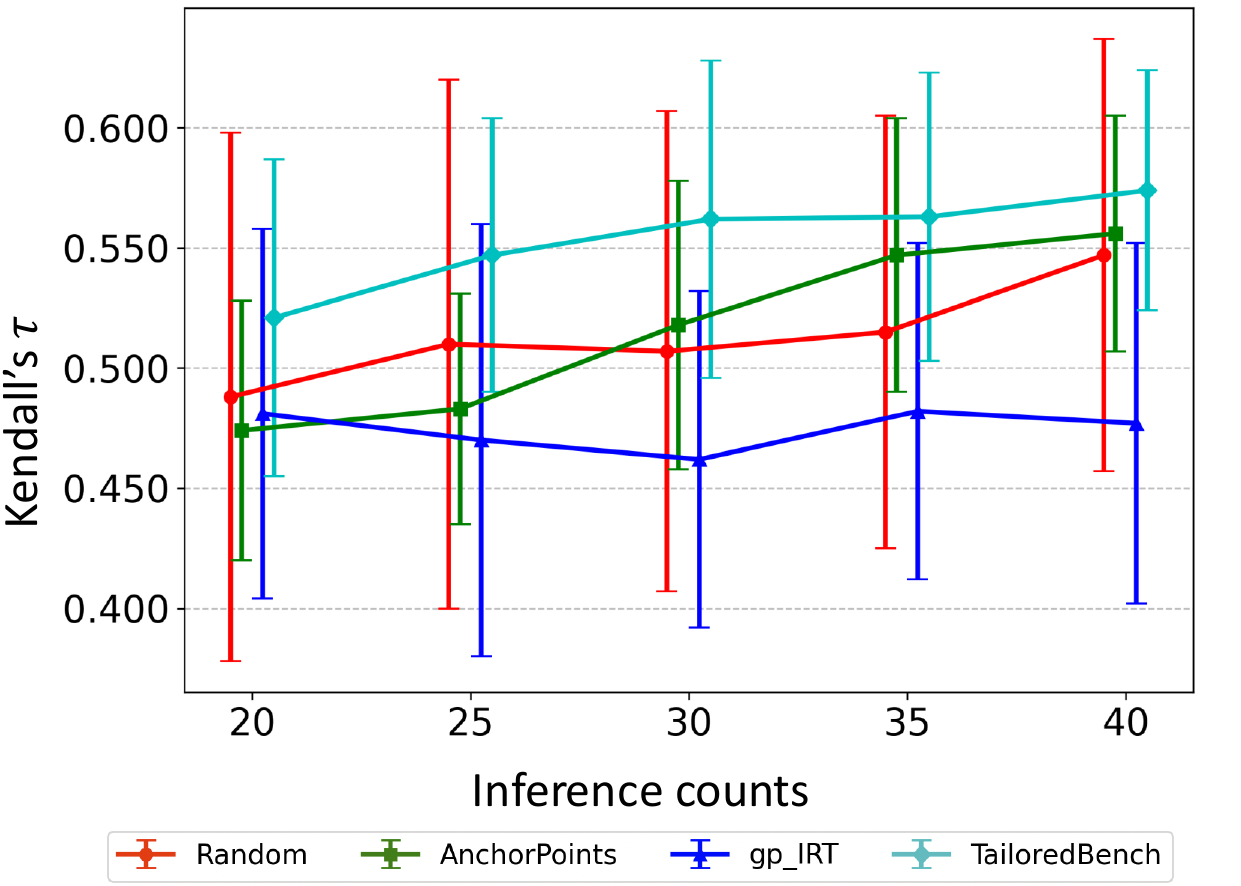} \hfill
  \includegraphics[width=0.49\linewidth]{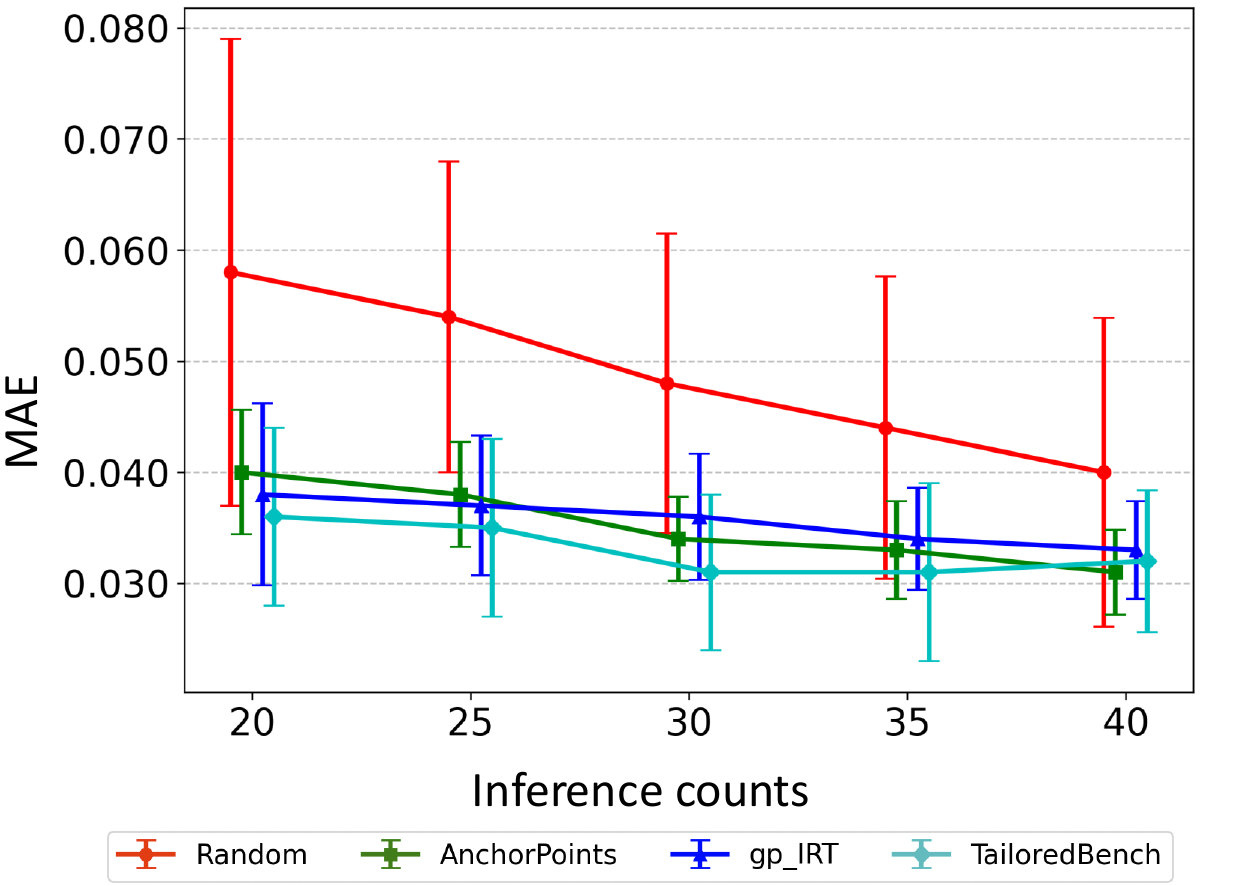}
  \caption {Demonstration of method effectiveness with variance on POPE benchmark.}
\vspace{-0.3cm}
\label{Variance_pope}
\end{figure*}

In this section, we present visual comparisons of our method and other approaches, including their respective variances, as illustrated in Figures \ref{Variance_arc} to \ref{Variance_pope}. The results demonstrate that our method outperforms the baseline methods on all datasets and exhibits greater robustness (with smaller variance).

\subsection{More Analyses On the Impact of Native Source Model Quantity on Our Method}
\label{apd:quantity}
\begin{figure*}[ht]
  \includegraphics[width=0.49\linewidth]{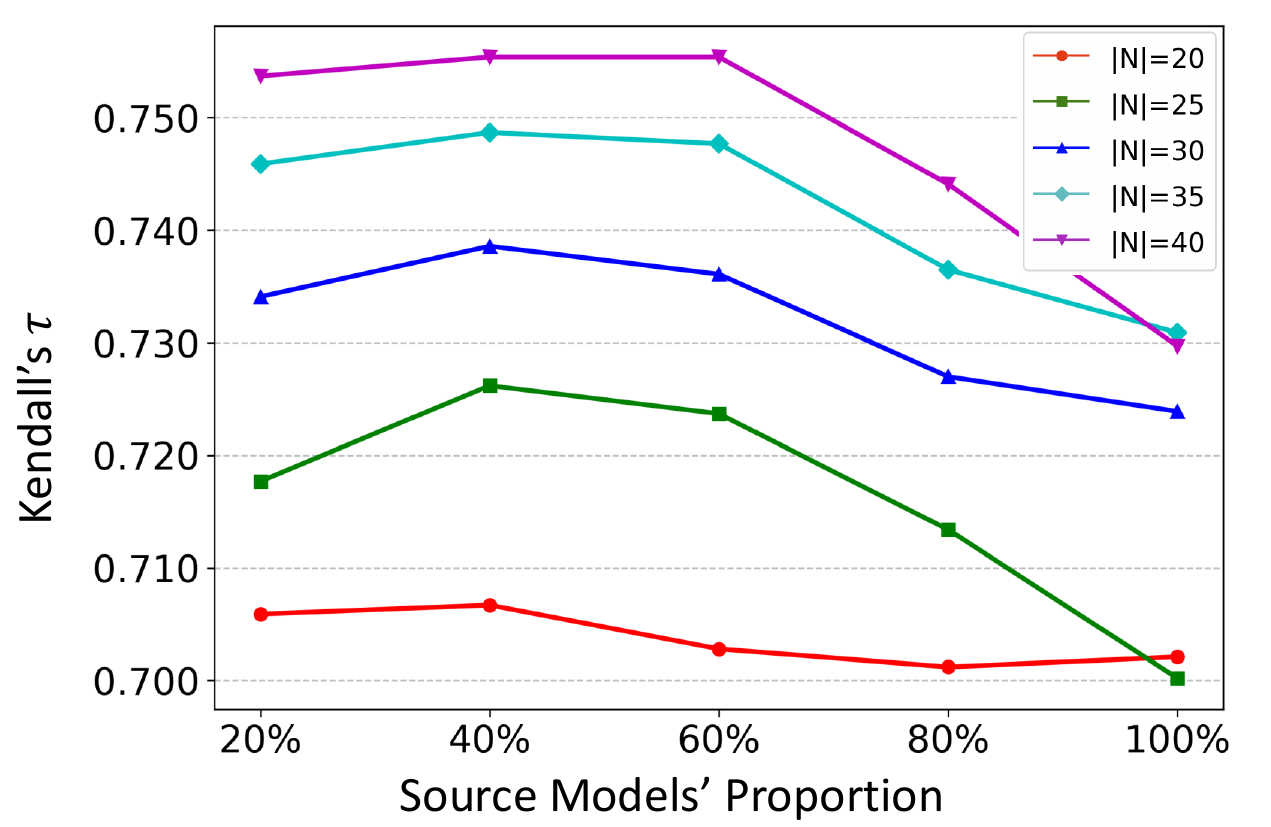} \hfill
  \includegraphics[width=0.49\linewidth]{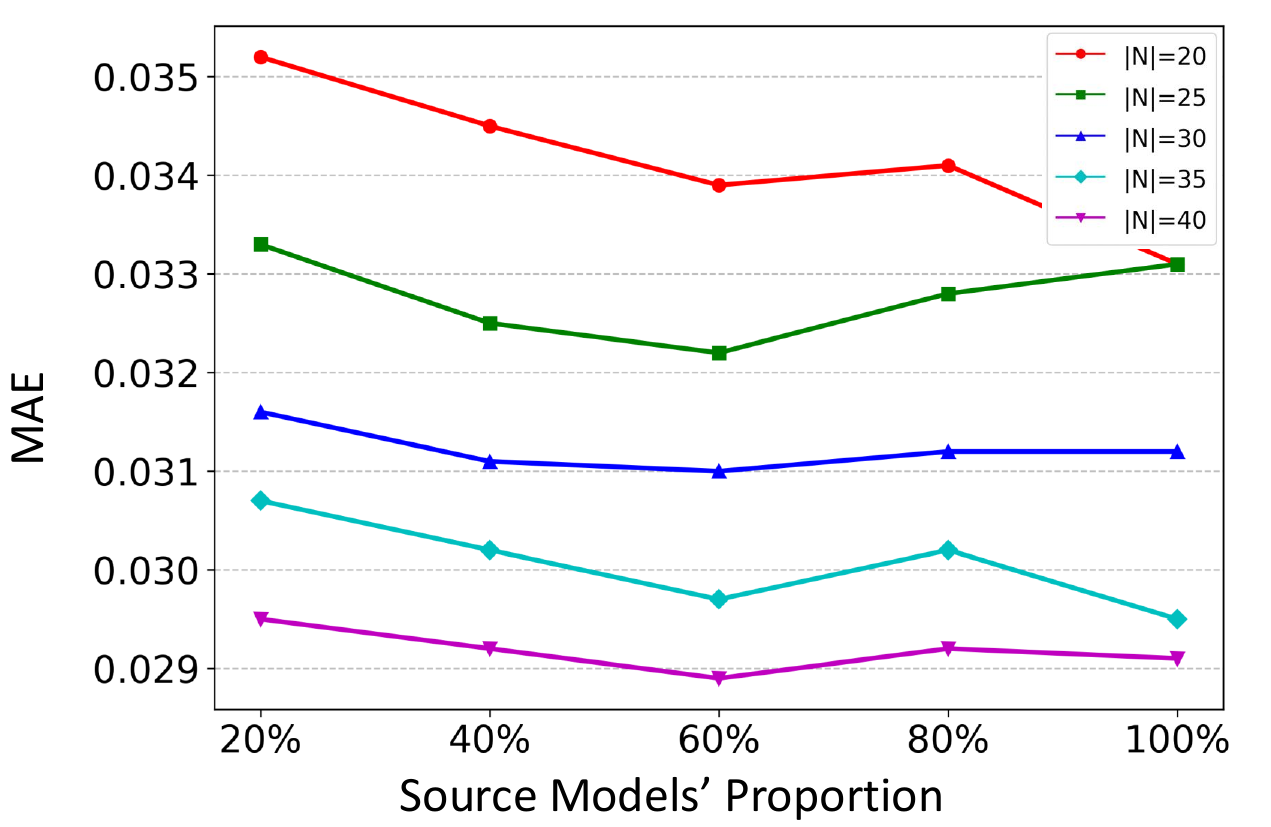}
  \caption {The impact of the quantity of Native Source Models (with
prediction consistency kept the same) on ARC Challenge benchmark.}
\vspace{-0.3cm}
\label{Quantity_arc}
\end{figure*}

\begin{figure*}[ht]
  \includegraphics[width=0.49\linewidth]{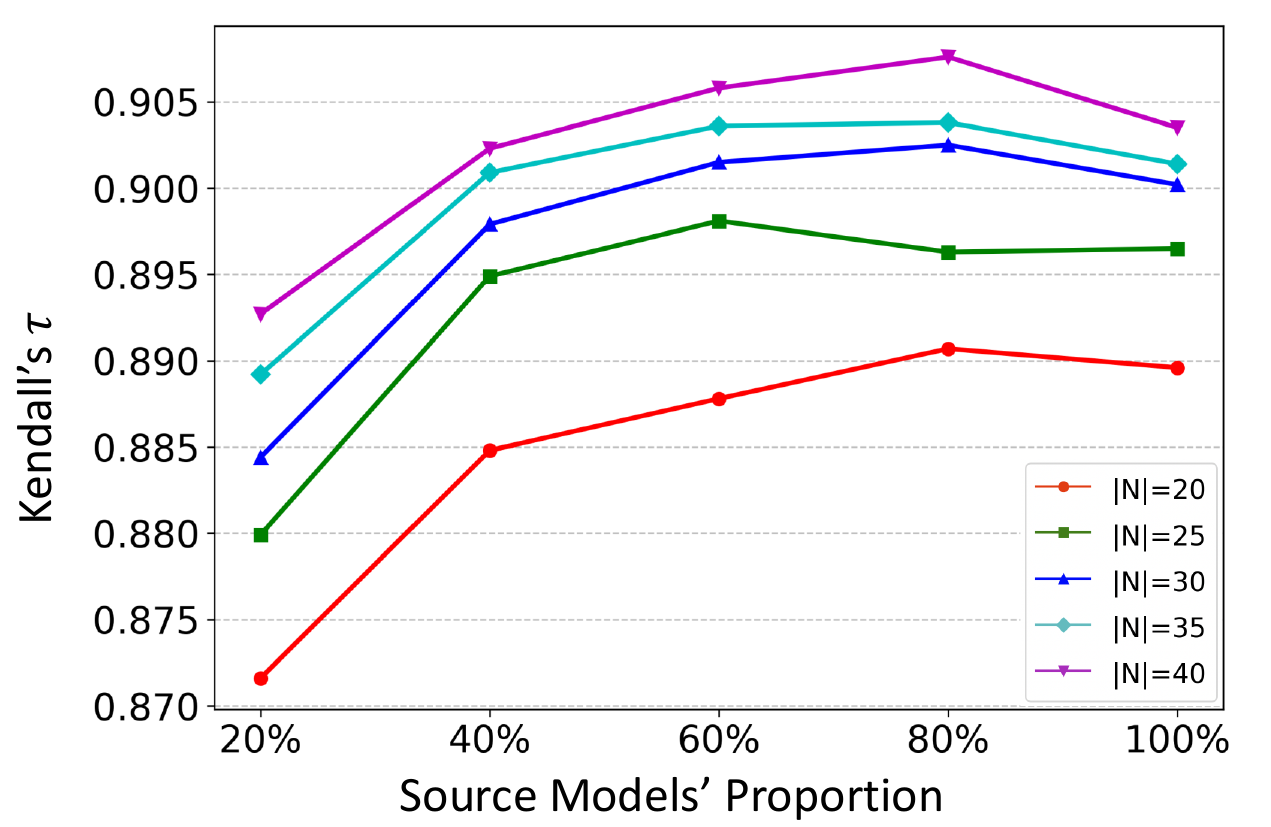} \hfill
  \includegraphics[width=0.49\linewidth]{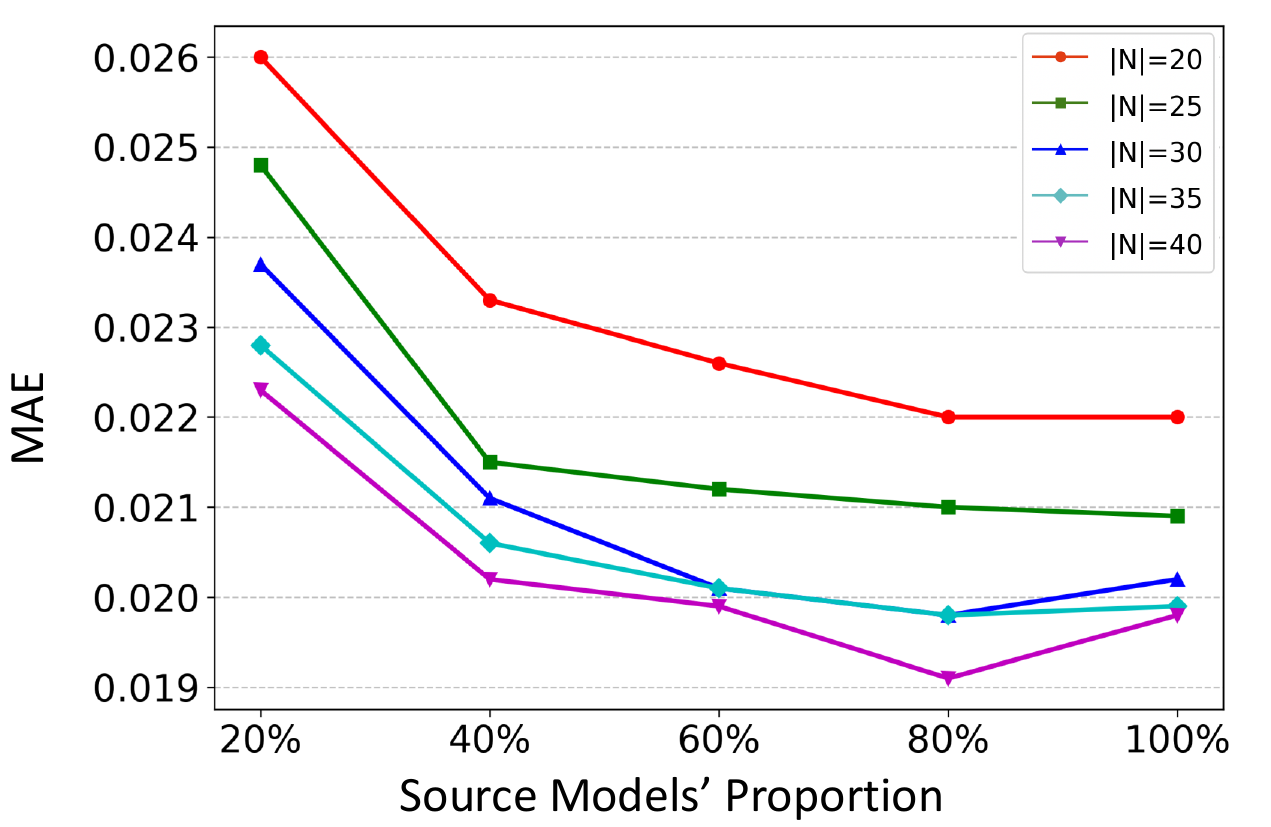}
  \caption {The impact of the quantity of Native Source Models (with prediction consistency kept the same) on Hellaswag benchmark.}
\vspace{-0.3cm}
\label{Quantity_hellaswag}
\end{figure*}

\begin{figure*}[ht]
  \includegraphics[width=0.49\linewidth]{figs/modelratio/GSM8K_modelratio_kendall.pdf} \hfill
  \includegraphics[width=0.49\linewidth]{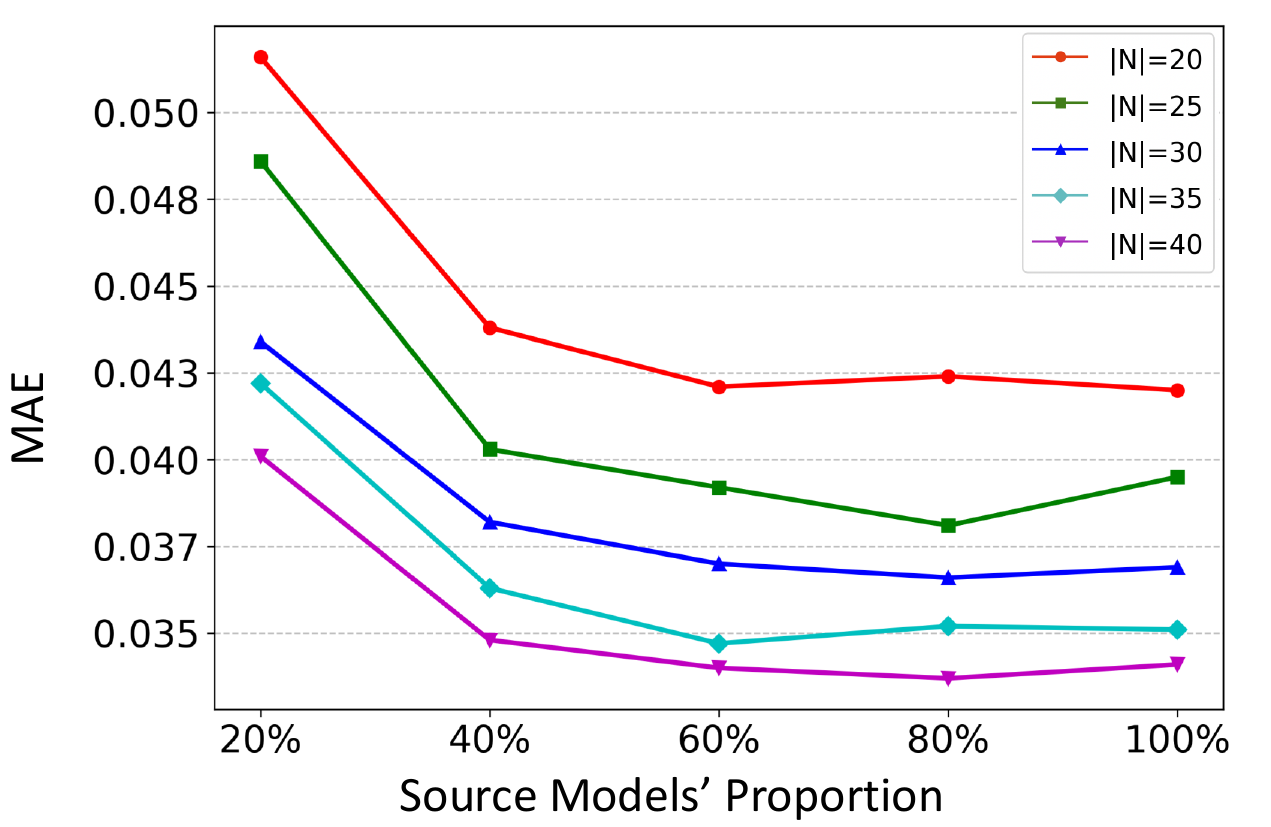}
  \caption {The impact of the quantity of Native Source Models (with prediction consistency kept the same) on GSM8K benchmark.}
\vspace{-0.3cm}
\label{Quantity_GSM8K}
\end{figure*}

\begin{figure*}[ht]
  \includegraphics[width=0.49\linewidth]{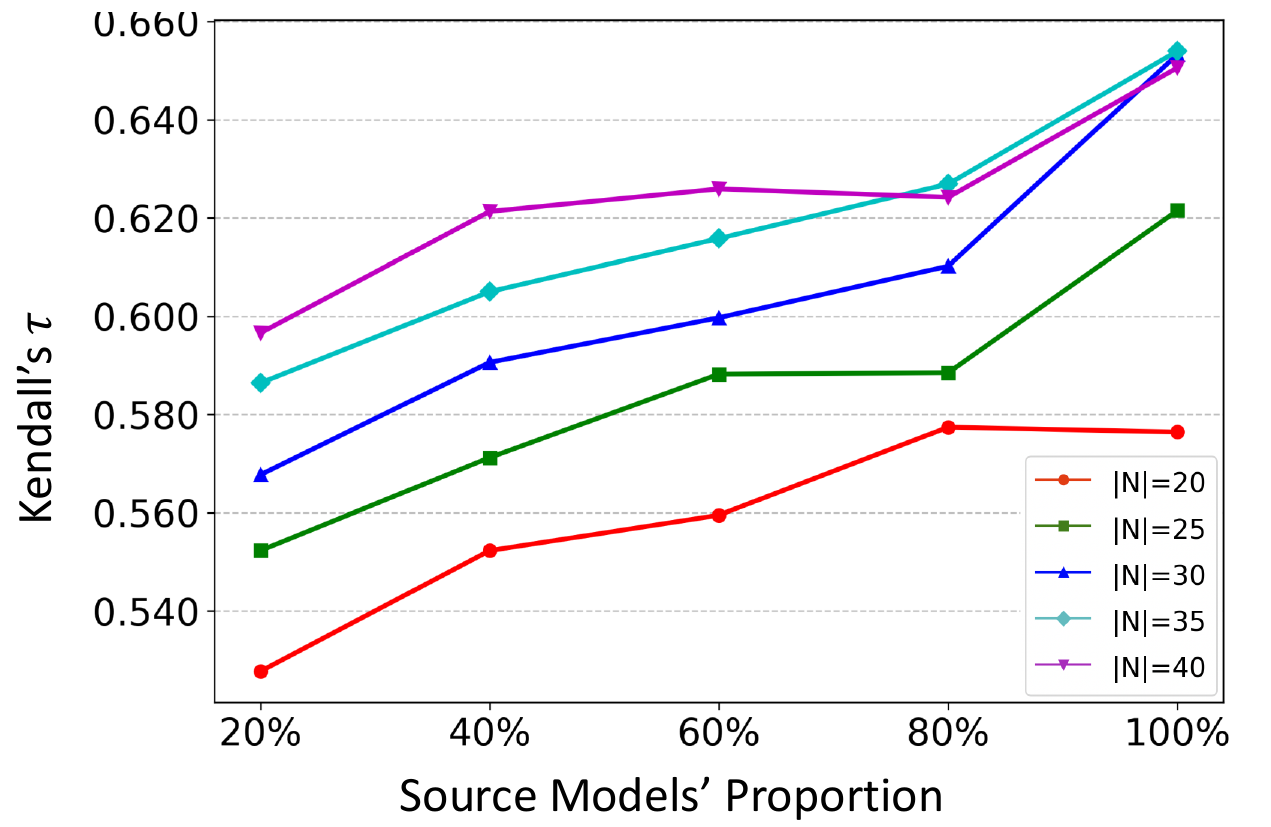} \hfill
  \includegraphics[width=0.49\linewidth]{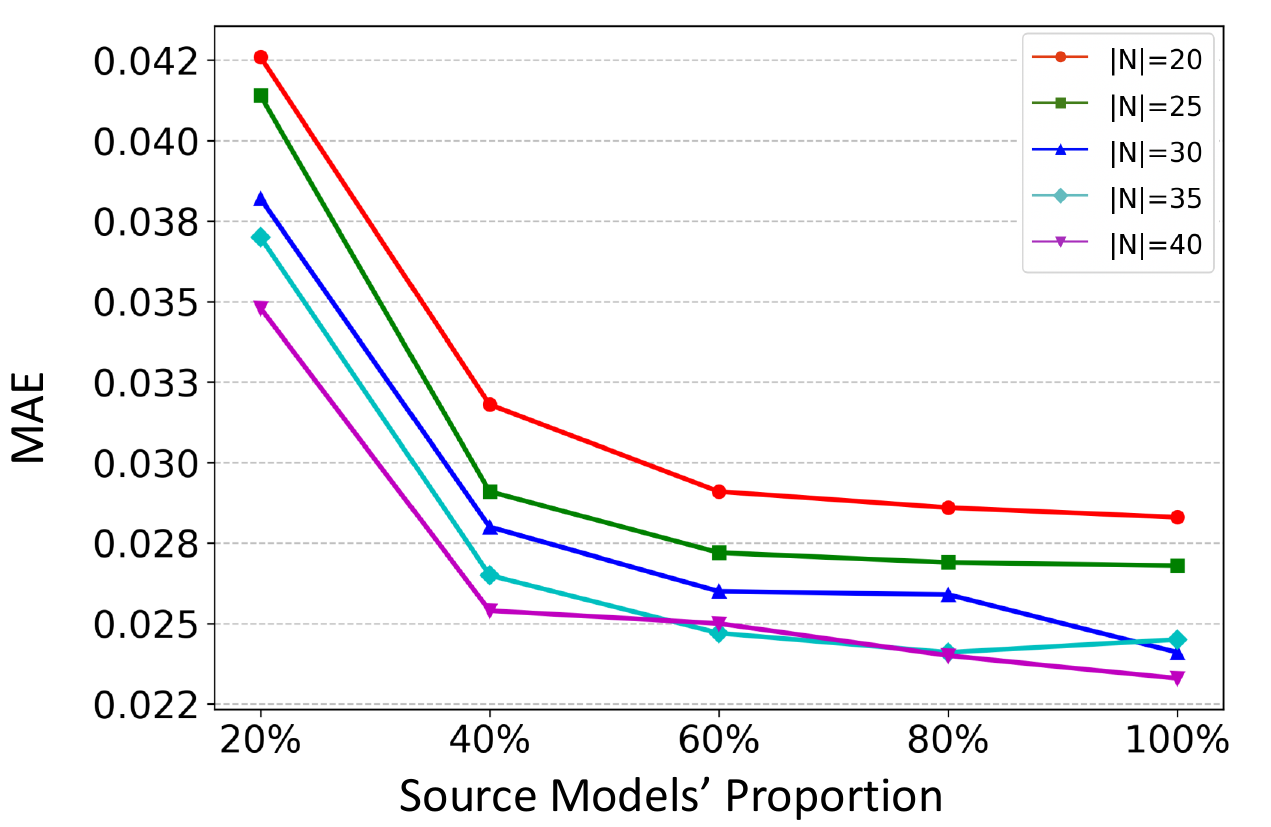}
  \caption {The impact of the quantity of Native Source Models (with prediction consistency kept the same) on Winogrande benchmark.}
\vspace{-0.3cm}
\label{Quantity_winogrande}
\end{figure*}

\begin{figure*}[ht]
  \includegraphics[width=0.49\linewidth]{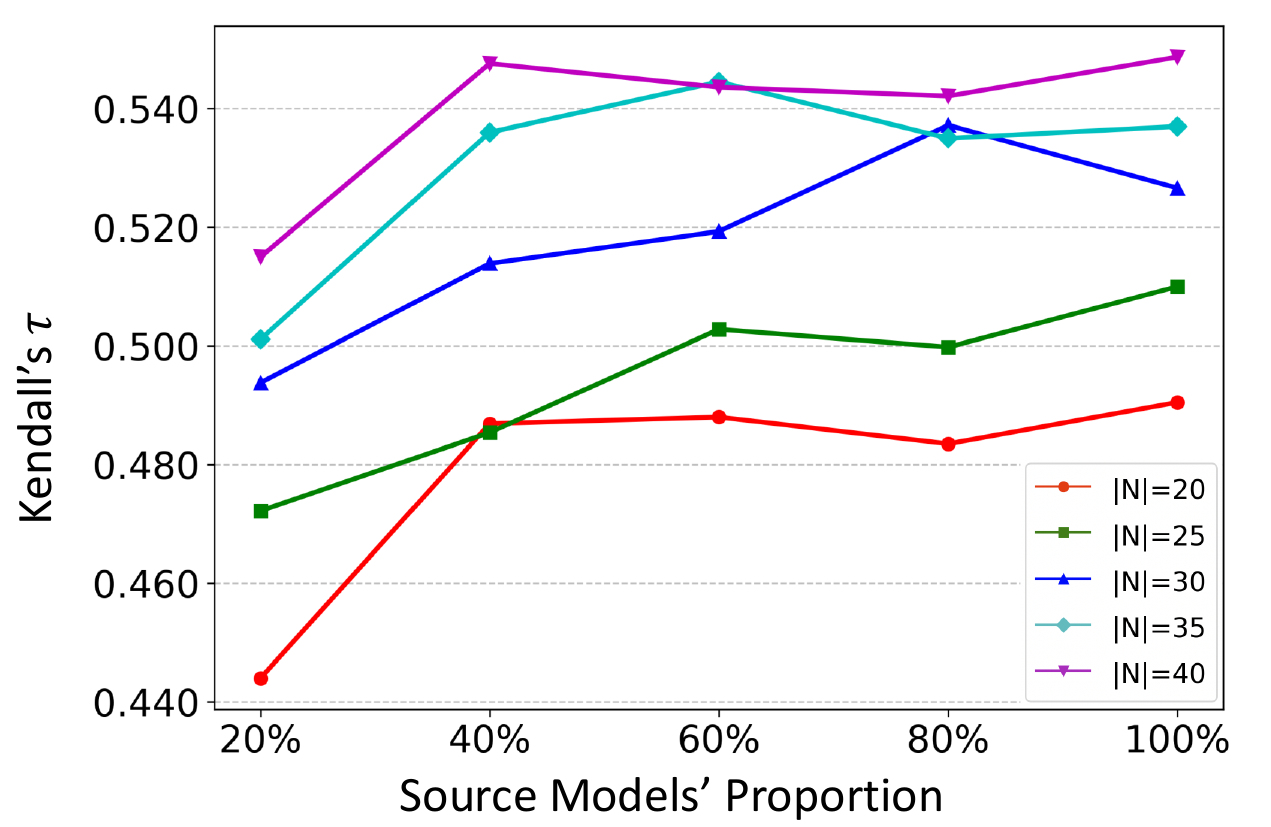} \hfill
  \includegraphics[width=0.49\linewidth]{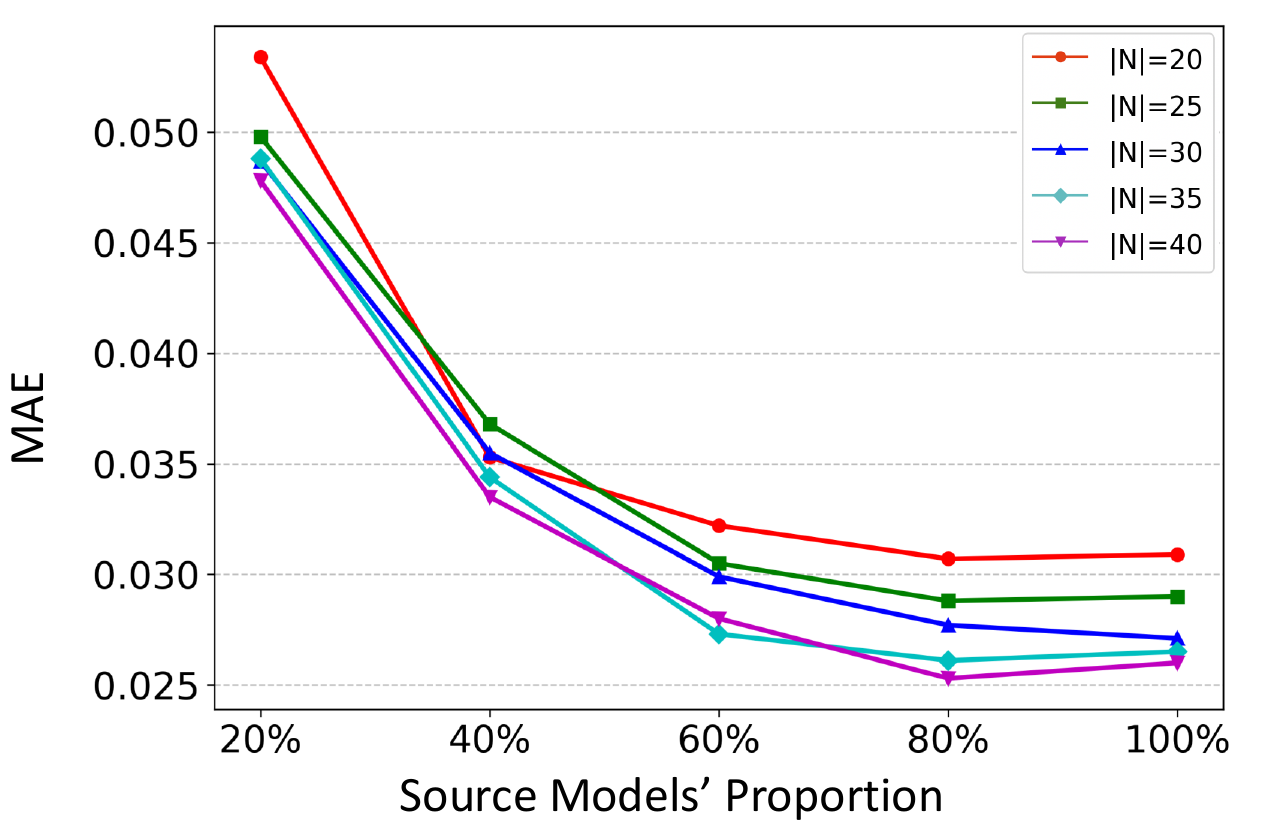}
  \caption {The impact of the quantity of Native Source Models (with prediction consistency kept the same) on POPE benchmark.}
\vspace{-0.3cm}
\label{Quantity_pope}
\end{figure*}

In this section, we maintain the overall prediction consistency between the native source models and the target models constant, while varying the proportion of the source models designated as native source models from 20\% to 100\% for the target models across various benchmarks. The results are illustrated in Figures \ref{Quantity_arc} to \ref{Quantity_pope}, indicating that, under the condition of maintaining the prediction consistency between the native source models and the target model, the number of native source models significantly influences the method's performance.

\subsection{More Analyses On the Impact of Native Source Models' Prediction Consistency on Our Method}
\label{apd:similarity}
\begin{figure*}[ht]
  \includegraphics[width=0.49\linewidth]{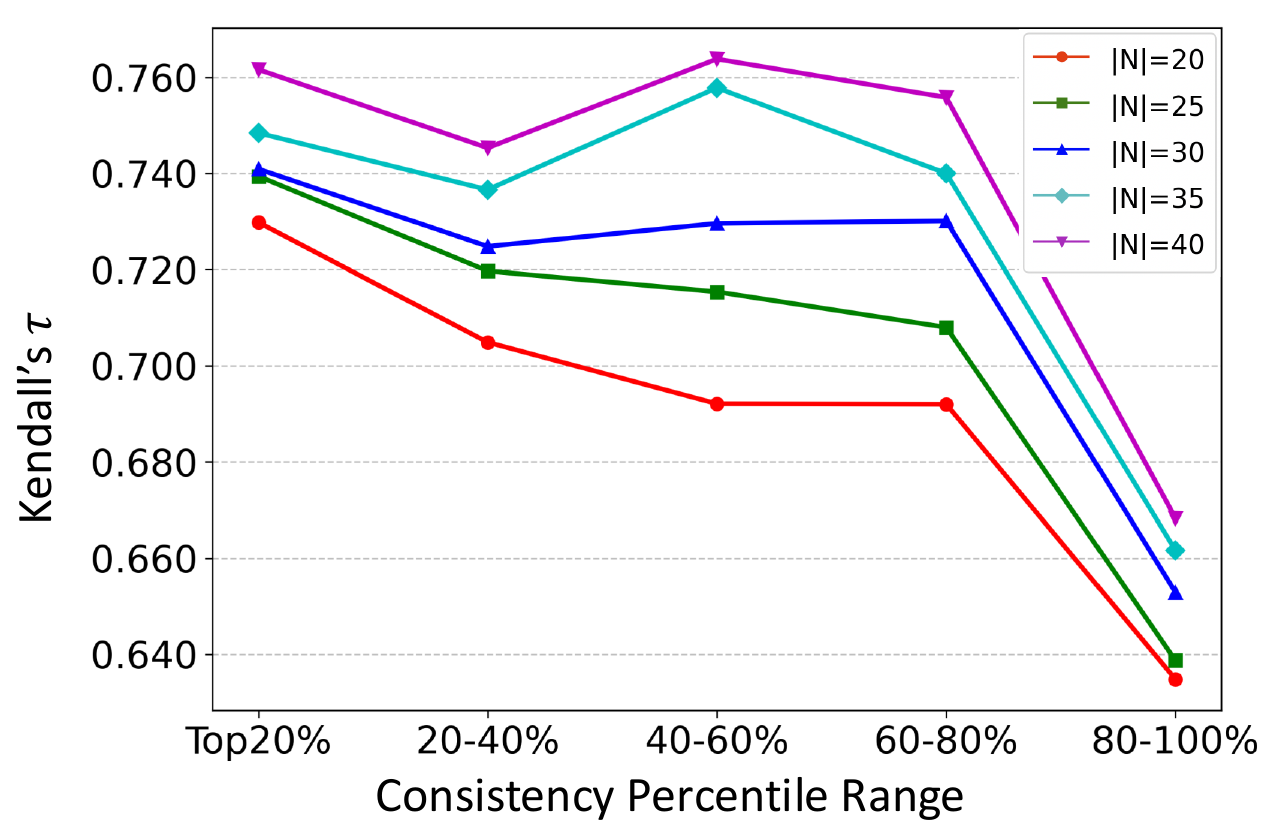} \hfill
  \includegraphics[width=0.49\linewidth]{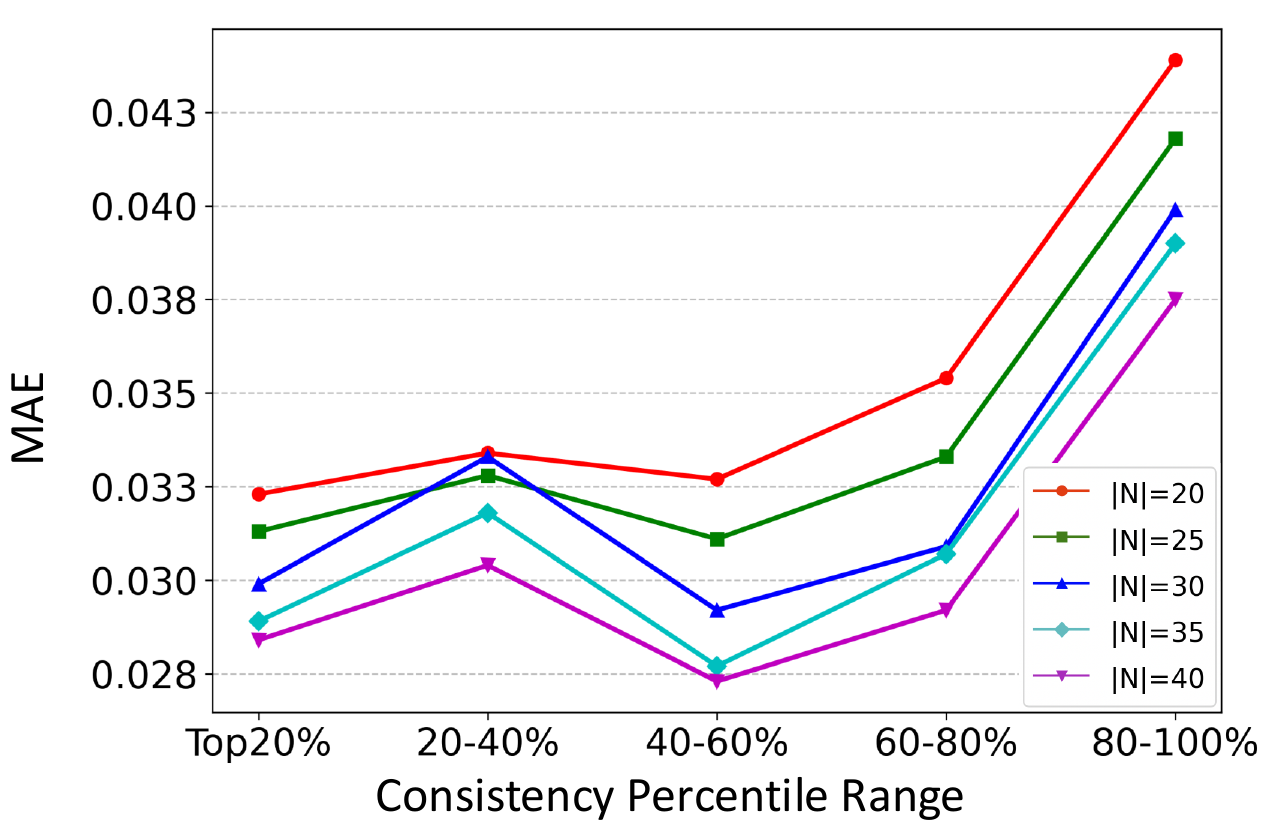}
  \caption {The impact of prediction consistency between the Native Source Model and Target Model (with quantity kept the same) on ARC Challenge benchmark.}
\vspace{-0.3cm}
\label{Similarity_arc}
\end{figure*}

\begin{figure*}[ht]
  \includegraphics[width=0.49\linewidth]{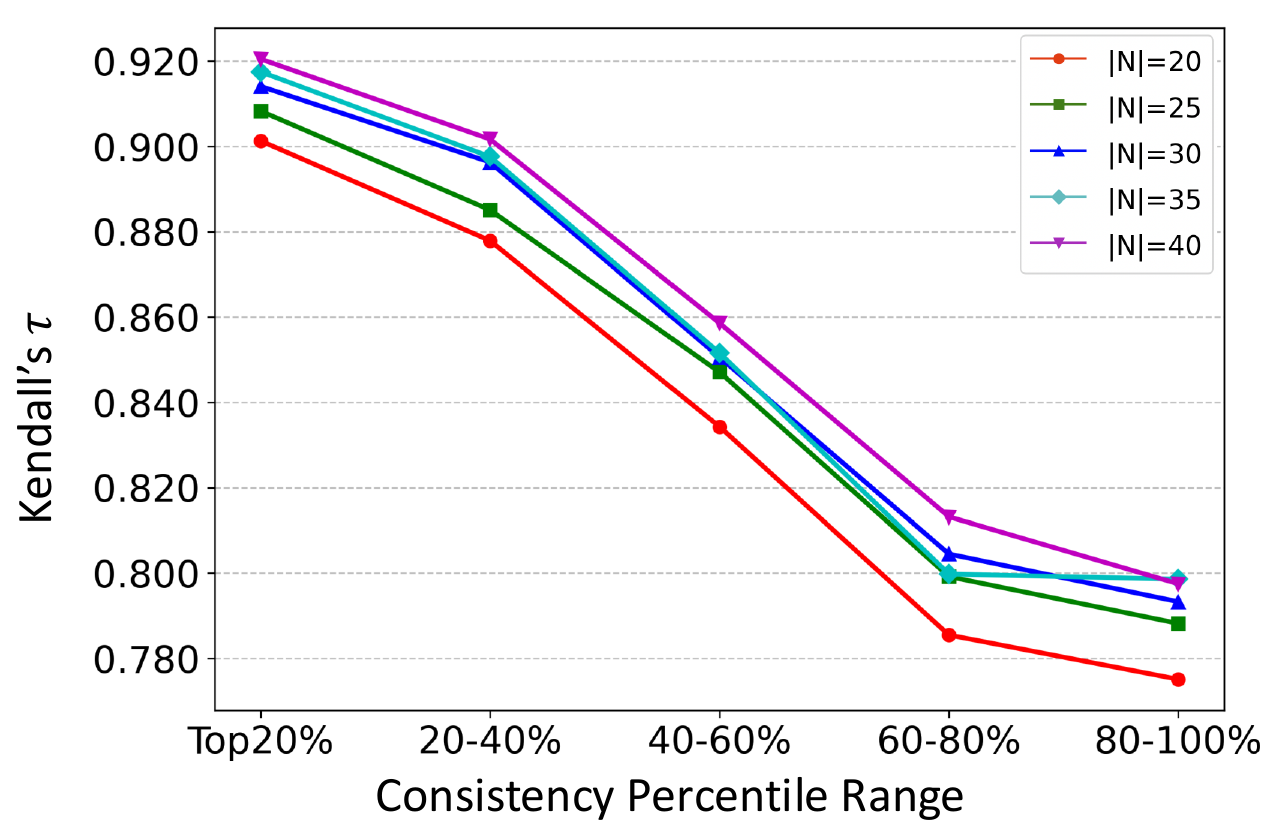} \hfill
  \includegraphics[width=0.49\linewidth]{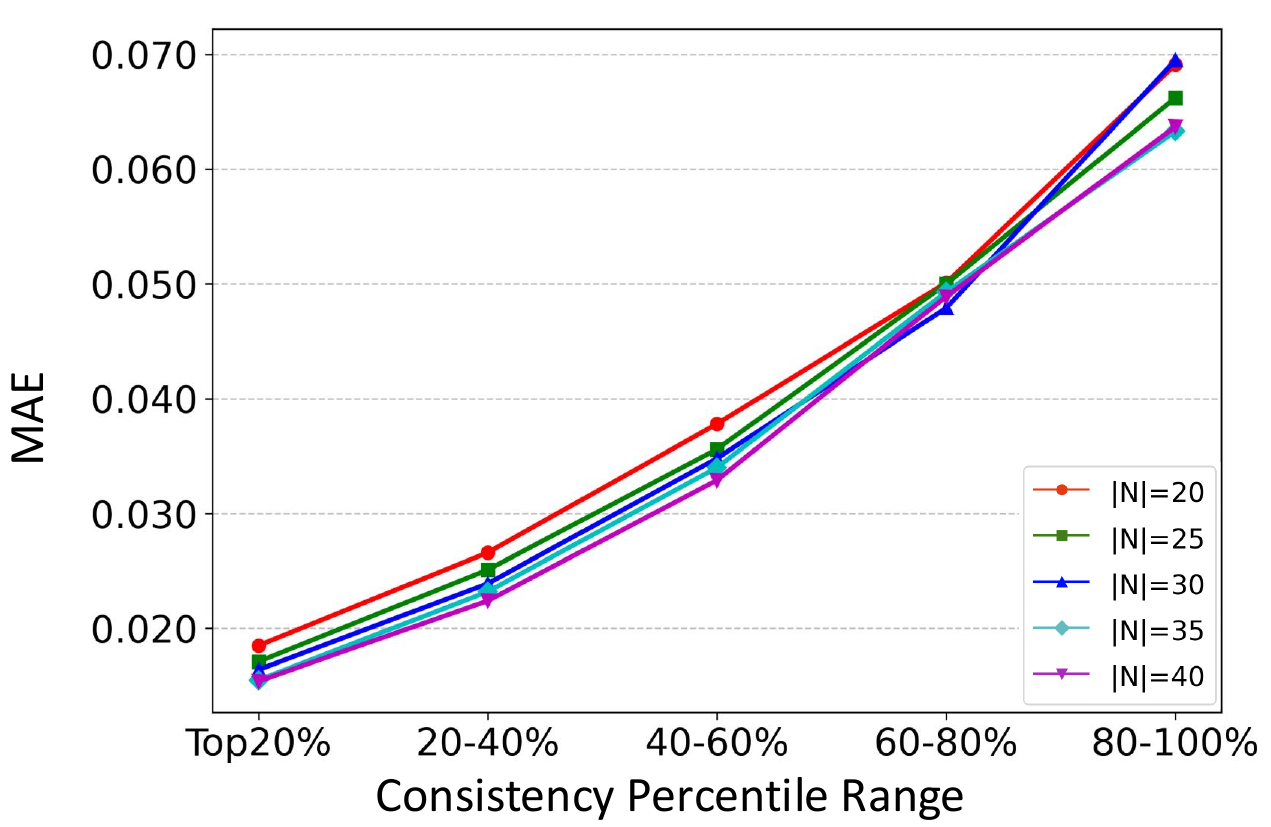}
  \caption {The impact of prediction consistency between the Native Source Model and Target Model (with quantity kept the same) on Hellaswag benchmark.}
\vspace{-0.3cm}
\label{Similarity_hella}
\end{figure*}

\begin{figure*}[ht]
  \includegraphics[width=0.49\linewidth]{figs/modelsimilarity/GSM8K_similarity_kendall.pdf} \hfill
  \includegraphics[width=0.49\linewidth]{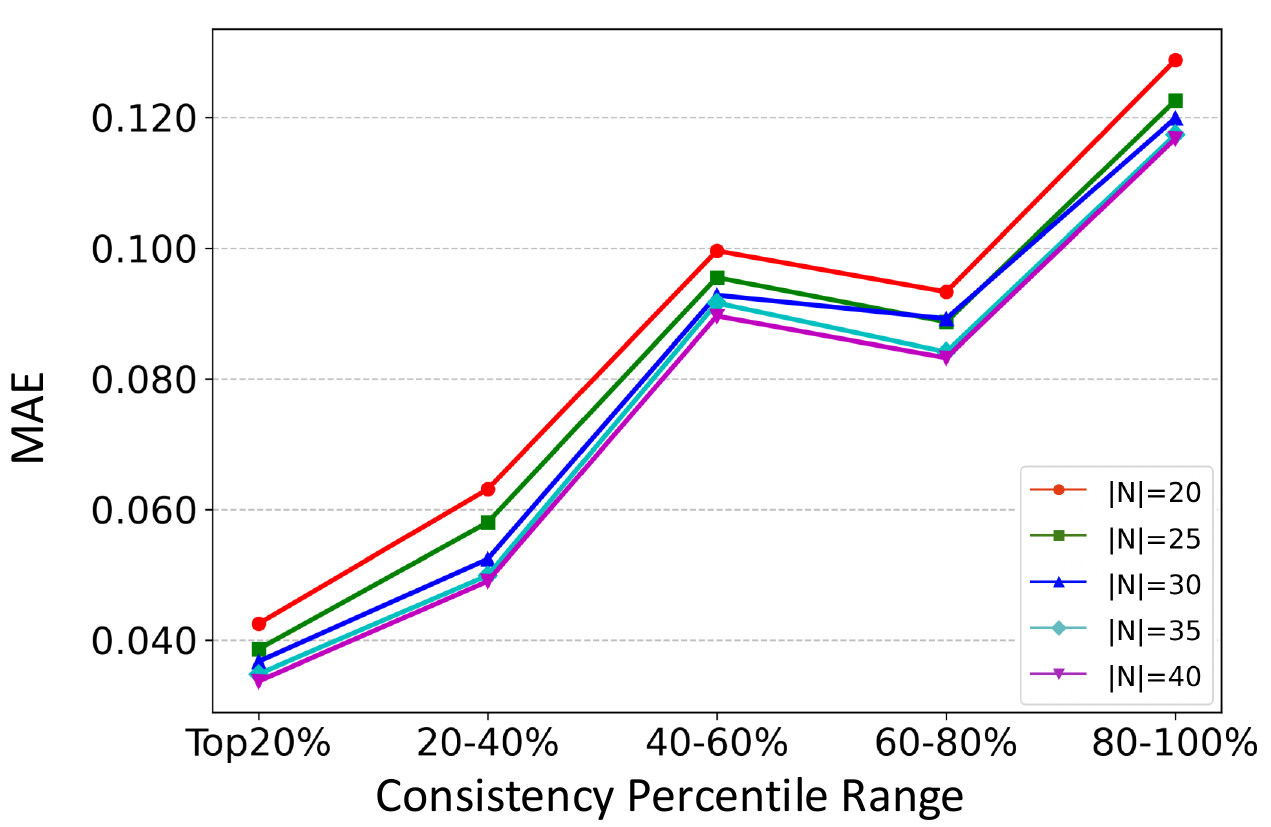}
  \caption {The impact of prediction consistency between the Native Source Model and Target Model (with quantity kept the same) on GSM8K benchmark.}
\vspace{-0.3cm}
\label{Similarity_GSM8K}
\end{figure*}

\begin{figure*}[ht]
  \includegraphics[width=0.49\linewidth]{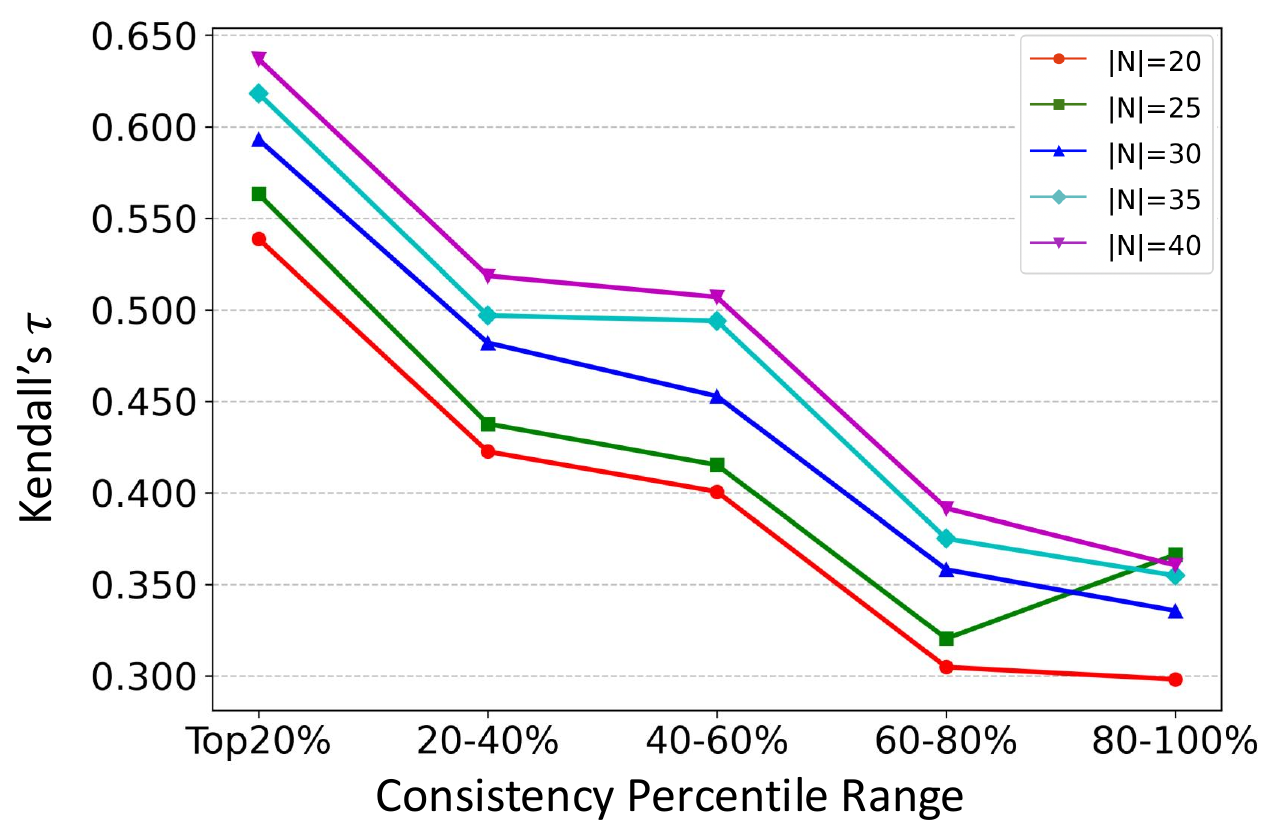} \hfill
  \includegraphics[width=0.49\linewidth]{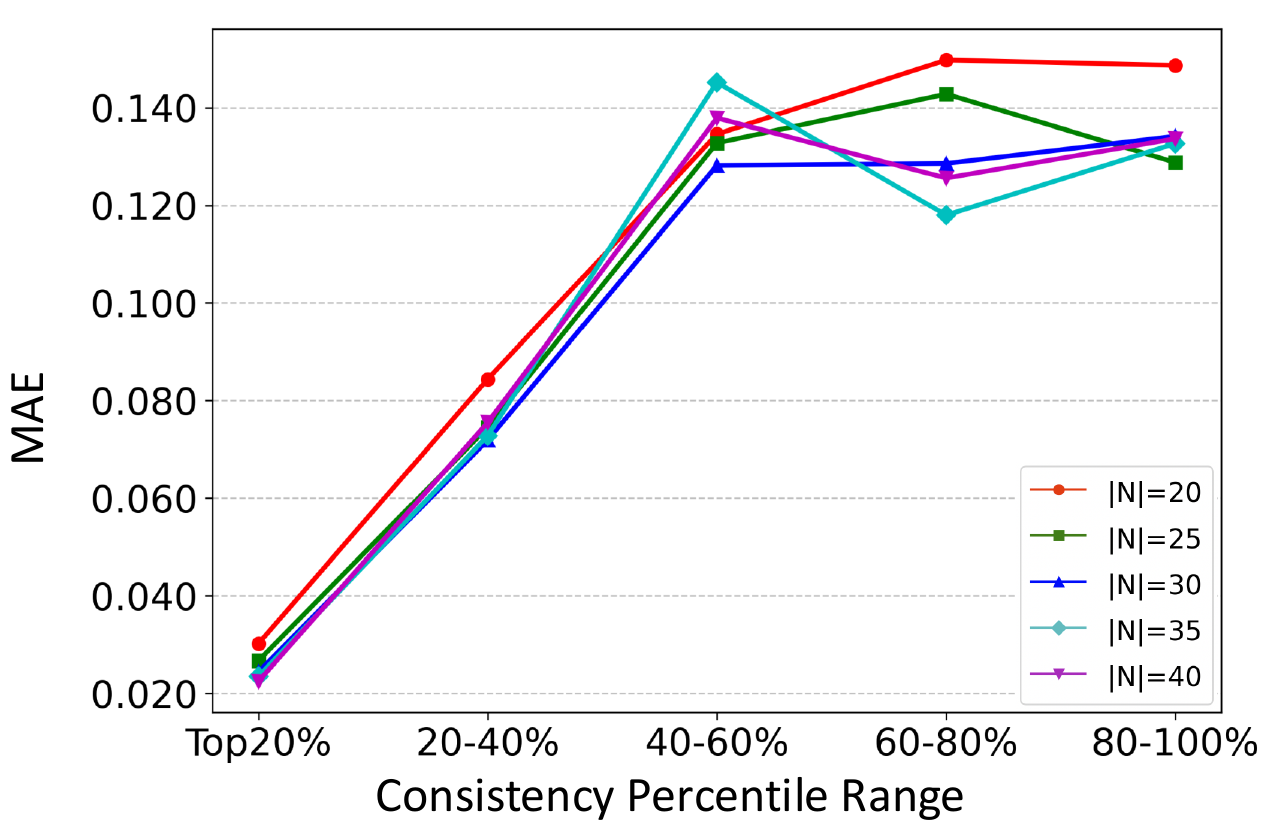}
  \caption {The impact of prediction consistency between the Native Source Model and Target Model (with quantity kept the same) on Winogrande benchmark.}
\vspace{-0.3cm}
\label{Similarity_winogrande}
\end{figure*}

\begin{figure*}[ht]
  \includegraphics[width=0.49\linewidth]{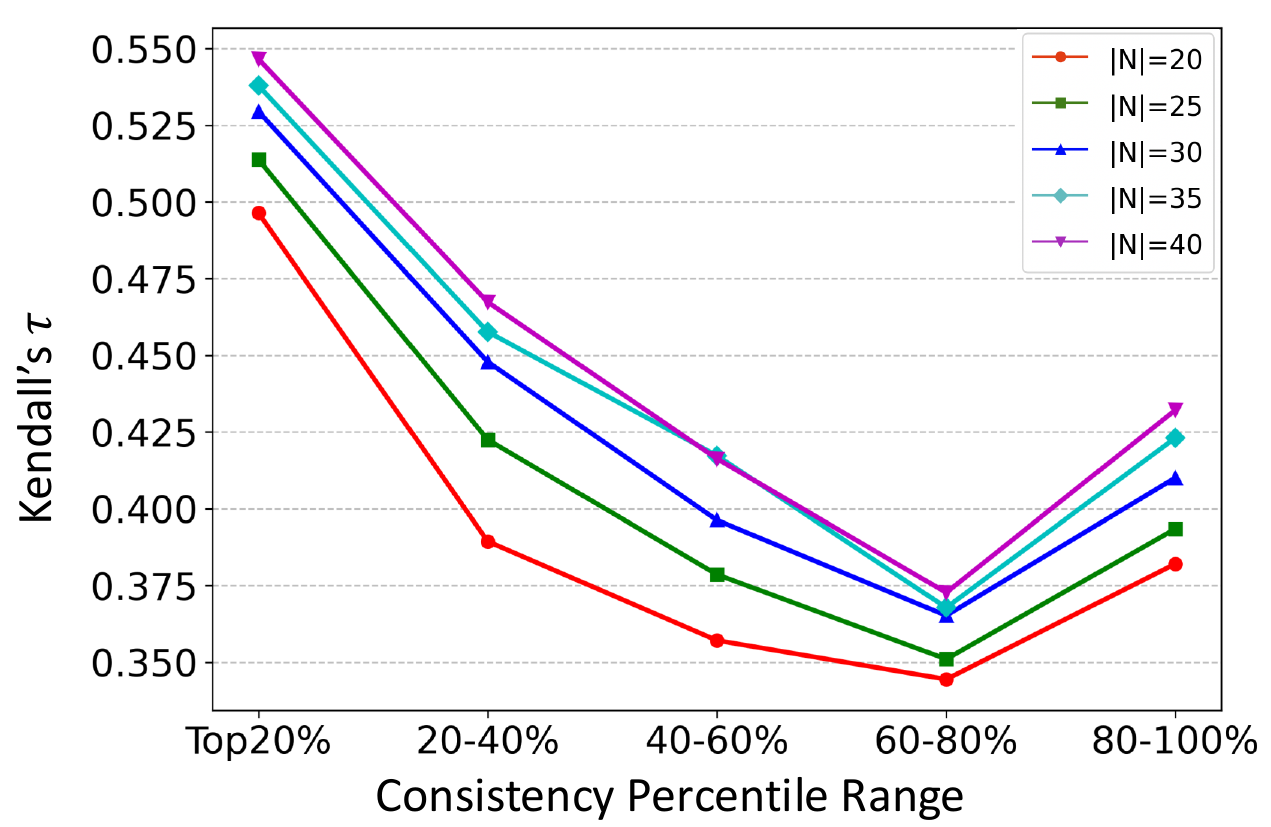} \hfill
  \includegraphics[width=0.49\linewidth]{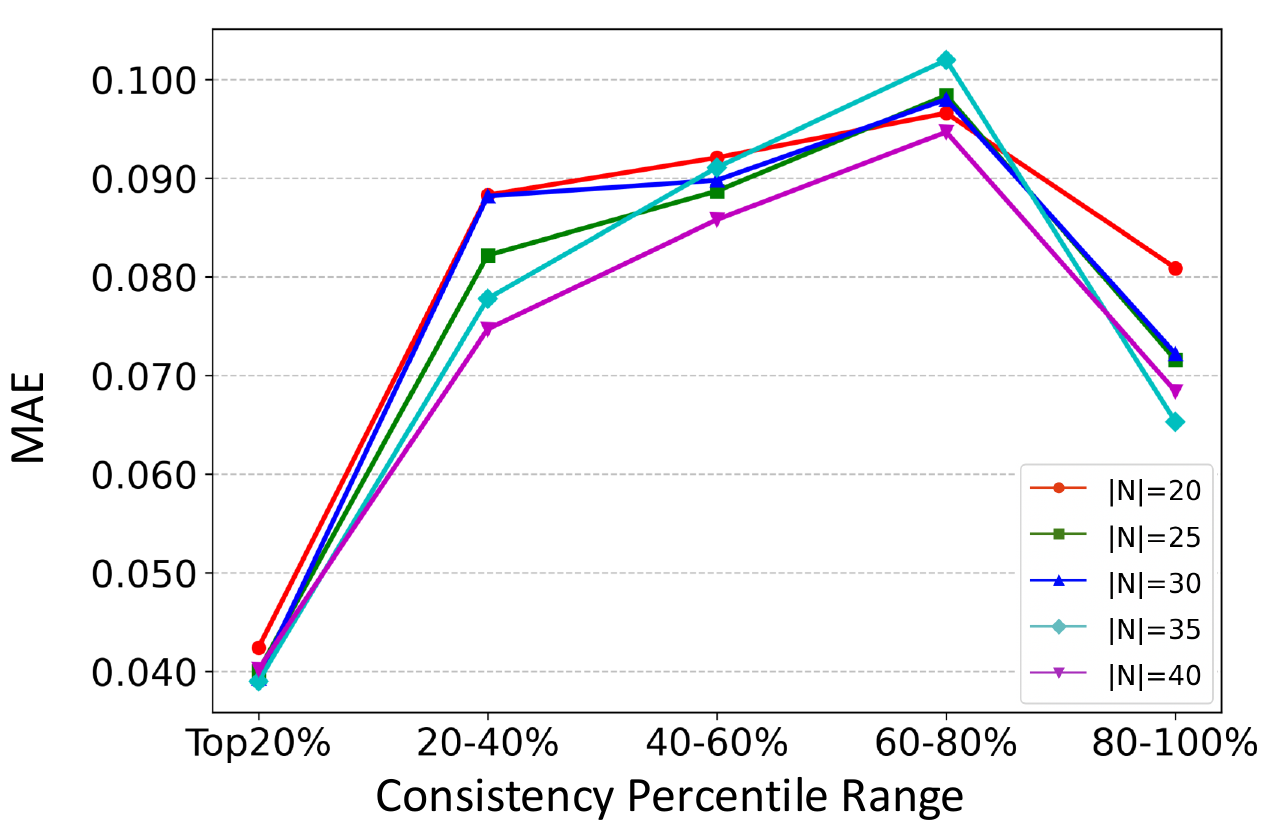}
  \caption {The impact of prediction consistency between the Native Source Model and Target Model (with quantity kept the same) on POPE benchmark.}
\vspace{-0.3cm}
\label{Similarity_pope}
\end{figure*}

We conduct ablation studies by selecting native source models based on their prediction consistency with the target model across various benchmarks, ranging from the top 20\% to the 80\%\textasciitilde100\% range. The results, presented in Figures \ref{Similarity_arc} to \ref{Similarity_pope}, indicate that the performance of the method significantly declines as the prediction consistency between the native source models and the target model decreases, under the condition of keeping the number of native source models constant.

\subsection{Extended Results on Optimal Native Source Model Selection}
\begin{figure*}[htbp]
  \includegraphics[width=0.49\linewidth]{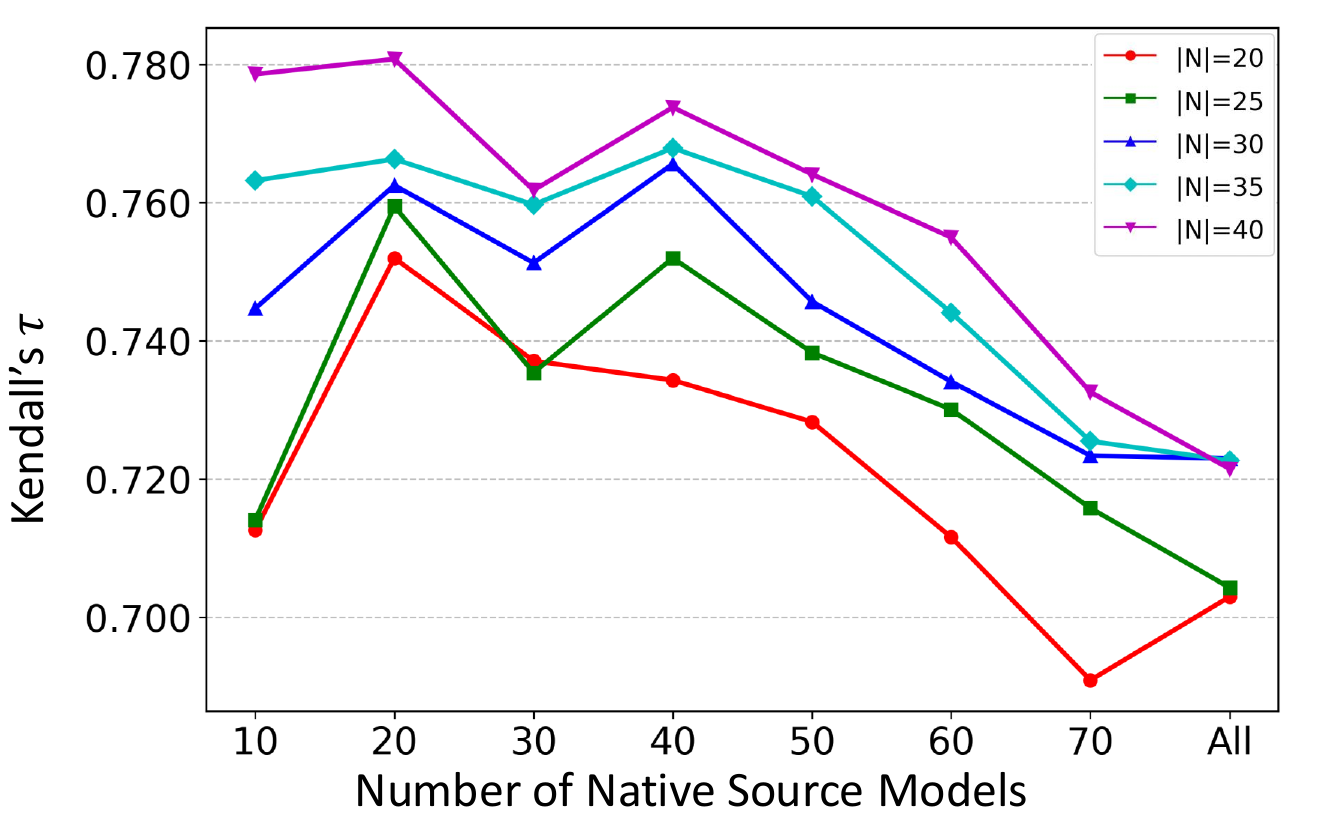} \hfill
  \includegraphics[width=0.49\linewidth]{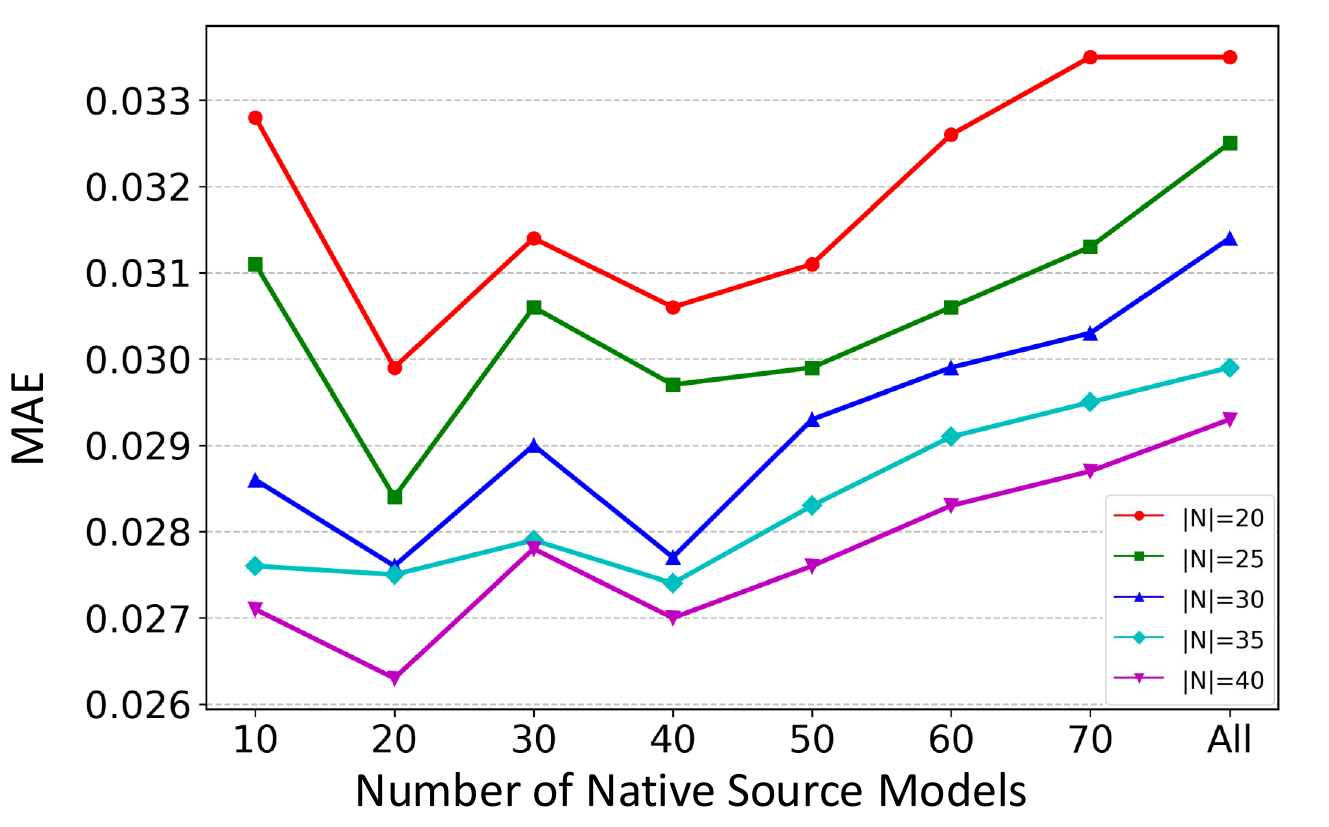}
  \caption {Performance of \textsc{TailoredBench} with varying numbers of Native Source Models on ARC Challenge benchmark.}
\vspace{-0.3cm}
\label{mainanalysis_arc}
\end{figure*}

\begin{figure*}[ht]
  \includegraphics[width=0.49\linewidth]{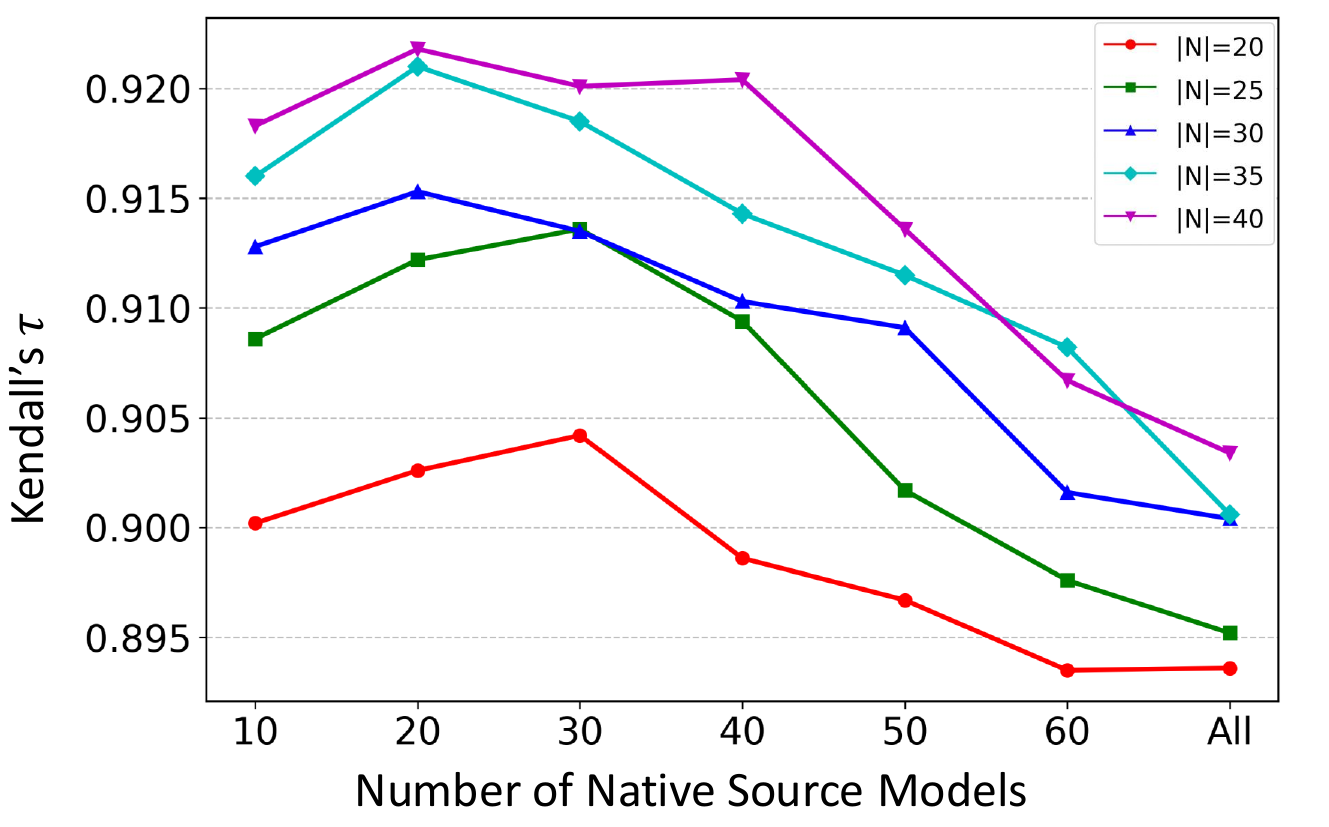} \hfill
  \includegraphics[width=0.49\linewidth]{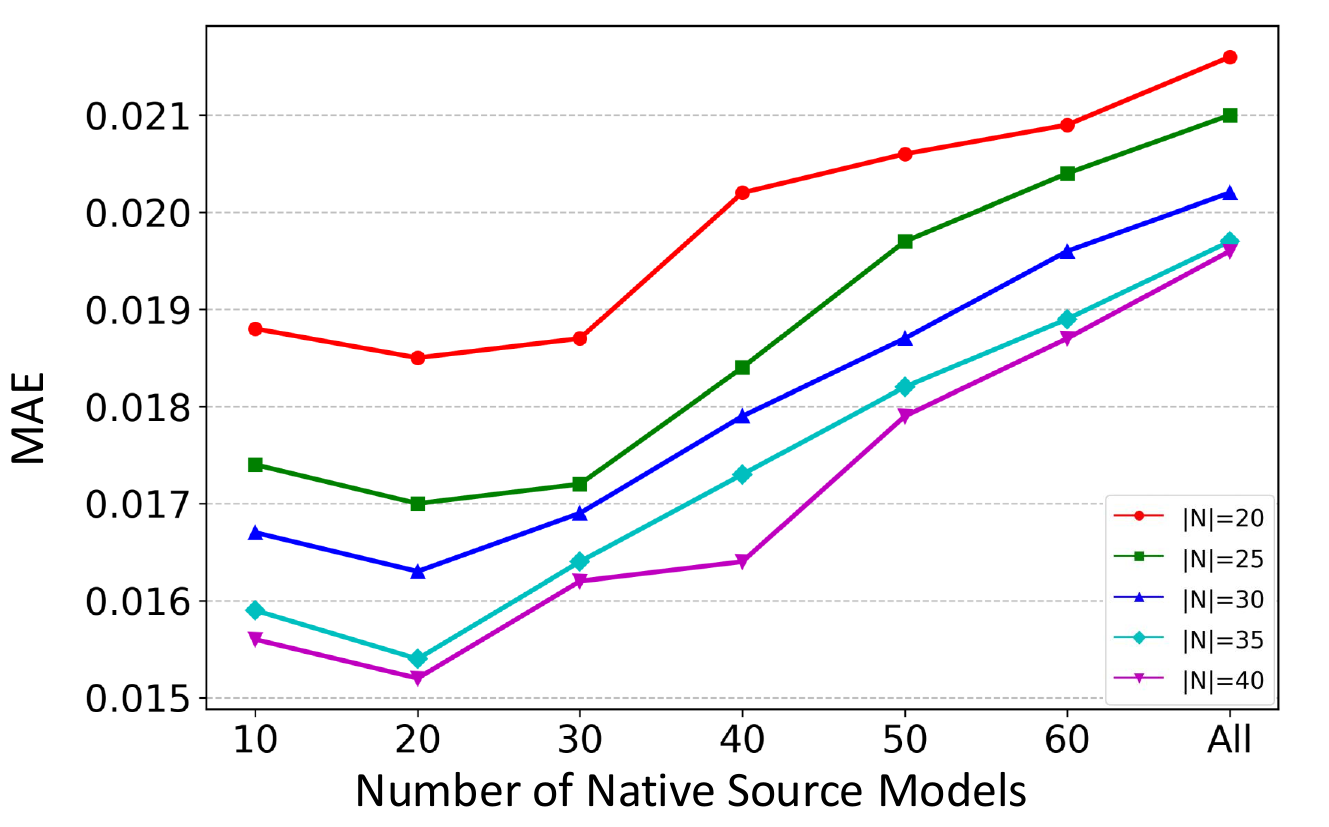}
  \caption {Performance of \textsc{TailoredBench} with varying numbers of Native Source Models on Hellaswag benchmark.}
\vspace{-0.3cm}
\label{mainanalysis_hella}
\end{figure*}

\begin{figure*}[ht]
  \includegraphics[width=0.49\linewidth]{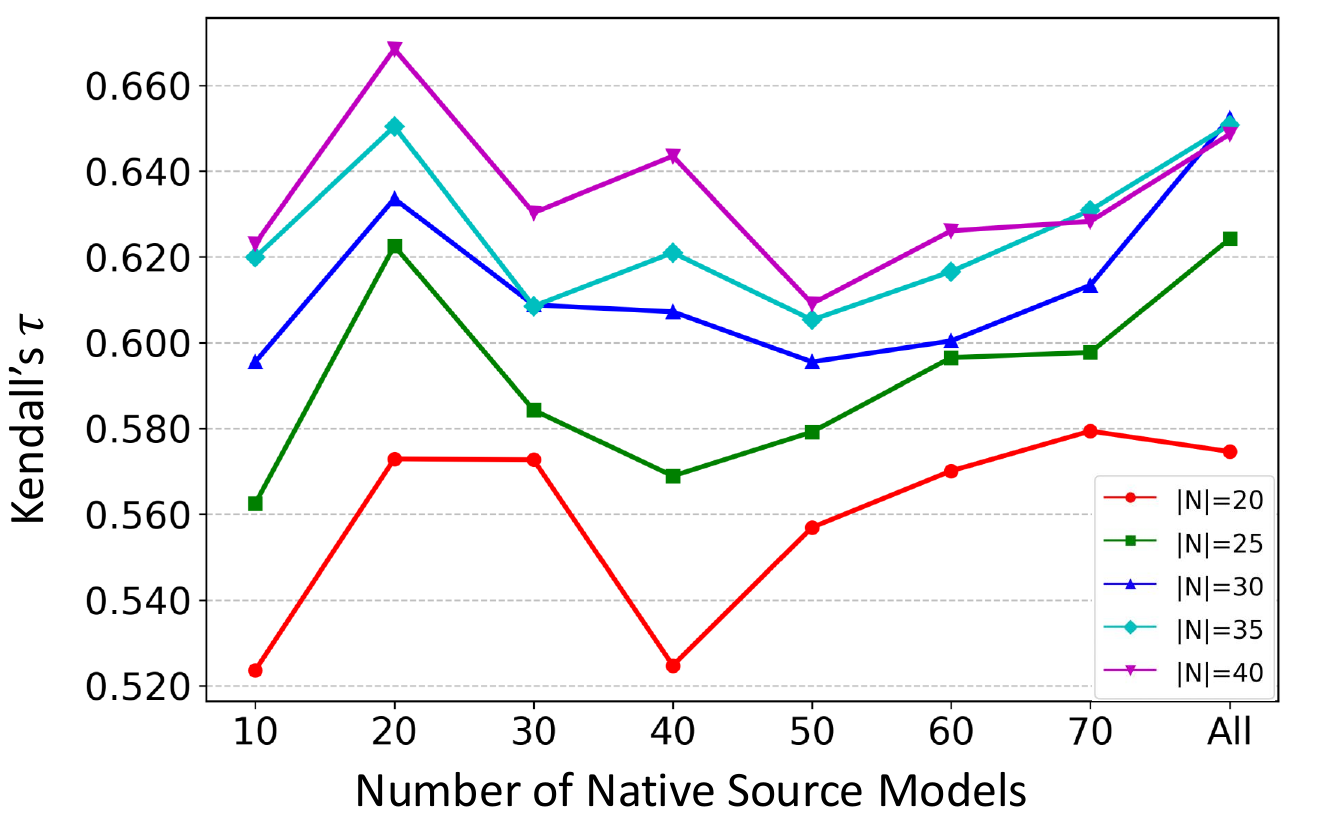} \hfill
  \includegraphics[width=0.49\linewidth]{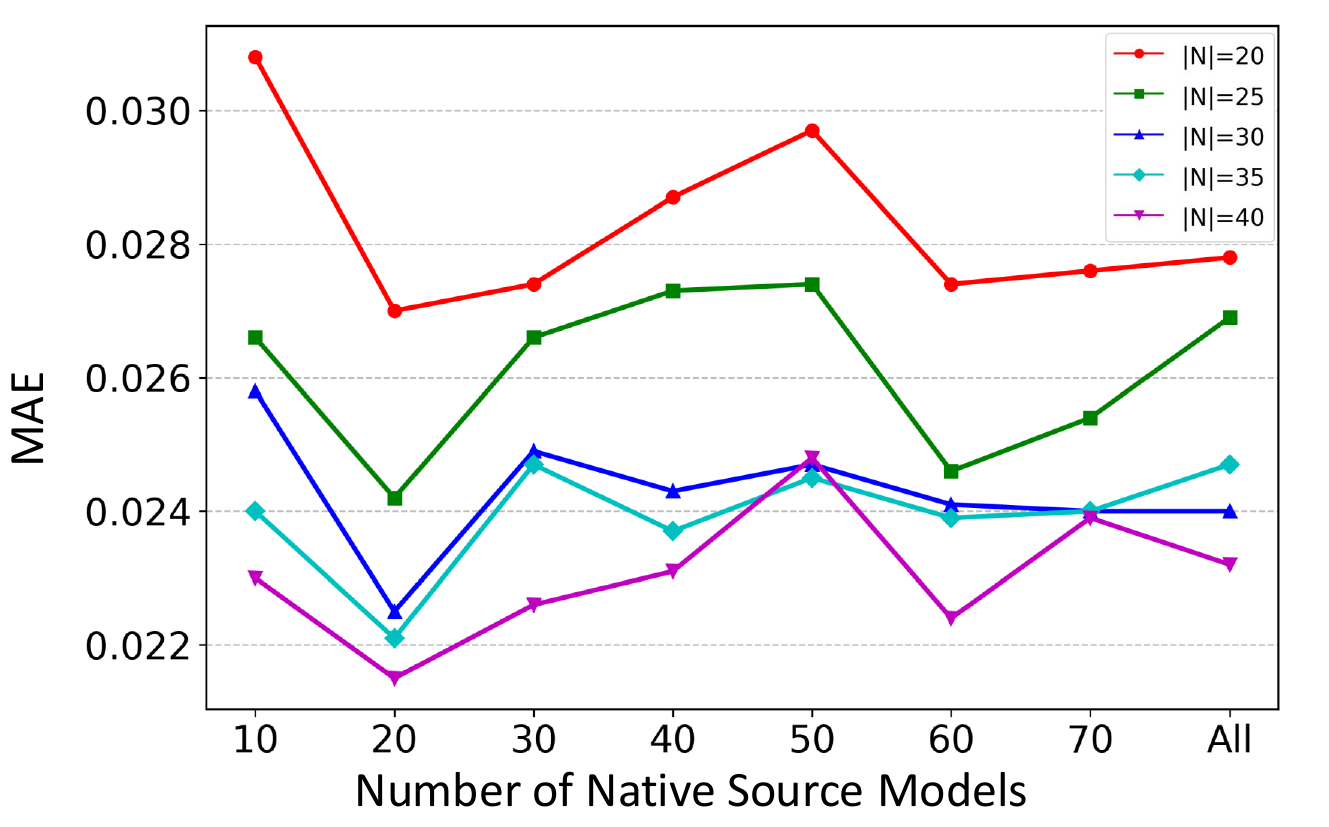}
  \caption {Performance of \textsc{TailoredBench} with varying numbers of Native Source Models on Winogrande benchmark.}
\vspace{-0.3cm}
\label{mainanalysis_winogrande}
\end{figure*}

\begin{figure*}[ht]
  \includegraphics[width=0.49\linewidth]{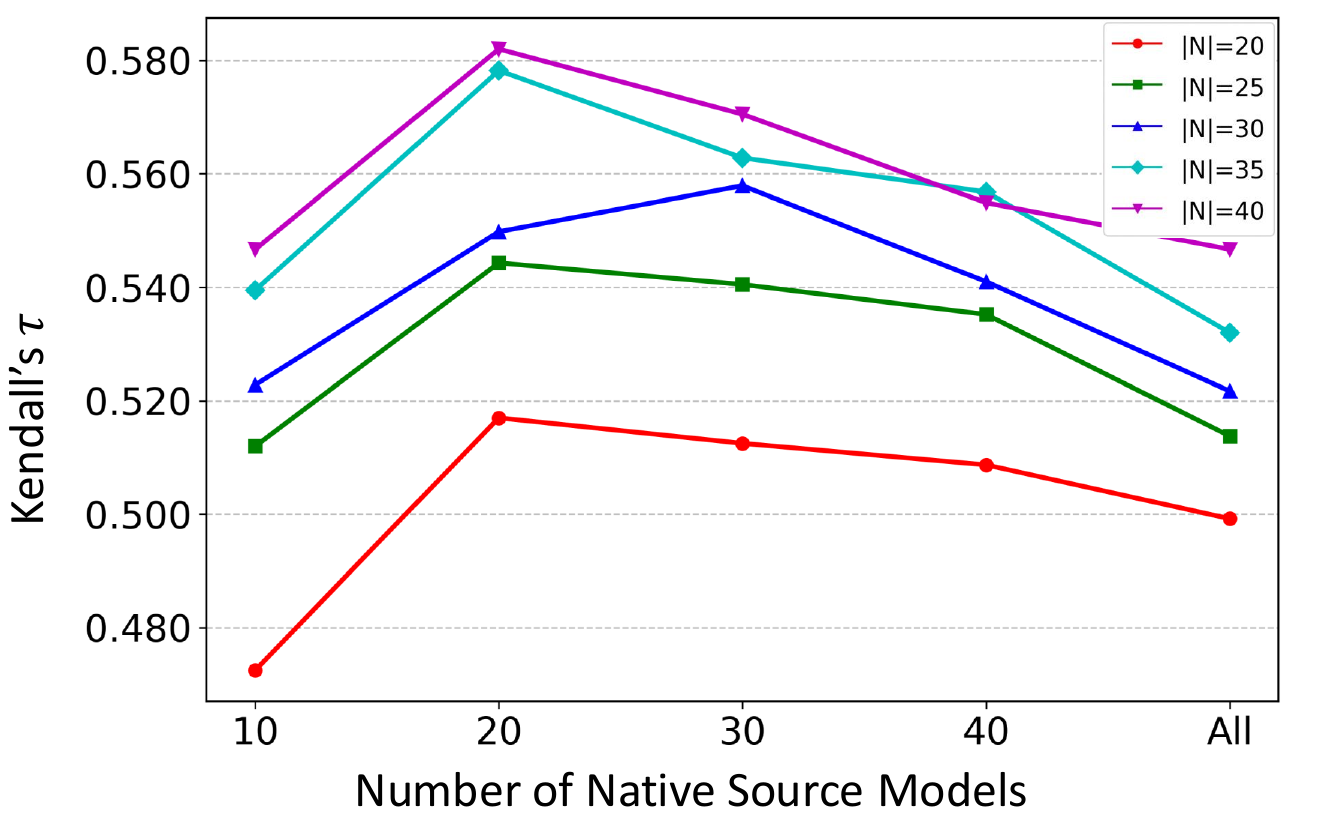} \hfill
  \includegraphics[width=0.49\linewidth]{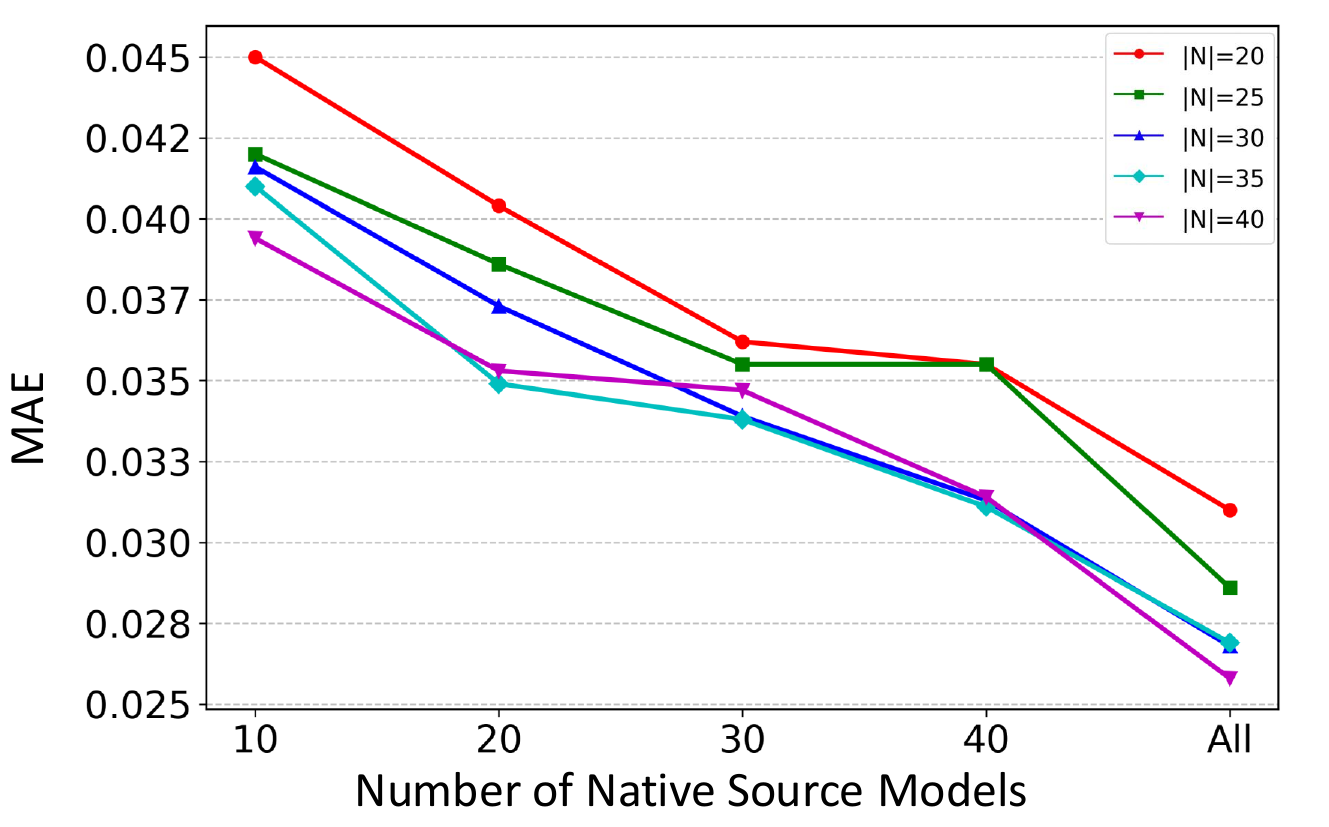}
  \caption {Performance of \textsc{TailoredBench} with varying numbers of Native Source Models on POPE benchmark.}
\vspace{-0.3cm}
\label{mainanalysis_pope}
\end{figure*}

This section presents the results of our method as the number of native source models is incrementally increased based on their prediction consistency with the target model. The results in Figures \ref{mainanalysis_arc} to \ref{mainanalysis_pope} show that, overall, Kendall's $\tau$ initially increases and then decreases as the number of native source models increases, while the MAE initially decreases and then increases with the increase in the number of native source models. 

Moreover, Our method adaptively selects 45 native source models for the ARC Challenge benchmark, 40 for the Hellaswag benchmark, 33 for the Winogrande benchmark, and 35 for the POPE benchmark. These selections represent near-optimal numbers of native source models, as demonstrated in Figures \ref{mainanalysis_arc} to \ref{mainanalysis_pope}.
\label{apd:main_analysis}



\section{Models Used in Our Experiments}
\label{apd:modellist}
Tables \ref{apdtab:arc-model-list}, \ref{apdtab:hella-model-list}, \ref{apdtab:winoandGSM8K-model-list}, \ref{apdtab:pope-model-list} provide comprehensive lists of models corresponding to each benchmark.
\begin{table*}[htbp]
    \renewcommand{\arraystretch}{1.2}
    \centering
    \small
    \setlength{\tabcolsep}{0.5em} 
    \begin{tabular}{p{2cm}p{12cm}}
        \toprule
        \textbf{Benchmark} & \textbf{Model Names} \\
        \midrule
        ARC Challenge & Qwen2-72B-Instruct, Meta-Llama-3-70B-Instruct, Qwen2-72B, zephyr-orpo-141b-A35b-v0.1, Phi-3-medium-4k-instruct, Yi-1.5-34B-Chat, c4ai-command-r-plus, Qwen1.5-110B, Smaug-72B-v0.1, Qwen1.5-110B-Chat, Yi-1.5-9B-Chat, Qwen1.5-32B-Chat, Nous-Hermes-2-Mixtral-8x7B-DPO, deepseek-llm-67b-chat, Qwen1.5-32B, Yi-1.5-34B-32K, Meta-Llama-3-70B, Phi-3-mini-4k-instruct, mixtral-8x22B-v0.3, Mixtral-8x22B-v0.1, Phi-3-mini-128k-instruct, Yi-1.5-34B, c4ai-command-r-v01, Qwen2-7B-Instruct, Hermes-2-Theta-Llama-3-8B, aya-23-35B, Mixtral-8x7B-Instruct-v0.1, notux-8x7b-v1, Meta-Llama-3-8B-Instruct, Yi-34B-Chat, Smaug-34B-v0.1, Qwen2-7B, Nous-Hermes-2-SOLAR-10.7B, K2-Chat, Yi-1.5-9B-Chat-16K, Llama-3-Refueled, WizardLM-70B-V1.0, Yi-34B, Yi-1.5-6B-Chat, NeuralDaredevil-8B-abliterated, Yi-1.5-9B, Nous-Hermes-2-Mixtral-8x7B-SFT, Hermes-2-Pro-Mistral-7B, Hermes-2-Pro-Llama-3-8B, openchat\_3.5, neural-chat-7b-v3-2, OpenHermes-2-Mistral-7B, OpenHermes-2.5-Mistral-7B, Qwen1.5-14B-Chat, Nous-Hermes-2-Mistral-7B-DPO, neural-chat-7b-v3-1, Starling-LM-7B-alpha, Qwen1.5-14B, neural-chat-7b-v3-3, Yi-34B-200K, SOLAR-10.7B-Instruct-v1.0, Yi-1.5-9B-32K, Mixtral-8x7B-v0.1, Mistral-7B-Instruct-v0.3, zephyr-7b-alpha, Mistral-7B-Instruct-v0.2, dolphin-2.9-llama3-8b, Llama-2-70b-hf, Orca-2-13b, Llama-3-8B-Instruct-Gradient-1048k, neural-chat-7b-v3, zephyr-7b-beta, Mistral-7B-OpenOrca, Yi-9B, Yi-9B-200K, DeciLM-7B-instruct, gemma-1.1-7b-it, SOLAR-10.7B-v1.0, merlinite-7b, Qwen1.5-7B-Chat, 14B, Yi-1.5-6B, stablelm-2-12b-chat, aya-23-8B, zephyr-7b-gemma-v0.1, Yarn-Solar-10b-32k, phi-2, phixtral-2x2\_8, gemma-7b, Qwen1.5-7B, WizardLM-13B-V1.2, LLaMA-Pro-8B-Instruct, Yarn-Solar-10b-64k, DeciLM-7B, OrpoLlama-3-8B, Qwen1.5-MoE-A2.7B-Chat, deepseek-llm-7b-chat, Mistral-7B-v0.1, CollectiveCognition-v1.1-Mistral-7B, Mistral\_Pro\_8B\_v0.1, Mistral-7B-v0.3, Orca-2-7b, Mistral-7B-v0.2, Yi-6B-Chat, Qwen2-1.5B-Instruct, stablelm-2-12b, openchat\_v3.2, falcon-11B, Yi-6B, Mistral-7B-Instruct-v0.1, Yarn-Mistral-7b-64k, Meta-Llama-3-8B, Yarn-Mistral-7b-128k, gemma-7b-it, openchat\_v3.2\_super, Llama-2-70b-chat-hf, Qwen1.5-MoE-A2.7B, stablelm-zephyr-3b, Qwen1.5-4B-Chat, starcoder2-15b, OpenHermes-13B, MetaMath-Mistral-Pro, Yi-6B-200K, falcon-40b, Qwen1.5-4B, Llama-2-13b-chat-hf, Llama-2-13b-hf, vicuna-7b-v1.5, OLMo-7B-Instruct-hf, internlm2-chat-1\_8b, falcon-40b-instruct, Qwen2-1.5B, deepseek-moe-16b-chat, OpenHermes-7B, Llama-2-7b-chat-hf, Nous-Hermes-llama-2-7b, stablelm-2-zephyr-1\_6b, Qwen1.5-1.8B, Qwen1.5-1.8B-Chat, LLaMA-Pro-8B, Llama-2-7b-hf, stablelm-2-1\_6b-chat, internlm2-1\_8b, Yarn-Llama-2-13b-128k, NexusRaven-V2-13B, starcoder2-7b, Llama-2-7B-32K-Instruct, deepseek-llm-7b-base, recurrentgemma-2b-it, gemma-1.1-2b-it, granite-7b-base, deepseek-moe-16b-base, gemma-2b, stablelm-3b-4e1t, gemma-2b-it, Yarn-Llama-2-7b-64k, Qwen2-0.5B, phi-1\_5 \\
        \bottomrule
    \end{tabular}
    \caption{Models used for ARC Challenge benchmark.}
    \label{apdtab:arc-model-list}
\end{table*}

\begin{table*}[htbp]
    \renewcommand{\arraystretch}{1.2}
    \centering
    \small
    \setlength{\tabcolsep}{0.5em} 
    \begin{tabular}{p{2cm}p{12cm}}
        \toprule
        \textbf{Benchmark} & \textbf{Model Names} \\
        \midrule
        HellaSwag & LLaMAntino-3-ANITA-8B-Inst-DPO-ITA, luxia-21.4b-alignment-v1.0, UNA-ThePitbull-21.4-v1, T3Q-ko-solar-dpo-v6.0, MultiVerse\_70B, RoleBeagle-11B, Capricorn-7B-DPO, Tess-2.0-Llama-3-70B, Truthful\_DPO\_MOE\_19B, multimaster-7b-v5, guanaco-65B-HF, FusionNet\_34Bx2\_MoE\_v0.1, Mixtral-8x7B-v0.1, Evangelion-7B, Lumina-5.5-Instruct, Mistral-Hermes-2x7b, Bagel-Hermes-2x34B, shqiponja-15b-v1, CollectiveCognition-v1.1-Mistral-7B-dare-0.85, etri-ones-solar, mpt-30b-instruct, openbuddy-mixtral-7bx8-v18.1-32k, bagel-dpo-7b-v0.4, OpenHermes-2.5-Mistral-7B, NeuralHermes-2.5-Mistral-7B, dolphin-2.1-mistral-7b-snr-math-laser, NeuralHermes-2.5-Mistral-7B, openbuddy-qwen1.5-32b-v21.1-32k, internlm2-20b-llama, Matter-0.2-7B-DPO, airoboros-13b-gpt4-1.2, L3-SnowStorm-v1.15-4x8B-B, Pallas-0.5-LASER-0.6, BgGPT-7B-Instruct-v0.1, Seagull-llama-3-8B-orpo-v0.5, vigogne-7b-instruct, Llama-2-7b-chat-hf-activity-fine-tuned-v4, Llama-2-7b-chat-hf-activity-fine-tuned-v3, vicuna-class-tutor-7b-ep3, Llama-2-7b-chat-hf-afr-200step-flan-v2, llama3-8b-instruct-align-test1-kto, MFANN3bv0.7, openbuddy-yi1.5-9b-v21.1-32k, openbuddy-mixtral-7bx8-v17.1-32k, odia\_llama2\_7B\_base, MT7Bi-alpha-dpo-v0.2, llama-shishya-7b-ep3-v2, Instruct\_Yi-6B\_Dolly15K, Gaja-v2.00-dpo, phi-2-OpenHermes-2.5, lion-gemma-7b-cn-v2, ToRoLaMa-7b-v1.0, gogpt-7b, Amber, open\_llama\_3b\_v2, openllama\_3b\_EvolInstruct\_lora\_merged, gemma-7B-it-firefly, Qwen1.5-4B, google-gemma-7b-it-dpo-v1, openhermes-2b-gemma-sft-qlora, RedPajama-INCITE-Chat-3B-v1, mistral\_v1, gpt-j-6b, GPT-J-Pyg\_PPO-6B, ScarletPajama-3B-HF, LLama2-7B-Structural-Prune-1.5x, illuni-llama-2-ko-7b-test, RedPajama-INCITE-Chat-3B-ShareGPT-11K, RedPajama-INCITE-Base-3B-v1, Guanaco-3B-Uncensored-v2-GPTQ, glaive-coder-7b, xglm-7.5B, gpt-sw3-6.7b, cisco-iNAM-1.1B, pythia-2.7b, qd-phi-1\_5, pythia-2.8b-deduped, LLmRa-2.7B, Tinyllama-1.3B-Cinder-Reason-Test-2, TinyPoliticaLlama-1.1B, Galpaca-30b-MiniOrca, finetune\_test\_qwen15-1-8b-sft-lora, TinyLlama-1.1B-Chat-v0.3, TinyLlama-1.1B-Chat-v0.1, CroissantLLMBase, pygmalion-2.7b, blossom-v2-3b, falcon\_1b\_stage3, MiniMerlin-3b-v0.1, DPO-miniguanaco-1.5T, CodeQwen1.5-7B-Chat, yayi2-30b-llama, rho-math-1b-v0.1, LLmRa-1.3B\_V2, TinyLlama-1.1B-intermediate-step-480k-1T, gemma-2b-ko-dev-pbmt192, gpt2-chatbot, CodeLlama-7b-Python-hf, Deita-500m, TinyWand-SFT, tinyllama-coder-py-v13, d-Qwen1.5-1.8B, TinyLlama-1.1B-intermediate-step-240k-503b, dlite-v1-1\_5b, pythia-1b-deduped, gpt2-large, WizardCoder-Guanaco-15B-V1.0, Qwen1.5-0.5B-vortex-v2, Sailor-0.5B-Chat, WizardCoder-Guanaco-15B-V1.1, Alpaca\_refine\_gpt2\_e1\_se0, deepseek-coder-1.3b-chat, speechless-coder-ds-1.3b, Instruct\_GPT, deepseek-coder-1.3b-chat-and-function-calling, megatron-gpt2-345m, starcoderbase-3b, dlite-v1-355m, gov-qna-ko-merged, SSH\_355M, CodeLlama-34b-Instruct-hf, CodeLlama-34B-Instruct-fp16, mptk-1b, KoAlpaca-Polyglot-5.8B, Llama-160M-Chat-v1, llama-160m, CodeLlama-34b-hf, KoAlpaca-KoRWKV-6B, Quokka\_590m, pruned-yi-3b-prerelease-ckpt01, gpt2\_test, finetuned-gpt2-tiny, Kaori-34b-v2, kaori-34b-v4, tiny\_starcoder\_py, GPT-2-Large-51k-steps, DialoGPT-small, test\_mistral2, pythia-31m-KI\_v1-2048-scratch \\
        \bottomrule
    \end{tabular}
    \caption{Models used for Hellaswag benchmark.}
    \label{apdtab:hella-model-list}
\end{table*}

\begin{table*}[htbp]
    \renewcommand{\arraystretch}{1.2}
    \centering
    \small
    \setlength{\tabcolsep}{0.5em} 
    \begin{tabular}{p{2cm}p{12cm}}
        \toprule
        \textbf{Benchmark} & \textbf{Model Names} \\
        \midrule
        GSM8K \& Winogrande & ExtremeDolphin-MoE, Mistral-7B-Instruct-v0.2-sparsity-20, PiVoT-SUS-RP, polyglot-math-4x7b, SOLAR-10B-Nector-DPO-Jawade, Starling-LM-11B-alpha, NeuralPipe-7B-slerp, oswald-7b, MistralTrixTest, Sensualize-Mixtral-bf16, Sensualize-Solar-10.7B, FusionNet\_passthrough, finance-chat, Kunoichi-7B, dolphin-2.2.1-mistral-7b, CarbonVillain-en-10.7B-v3, xDAN-SlimOrca, Mistral-11B-v0.1, dm7b\_sft\_gpt88w\_merge, Loyal-Macaroni-Maid-7B, Yi-34B-200K-DARE-merge-v5, WinterGoddess-1.4x-70B-L2, vicuna-class-shishya-ac-hal-13b-ep3, Kaori-34B-v1, mistral-megamerge-dare-7b, Chupacabra-8x7B-MoE, bagel-7b-v0.1, Mixtral-8x7B-v0.1, openbuddy-deepseek-67b-v15-base, Falkor-7b, synapsellm-7b-mistral-v0.4-preview3, llama2-13b-ft-openllm-leaderboard-v1, synapsellm-7b-mistral-v0.3-preview, Tess-M-v1.3, monika-ddlc-7b-v1, speechless-mistral-7b-dare-0.85, mistral-7b-v0.1-layla-v1, Mistral-v0.1-PeanutButter-v0.0.2-7B, chronos-70b-v2, L2-7b-Beluga-WVG-Test, llama-2-13b-FINETUNE3\_3.3w-r8-gate\_up\_down, airoboros-c34b-2.2.1, llama-2-13b-FINETUNE4\_3.8w-r8-q\_k\_v\_o, llama-2-13b-FINETUNE3\_3.3w-r16-gate\_up\_down, MLewd-Chat-v2-13B, Mistral-7B-v0.1-Open-Platypus, llama-2-13b-FINETUNE1\_17w-r4, EverythingLM-13b-V3-peft, Llama2-7B-guanaco-1k, llama-2-13b-FINETUNE4\_3.8w-r8-q\_k\_v\_o\_gate\_up\_down, Koss-7B-chat, ReMM-v2.2-L2-13B, WizardLM-1.0-Uncensored-CodeLlama-34b, airoboros-13b, airoboros-7b-gpt4-1.4.1-qlora, Wizard-Vicuna-7B-Uncensored-HF, Luban-Platypus2-13B-QLora-0.80-epoch, CodeLlama-34b-hf, airoboros-33b-gpt4-m2.0, llama2-22b-blocktriangular, GPT-JT-6B-v0, llama2-70b-oasst-sft-v10, vigogne-7b-instruct, based-30b, mpt-30b-chat, qCammel-70x, GiftedConvo13bLoraNoEconsE4, llama-2-13b-platypus-vicuna-wizard, GOAT-7B-Community, genz-13b-v2, chronolima-airo-grad-l2-13B, Vicuna-13B-CoT, Llama-2-7b-ft-instruct-es, OpenOrca-Preview1-13B, Tulpar-7b-v0, zephyr-7b-sft-full, Mixtral-Orca-v0.1, Marcoroni-7b-DPO-Merge, Aquila2-34B, SOLAR-10.7B-Instruct-v1.0-128k, dolphin-2.6-mistral-7b-dpo-orca-v3, flux-7b-v0.1, Turdus, A0110, yayi2-30b-llama, NeuralMarcoro14-7B, Deacon-34b-Adapter, test0, Pallas-0.5-LASER-0.4, Marcoro14-7B-ties, Antares-11b-v1, CodegebraGPT-10b, Mistral-Syndicate-7B, Nous-Hermes-2-Yi-34B, Half-NSFW\_Noromaid-7b, neural-chat-7b-v3-3-wizardmath-dare-me, apricot-wildflower-20, SauerkrautLM-UNA-SOLAR-Instruct, kalomaze-stuff, Walter-Mistral-7B, Starling-LM-alpha-8x7B-MoE, una-neural-chat-v3-3-P2-OMA, Dans-07YahooAnswers-7b, Chupacabra-7B-v2.03, PlatYi-34B-200K-Q, chinese-alpaca-2-13b-16k, ALMA-7B-Ja-V2, speechless-code-mistral-7b-v2.0, Mistral7B\_adaptor\_v1, notus-7b-v1, Chupacabra-7B-v2, SciPhi-Self-RAG-Mistral-7B-32k, Ferret-7B, llama-2-13B-instructed, glaive-coder-7b, Mistralic-7B-1, kuchiki-l2-7b, llama\_7b\_lora, Slerpeno, Llama2-7b-openorca-mc-v2-dpo, CAMEL-13B-Role-Playing-Data, starchat-beta, testmodel2, Huginn-13b-v1.2, Dans-AdventurousWinds-7b, Wizard-Vicuna-13B-Uncensored-HF, Llama-2-13b-hf-ds\_wiki\_1024\_full\_r\_64\_alpha\_16\_merged, Emerald-13B, koala-13B-HF, tulu-7B-fp16, airoboros-c34b-2.1, airoboros-7b-gpt4-1.1, 13B-Chimera, Nous-Hermes-Platypus2-13B-QLoRA-0.80-epoch, airoboros-l2-7b-gpt4-m2.0, llama-7b, llama-65b-instruct, Flash-Llama-7B, StableBeluga-13B, huginnv1.2, llama\_13b\_sharegpt94k\_fastchat, CAMEL-13B-Combined-Data, MelangeC-70b, chronos-13b-v2, stack-llama-2, CodeLlama-34b-Python-hf, UltraLM-65b, Platypus-30B, bimoGPT-llama2-13b, test-llama2-7b \\
        \bottomrule
    \end{tabular}
    \caption{Models used for GSM8K and Winogrande benchmark.}
    \label{apdtab:winoandGSM8K-model-list}
\end{table*}

\begin{table*}[htbp]
    \renewcommand{\arraystretch}{1.2}
    \centering
    \small
    \setlength{\tabcolsep}{0.5em} 
    \begin{tabular}{p{2cm}p{12cm}}
        \toprule
        \textbf{Benchmark} & \textbf{Model Names} \\
        \midrule
        POPE & InternVL2-76B, paligemma-3b-mix-448, InternVL-Chat-V1-5, cambrian\_13b, cogvlm-chat, CloudWalk, Ovis1.5-Gemma2-9B, cambrian\_8b, InternVL2-26B, Ovis1.5-Llama3-8B, llava\_next\_vicuna\_13b, glm-4v-9b, emu2\_chat, llava\_next\_mistral\_7b, llava\_next\_vicuna\_7b, WeMM, cambrian\_34b, llava\_next\_llama3, 360VL-70B, Bunny-llama3-8B, GLM4V, MiniCPM-V-2, llava\_next\_qwen\_32b, Yi-Vision, InternVL2-2B, GeminiPro1-5, InternVL2-8B, llava\_next\_interleave\_7b\_dpo, XComposer2d5, MiniCPM-V-2\_6, Mini-InternVL-Chat-2B-V1-5, cogvlm2-llama3-chat-19B, llava\_next\_yi\_34b, Step1V, InternVL2-1B, InternVL2-4B, Phi-3-Vision, llava\_next\_interleave\_7b, monkey-chat, OmniLMM\_12B, InternVL2-40B, idefics2\_8b, deepseek\_vl\_7b, GPT4o\_20240806, sharecaptioner, monkey, llava-v1.5-7b-xtuner, GPT4o\_HIGH, RekaEdge, GPT4o, Mantis-8B-Idefics2, MiniCPM-Llama3-V-2\_5, llava-llama-3-8b, sharegpt4v\_7b, Mini-InternVL-Chat-4B-V1-5, llava-internlm-7b, llava-v1.5-13b-xtuner, sharegpt4v\_13b, llava\_v1.5\_7b, GPT4o\_MINI, deepseek\_vl\_1.3b, RekaFlash, llava\_v1.5\_13b, Mantis-8B-siglip-llama3, MiniCPM-V, QwenVLPlus, Mantis-8B-clip-llama3, Yi\_VL\_6B, llava-internlm2-20b, XComposer2\_1.8b, mPLUG-Owl2, GPT4V, Yi\_VL\_34B, llava-internlm2-7b, Claude3-5V\_Sonnet, MMAlaya, instructblip\_7b, XComposer2, XComposer2\_POPE\_TEST, TransCore\_M, Claude3V\_Haiku, Claude3V\_Sonnet, Claude3V\_Opus, idefics\_9b\_instruct, chameleon\_30b, QwenVLMax, qwen\_chat, llava\_v1\_7b, PandaGPT\_13B, qwen\_base, XComposer, MiniGPT-4-v1-7B, VisualGLM\_6b, flamingov2, MiniGPT-4-v2, VXVERSE, idefics\_80b\_instruct, chameleon\_7b, XComposer2\_4KHD \\
        \bottomrule
    \end{tabular}
    \caption{Models used for POPE benchmark.}
    \label{apdtab:pope-model-list}
\end{table*}

%% file: acl_latex.bbl
\begin{thebibliography}{28}
\providecommand{\natexlab}[1]{#1}

\bibitem[{Awadalla et~al.(2022)Awadalla, Wortsman, Ilharco, Min, Magnusson, Hajishirzi, and Schmidt}]{awadalla2022exploring}
Anas Awadalla, Mitchell Wortsman, Gabriel Ilharco, Sewon Min, Ian Magnusson, Hannaneh Hajishirzi, and Ludwig Schmidt. 2022.
\newblock Exploring the landscape of distributional robustness for question answering models.
\newblock \emph{arXiv preprint arXiv:2210.12517}.

\bibitem[{Baek et~al.(2022)Baek, Jiang, Raghunathan, and Kolter}]{baek2022agreement}
Christina Baek, Yiding Jiang, Aditi Raghunathan, and J~Zico Kolter. 2022.
\newblock Agreement-on-the-line: Predicting the performance of neural networks under distribution shift.
\newblock \emph{Advances in Neural Information Processing Systems}, 35:19274--19289.

\bibitem[{Beeching et~al.(2023)Beeching, Fourrier, Habib, Han, Lambert, Rajani, Sanseviero, Tunstall, and Wolf}]{open-llm-leaderboard}
Edward Beeching, Clémentine Fourrier, Nathan Habib, Sheon Han, Nathan Lambert, Nazneen Rajani, Omar Sanseviero, Lewis Tunstall, and Thomas Wolf. 2023.
\newblock Open llm leaderboard.
\newblock \url{https://huggingface.co/spaces/open-llm-leaderboard-old/open_llm_leaderboard}.

\bibitem[{Clark et~al.(2018)Clark, Cowhey, Etzioni, Khot, Sabharwal, Schoenick, and Tafjord}]{clark2018arc}
Peter Clark, Isaac Cowhey, Oren Etzioni, Tushar Khot, Ashish Sabharwal, Carissa Schoenick, and Oyvind Tafjord. 2018.
\newblock Think you have solved question answering? try arc, the ai2 reasoning challenge.
\newblock \emph{arXiv preprint arXiv:1803.05457}.

\bibitem[{Cobbe et~al.(2021)Cobbe, Kosaraju, Bavarian, Chen, Jun, Kaiser, Plappert, Tworek, Hilton, Nakano et~al.}]{cobbe2021GSM8K}
Karl Cobbe, Vineet Kosaraju, Mohammad Bavarian, Mark Chen, Heewoo Jun, Lukasz Kaiser, Matthias Plappert, Jerry Tworek, Jacob Hilton, Reiichiro Nakano, et~al. 2021.
\newblock Training verifiers to solve math word problems.
\newblock \emph{arXiv preprint arXiv:2110.14168}.

\bibitem[{Contributors(2023)}]{2023opencompass}
OpenCompass Contributors. 2023.
\newblock Opencompass: A universal evaluation platform for foundation models.
\newblock \url{https://github.com/open-compass/opencompass}.

\bibitem[{Hu et~al.(2023)Hu, Liu, Han, Zhang, He, Zhao, Lin, Ding, Ou, Zeng et~al.}]{hu2023predicting}
Shengding Hu, Xin Liu, Xu~Han, Xinrong Zhang, Chaoqun He, Weilin Zhao, Yankai Lin, Ning Ding, Zebin Ou, Guoyang Zeng, et~al. 2023.
\newblock Predicting emergent abilities with infinite resolution evaluation.
\newblock In \emph{The Twelfth International Conference on Learning Representations}.

\bibitem[{Isik et~al.(2024)Isik, Ponomareva, Hazimeh, Paparas, Vassilvitskii, and Koyejo}]{isik2024scaling}
Berivan Isik, Natalia Ponomareva, Hussein Hazimeh, Dimitris Paparas, Sergei Vassilvitskii, and Sanmi Koyejo. 2024.
\newblock Scaling laws for downstream task performance of large language models.
\newblock \emph{arXiv preprint arXiv:2402.04177}.

\bibitem[{Kaufman and Rousseeuw(2009)}]{pam}
Leonard Kaufman and Peter~J Rousseeuw. 2009.
\newblock \emph{Finding groups in data: an introduction to cluster analysis}.
\newblock John Wiley \& Sons.

\bibitem[{Li et~al.(2023)Li, Du, Zhou, Wang, Zhao, and Wen}]{li2023pope}
Yifan Li, Yifan Du, Kun Zhou, Jinpeng Wang, Wayne~Xin Zhao, and Ji-Rong Wen. 2023.
\newblock Evaluating object hallucination in large vision-language models.
\newblock \emph{arXiv preprint arXiv:2305.10355}.

\bibitem[{Liang et~al.(2022)Liang, Bommasani, Lee, Tsipras, Soylu, Yasunaga, Zhang, Narayanan, Wu, Kumar et~al.}]{liang2022holistic}
Percy Liang, Rishi Bommasani, Tony Lee, Dimitris Tsipras, Dilara Soylu, Michihiro Yasunaga, Yian Zhang, Deepak Narayanan, Yuhuai Wu, Ananya Kumar, et~al. 2022.
\newblock Holistic evaluation of language models.
\newblock \emph{arXiv preprint arXiv:2211.09110}.

\bibitem[{Mehra et~al.(2024)Mehra, Saxena, Kim, Baek, Kolter, and Raghunathan}]{mehra2024predicting}
Aman Mehra, Rahul Saxena, Taeyoun Kim, Christina Baek, Zico Kolter, and Aditi Raghunathan. 2024.
\newblock Predicting the performance of foundation models via agreement-on-the-line.
\newblock \emph{arXiv preprint arXiv:2404.01542}.

\bibitem[{Miller et~al.(2021)Miller, Taori, Raghunathan, Sagawa, Koh, Shankar, Liang, Carmon, and Schmidt}]{miller2021accuracy}
John~P Miller, Rohan Taori, Aditi Raghunathan, Shiori Sagawa, Pang~Wei Koh, Vaishaal Shankar, Percy Liang, Yair Carmon, and Ludwig Schmidt. 2021.
\newblock Accuracy on the line: on the strong correlation between out-of-distribution and in-distribution generalization.
\newblock In \emph{International conference on machine learning}, pages 7721--7735. PMLR.

\bibitem[{Ouyang et~al.(2022)Ouyang, Wu, Jiang, Almeida, Wainwright, Mishkin, Zhang, Agarwal, Slama, Ray, Schulman, Hilton, Kelton, Miller, Simens, Askell, Welinder, Christiano, Leike, and Lowe}]{insgpt}
Long Ouyang, Jeffrey Wu, Xu~Jiang, Diogo Almeida, Carroll~L. Wainwright, Pamela Mishkin, Chong Zhang, Sandhini Agarwal, Katarina Slama, Alex Ray, John Schulman, Jacob Hilton, Fraser Kelton, Luke Miller, Maddie Simens, Amanda Askell, Peter Welinder, Paul~F. Christiano, Jan Leike, and Ryan Lowe. 2022.
\newblock \href {http://papers.nips.cc/paper\_files/paper/2022/hash/b1efde53be364a73914f58805a001731-Abstract-Conference.html} {Training language models to follow instructions with human feedback}.
\newblock In \emph{Advances in Neural Information Processing Systems 35: Annual Conference on Neural Information Processing Systems 2022, NeurIPS 2022, New Orleans, LA, USA, November 28 - December 9, 2022}.

\bibitem[{Pacchiardi et~al.(2024)Pacchiardi, Cheke, and Hern{\'a}ndez-Orallo}]{100instances}
Lorenzo Pacchiardi, Lucy~G Cheke, and Jos{\'e} Hern{\'a}ndez-Orallo. 2024.
\newblock 100 instances is all you need: predicting the success of a new llm on unseen data by testing on a few instances.
\newblock \emph{arXiv preprint arXiv:2409.03563}.

\bibitem[{Perlitz et~al.(2023)Perlitz, Bandel, Gera, Arviv, Ein-Dor, Shnarch, Slonim, Shmueli-Scheuer, and Choshen}]{perlitz2023efficient}
Yotam Perlitz, Elron Bandel, Ariel Gera, Ofir Arviv, Liat Ein-Dor, Eyal Shnarch, Noam Slonim, Michal Shmueli-Scheuer, and Leshem Choshen. 2023.
\newblock Efficient benchmarking (of language models).
\newblock \emph{arXiv preprint arXiv:2308.11696}.

\bibitem[{Polo et~al.(2024)Polo, Weber, Choshen, Sun, Xu, and Yurochkin}]{tiny}
Felipe~Maia Polo, Lucas Weber, Leshem Choshen, Yuekai Sun, Gongjun Xu, and Mikhail Yurochkin. 2024.
\newblock \href {https://openreview.net/forum?id=qAml3FpfhG} {tinybenchmarks: evaluating llms with fewer examples}.
\newblock In \emph{Forty-first International Conference on Machine Learning, {ICML} 2024, Vienna, Austria, July 21-27, 2024}. OpenReview.net.

\bibitem[{Prabhu et~al.(2024)Prabhu, Udandarao, Torr, Bethge, Bibi, and Albanie}]{prabhu2024lifelong}
Ameya Prabhu, Vishaal Udandarao, Philip Torr, Matthias Bethge, Adel Bibi, and Samuel Albanie. 2024.
\newblock Lifelong benchmarks: Efficient model evaluation in an era of rapid progress.
\newblock \emph{arXiv preprint arXiv:2402.19472}.

\bibitem[{Rodgers and Nicewander(1988)}]{ref_pearsonr}
Joseph~Lee Rodgers and W.~Alan Nicewander. 1988.
\newblock \href {http://www.jstor.org/stable/2685263} {Thirteen ways to look at the correlation coefficient}.
\newblock \emph{The American Statistician}, 42(1):59--66.

\bibitem[{Ruan et~al.(2024)Ruan, Maddison, and Hashimoto}]{ruan2024observational}
Yangjun Ruan, Chris~J Maddison, and Tatsunori Hashimoto. 2024.
\newblock Observational scaling laws and the predictability of language model performance.
\newblock \emph{arXiv preprint arXiv:2405.10938}.

\bibitem[{Sakaguchi et~al.(2021)Sakaguchi, Bras, Bhagavatula, and Choi}]{sakaguchi2021winogrande}
Keisuke Sakaguchi, Ronan~Le Bras, Chandra Bhagavatula, and Yejin Choi. 2021.
\newblock Winogrande: An adversarial winograd schema challenge at scale.
\newblock \emph{Communications of the ACM}, 64(9):99--106.

\bibitem[{Taori et~al.(2020)Taori, Dave, Shankar, Carlini, Recht, and Schmidt}]{taori2020measuring}
Rohan Taori, Achal Dave, Vaishaal Shankar, Nicholas Carlini, Benjamin Recht, and Ludwig Schmidt. 2020.
\newblock Measuring robustness to natural distribution shifts in image classification.
\newblock \emph{Advances in Neural Information Processing Systems}, 33:18583--18599.

\bibitem[{Touvron et~al.(2023)Touvron, Martin, Stone, Albert, Almahairi, Babaei, Bashlykov, Batra, Bhargava, Bhosale, Bikel, Blecher, Canton{-}Ferrer, Chen, Cucurull, Esiobu, Fernandes, Fu, Fu, Fuller, Gao, Goswami, Goyal, Hartshorn, Hosseini, Hou, Inan, Kardas, Kerkez, Khabsa, Kloumann, Korenev, Koura, Lachaux, Lavril, Lee, Liskovich, Lu, Mao, Martinet, Mihaylov, Mishra, Molybog, Nie, Poulton, Reizenstein, Rungta, Saladi, Schelten, Silva, Smith, Subramanian, Tan, Tang, Taylor, Williams, Kuan, Xu, Yan, Zarov, Zhang, Fan, Kambadur, Narang, Rodriguez, Stojnic, Edunov, and Scialom}]{llama2}
Hugo Touvron, Louis Martin, Kevin Stone, Peter Albert, Amjad Almahairi, Yasmine Babaei, Nikolay Bashlykov, Soumya Batra, Prajjwal Bhargava, Shruti Bhosale, Dan Bikel, Lukas Blecher, Cristian Canton{-}Ferrer, Moya Chen, Guillem Cucurull, David Esiobu, Jude Fernandes, Jeremy Fu, Wenyin Fu, Brian Fuller, Cynthia Gao, Vedanuj Goswami, Naman Goyal, Anthony Hartshorn, Saghar Hosseini, Rui Hou, Hakan Inan, Marcin Kardas, Viktor Kerkez, Madian Khabsa, Isabel Kloumann, Artem Korenev, Punit~Singh Koura, Marie{-}Anne Lachaux, Thibaut Lavril, Jenya Lee, Diana Liskovich, Yinghai Lu, Yuning Mao, Xavier Martinet, Todor Mihaylov, Pushkar Mishra, Igor Molybog, Yixin Nie, Andrew Poulton, Jeremy Reizenstein, Rashi Rungta, Kalyan Saladi, Alan Schelten, Ruan Silva, Eric~Michael Smith, Ranjan Subramanian, Xiaoqing~Ellen Tan, Binh Tang, Ross Taylor, Adina Williams, Jian~Xiang Kuan, Puxin Xu, Zheng Yan, Iliyan Zarov, Yuchen Zhang, Angela Fan, Melanie Kambadur, Sharan Narang, Aur{\'{e}}lien Rodriguez, Robert Stojnic, Sergey Edunov,
  and Thomas Scialom. 2023.
\newblock \href {https://doi.org/10.48550/ARXIV.2307.09288} {Llama 2: Open foundation and fine-tuned chat models}.
\newblock \emph{CoRR}, abs/2307.09288.

\bibitem[{Van~der Maaten and Hinton(2008)}]{tsne}
Laurens Van~der Maaten and Geoffrey Hinton. 2008.
\newblock Visualizing data using t-sne.
\newblock \emph{Journal of machine learning research}, 9(11).

\bibitem[{Vivek et~al.(2024)Vivek, Ethayarajh, Yang, and Kiela}]{AP}
Rajan Vivek, Kawin Ethayarajh, Diyi Yang, and Douwe Kiela. 2024.
\newblock \href {https://aclanthology.org/2024.eacl-long.95} {Anchor points: Benchmarking models with much fewer examples}.
\newblock In \emph{Proceedings of the 18th Conference of the European Chapter of the Association for Computational Linguistics, {EACL} 2024 - Volume 1: Long Papers, St. Julian's, Malta, March 17-22, 2024}, pages 1576--1601. Association for Computational Linguistics.

\bibitem[{Xu et~al.(2024)Xu, Saranathan, Alam, Shah, Lim, Wong, Martin, and Bhattacharya}]{xu2024data}
Cong Xu, Gayathri Saranathan, Mahammad~Parwez Alam, Arpit Shah, James Lim, Soon~Yee Wong, Foltin Martin, and Suparna Bhattacharya. 2024.
\newblock Data efficient evaluation of large language models and text-to-image models via adaptive sampling.
\newblock \emph{arXiv preprint arXiv:2406.15527}.

\bibitem[{Zellers et~al.(2019)Zellers, Holtzman, Bisk, Farhadi, and Choi}]{zellers2019hellaswag}
Rowan Zellers, Ari Holtzman, Yonatan Bisk, Ali Farhadi, and Yejin Choi. 2019.
\newblock Hellaswag: Can a machine really finish your sentence?
\newblock \emph{arXiv preprint arXiv:1905.07830}.

\bibitem[{Zhang et~al.(2024)Zhang, Lyu, Liu, and Ma}]{zhang2024collaborative}
Qiyuan Zhang, Fuyuan Lyu, Xue Liu, and Chen Ma. 2024.
\newblock Collaborative performance prediction for large language models.
\newblock \emph{arXiv preprint arXiv:2407.01300}.

\end{thebibliography}
